%% file: main.tex
\documentclass[]{informs4} 

\OneAndAHalfSpacedXI
\TheoremsNumberedThrough
\EquationsNumberedThrough
\ECRepeatTheorems

\usepackage{amsmath,amssymb,mathtools}
\usepackage{subcaption}
\usepackage{booktabs}
\usepackage[sort]{natbib}
\bibpunct[, ]{(}{)}{,}{a}{}{,}%
\def\bibfont{\small}%
\usepackage[colorlinks=true,linkcolor=black,citecolor=blue,urlcolor=blue]{hyperref}
\usepackage{bm,enumerate}
\usepackage[capitalize]{cleveref}
\crefname{assumption}{Assumption}{Assumptions}
\Crefname{assumption}{Assumption}{Assumptions}
\usepackage{algorithm}
\usepackage[noEnd=false,indLines=false]{algpseudocodex}
\usepackage{algpseudocode}

\newcommand{\E}{\mathbb{E}}

\newcommand{\cF}{\mathcal{F}}

\newcommand{\thmref}[2]{\hyperref[#1]{#2 \ref*{#1}}}

\MANUSCRIPTNO{}
\RUNTITLE{Randomization for Risk-Sensitive Newsvendor Learning}

\renewcommand{\theARTICLETOP}{}

\begin{document}

\TITLE{Overcoming Demand Censoring through Randomization for Risk-Sensitive Newsvendor Learning}

\ARTICLEAUTHORS{
\AUTHOR{Zhengchao Wang}
\AFF{The University of Sydney, \EMAIL{zhengchao.wang@sydney.edu.au}}
\AUTHOR{Nam Ho-Nguyen}
\AFF{The University of Sydney, \EMAIL{nam.ho-nguyen@sydney.edu.au}}
}

\ABSTRACT{
We study online newsvendor learning with bounded demand, censored observations, and convex nonlinear shortage costs. Because stockouts can conceal the shortage magnitudes needed by gradient-based algorithms, we use randomized ordering to construct unbiased subgradient estimates. We characterize admissible randomization distributions within this framework and develop an online inventory policy that combines the resulting subgradient estimates with gradient-based updates. Our regret analysis quantifies the cost of randomization via the estimation variance and an additional loss term. We characterize the randomization distributions that minimize this objective. For piecewise-linear shortage costs, the optimal distribution is computed through a linear-time recursion; for costs with bounded curvature density, it is characterized by a scalar ordinary differential equation. The policy achieves expected regret of $O(T^{2/3})$ against the best fixed order in hindsight over $T$ periods. Under independent demand and suitable curvature of expected losses, the bound improves to $O(\sqrt{T})$. Matching minimax lower bounds establish rate optimality over the corresponding classes with a fixed marginal-cost bound. Both rates extend to nonperishable inventory under uniformly positive conditional expected demand. Numerical experiments on synthetic and retail data show improvements over alternative policies using censored observations.
}


\maketitle




\input{_vcurrent/introduction_polished}

\input{_vcurrent/polished_section_1}

\input{_vcurrent/polished_section_2}

\input{_vcurrent/polished_section_3}

\input{_vcurrent/numerical_section}

\input{_vcurrent/conclusion}

\bibliographystyle{informs2014}
\bibliography{bibliograph}

\newpage

\begin{APPENDICES}
\renewcommand{\theHsection}{appendix.\Alph{section}}
\input{_vcurrent/appendix}
\end{APPENDICES}

\end{document}

%% file: _vcurrent/introduction_polished.tex
\section{Introduction}
\label{sec:introduction}

Inventory decisions are often made with only a partial view of demand.
A retailer observes how many units were sold, but once the available inventory is exhausted, the retailer cannot observe how much additional demand went unserved.
This is \emph{demand censoring}.
Such situations call for carefully chosen inventory decisions that balance operating costs against the information gained for future decisions.

The quality of an inventory decision depends on various factors such as the production cost, sales revenue, holding or disposal costs for unsold inventory, and penalties for unmet demand in case of an inventory shortage.
While it is common to model shortage penalties as linear, in general the consequences of a shortage need not be proportional to the unmet demand.
A small shortage may be absorbed through routine service recovery, whereas a severe shortage may require increasingly expensive fulfillment alternatives, incur escalating service penalties, or increase the risk of reputational harm.
A nonlinear convex shortage penalty models this aspect of a \emph{risk-sensitive} decision maker who is averse to large shortages.


This difference also changes what can be learned from sales.
A gradient-based inventory policy adjusts orders using marginal loss information, which measures the effect of a small change in the order quantity.
With a linear shortage cost, the marginal cost is constant whenever demand exceeds inventory, so the sales observation provides a valid subgradient even when demand is censored.
With a nonlinear shortage cost, however, the marginal loss can depend on how much demand went unserved.
Two demand realizations may therefore produce identical sales but imply different marginal costs.
Censoring can thus hide the first-order information needed for learning, even though the inventory loss remains convex.

We study this information problem in a repeated newsvendor model with continuous quantities, a known demand bound, and known inventory costs.
Our starting point is perishable inventory, with unsold units discarded at the end of each period and only sales observed.
Demand may depend arbitrarily on the full past, including earlier orders and random draws, but is conditionally independent of the current period's order given that past.
No stationary demand distribution is required although our developed algorithm works for independent stationary and nonstationary demands.
Performance is measured by expected regret relative to the best fixed order in hindsight, evaluated on the same demand sequence.

We address the missing marginal loss information through \emph{randomization}.
The policy maintains a \emph{virtual order}, which is the quantity updated by online gradient descent, and adds a randomized nonnegative amount to obtain the \emph{physical order} placed in that period.
The additional inventory can reveal information that would remain censored at the virtual order.
Using the resulting sales and the known randomization distribution, we construct an unbiased subgradient estimate at the virtual order for each fixed demand realization.
This estimate determines the next virtual order, without estimating a demand distribution.
Loss and regret are evaluated at the physical order, so our analysis accounts for the inventory used to acquire information.

We consider estimators that use the exact subgradient when sales are strictly below the physical order quantity. 
When sales exhaust that quantity, the estimator uses a correction determined by the randomization distribution and evaluated at the realized additional inventory. 
Within this class, we establish necessary and sufficient conditions on the distribution for unbiased estimation of the loss subgradient at the virtual order for every fixed demand realization. 
We also derive an expression for the worst-case conditional variance, allowing us to compare estimation precision across randomization designs.

The distribution of additional inventory also affects the operating loss.
Thus, in each period, we choose the distribution to minimize the corresponding contribution to our regret bound: a weighted sum of the worst-case conditional variance and expected additional inventory. 
The optimal distribution depends on where and by how much the marginal shortage cost increases. 
For piecewise-linear shortage costs, an optimal distribution assigns probability only to zero and the feasible breakpoints, with probabilities computed by a linear-time forward–backward recursion. 
For shortage costs with a bounded curvature density, the optimal distribution is characterized by a scalar ordinary differential equation.

For general convex shortage costs, our randomized ordering policies achieve $O(T^{2/3})$ expected regret over $T$ periods under the adaptive demand model described above. 
A matching lower bound establishes minimax optimality over classes with a fixed upper bound on marginal shortage costs. 
This bound holds even under independent and identically distributed demand, showing that the learning difficulty persists even when demand is stationary.
Under independent demand from distributions fixed in advance but allowed to vary across periods, the regret bound improves to $O(\sqrt{T})$ when the expected losses are uniformly strongly convex, or when they share a minimizer and satisfy a uniform restricted secant inequality. 
Matching lower bounds establish minimax optimality over both structural classes with fixed curvature and marginal-cost bounds. 
When the shortage cost is linear, sales reveal the required subgradient, so additional inventory is unnecessary \citep{huhRusmevichientong2009,hihatEtAl2023}.
Our framework then recovers an $O(\sqrt{T})$ regret guarantee under the adaptive demand model.

Our method extends to nonperishable inventory, where unsold units from a period can be kept to service demand in the next period. Inventory remaining from earlier periods may exceed the current randomized order-up-to target, preventing the policy from implementing that target exactly. 
We show how to recover the sales observations required by the estimator from actual sales and bound the additional cost of inventory held above the target. 
When expected demand conditional on the full past is uniformly bounded away from zero, the policy retains $O(T^{2/3})$ expected regret relative to the best fixed base-stock policy in hindsight. 
Under the same lower bound, the faster $O(\sqrt{T})$ rate is also preserved when demands are independent across periods, and the expected losses satisfy either curvature condition stated above.

\subsection{Related work}
\label{sec:related-work}

\paragraph{Inventory learning with censored demand.}
Research on censored-demand inventory control examines how ordering decisions affect both operating costs and subsequent learning.
Bayesian studies investigate the value of information and its implications for ordering policies \citep{larivierePorteus1999,dingPutermanBisi2002,luSongZhu2008,besbesChanetonMoallemi2022}.
Nonparametric approaches instead learn inventory policies or demand distributions directly from sales observations \citep{godfreyPowell2001,huhEtAl2011,bisiKalsiAbdollahian2015}.
\citet{besbesMuharremoglu2013} show that, under their regularity conditions, censoring leaves the regret order unchanged for continuous demand, while a lost-sales indicator can improve learning rates in the discrete-demand setting.
Our setting raises a related question: whether sales reveal the marginal loss needed to improve an order when shortage costs are nonlinear.

Stochastic approximation and gradient-based policies provide a direct way to update inventory decisions \citep{kunnumkalTopaloglu2008,huhRusmevichientong2009,huhRusmevichientong2014}.
For the classical newsvendor loss with linear shortage costs, sales reveal a valid subgradient, which makes such updates possible without observing all demand.
Other first-order frameworks accommodate more general losses by assuming access to the required subgradients.
In particular, \citet{hihatEtAl2023} study inventory optimization beyond the iid setting under subgradient feedback, including systems with inventory carryover.
Our method uses online gradient descent {\citep{zinkevich2003}}, but constructs the subgradient estimates from sales and accounts for the inventory required to obtain them.

Nonstationary censored-demand learning is also studied by \citet{lugosiMarkakisNeu2024}, who use randomized exponential weighting over a finite set of inventory levels under linear overage and shortage costs.
Their loss-estimation approach establishes guarantees without a stationary demand model.
We consider continuous inventory quantities and nonlinear shortage costs, for which sales may conceal the marginal information available in the linear model.
Related work addresses capacity constraints \citep{shiChenDuenyas2016}, finite shelf lives \citep{zhangChaoShi2018}, and replenishment lead times \citep{zhangChaoShi2020,lyuZhangXin2024}; these studies address operational constraints that are complementary to the information problem considered here.

\paragraph{Inventory models with nonlinear costs.}
\citet{kyparisisKoulamas2018} characterize optimal pricing and inventory decisions in a newsvendor model with nonlinear salvage revenue and shortage costs.
\citet{huhJanakiramanEtAl2011} establish conditions for the optimality of base-stock policies in backorder systems with penalties depending on backlog quantities and customer waiting times.
\citet{sillanpaaEtAl2021} study procurement contracts with piecewise-linear shortage costs that assign higher marginal costs to larger shortages.
These studies analyze how richer cost structures affect optimal decisions under specified demand models.
They do not address the sequential learning problem that arises when sales observations hide the shortage amount and hence the marginal loss.


\citet{yangShi2023} directly address this problem for integer-valued demand and inventory under iid demand from an unknown distribution.
Their policy separates learning and optimization: designated learning periods use elevated inventory levels to estimate a truncated demand distribution, and intervening periods choose inventory levels using that estimate.
They establish regret of order $T^{2/3}$ up to logarithmic factors, together with a lower bound having the same horizon exponent.
Our policy instead estimates marginal losses through randomized additional inventory in each period, and its general regret guarantee holds {under the demand model described above}.
The decision and demand assumptions also differ: our quantities are continuous and demand has a known finite bound, whereas their analysis allows unbounded integer-valued demand under additional restrictions.
Continuous quantities require a separate lower-bound argument, because intermediate orders can sometimes acquire information at arbitrarily small additional cost.
Our lower-bound construction rules out such inexpensive information acquisition and establishes the $T^{2/3}$ lower bound for continuous orders.
Under independent demand, the expected-loss curvature conditions further identify settings in which our regret rate improves to $O(\sqrt{T})$.

\paragraph{Randomization and information acquisition.}
Randomization has several roles in operations and optimization.
The ``Dice''-sion framework of \citet{delageKuhnWiesemann2019}, randomized assortment optimization of \citet{wangPeuraWiesemann2024}, and randomized robust pricing of \citet{guanMisic2026} examine its value under risk or distributional ambiguity.
In sequential learning, randomized decisions also acquire information, as in dynamic pricing and revenue management \citep{nambiarSimchiLeviWang2019,ferreiraSimchiLeviWang2018} and Thompson sampling for the repeated newsvendor problem \citep{chenEtAl2026Thompson}.
Optimizing sampling allocations is also central to pure exploration, where the objective is accurate identification rather than cumulative operating performance \citep{russo2020BestArm,qinYou2026}.

The closest connection to our estimator construction is the comparison-based stochastic optimization framework of \citet{chenLinWang2020}.
They construct unbiased gradient estimates by comparing a sample with the current decision, then drawing an auxiliary point based on this result for a second comparison with the same sample.
They optimize sampling densities to reduce the estimator's second moment, with loss evaluated at the original decision.
However, the two-comparison approach cannot be applied to newsvendor problems in general, since we typically only observe censored demand once in each period. Our method circumvents this by constructing a subgradient estimate from this one observation.
Because the physical order also determines the operating loss, our distribution design balances estimation variance with a bound on the loss increase from additional inventory.

More broadly, the choice of an exploration distribution can be guided by its contribution to learning performance.
Information-directed sampling balances immediate regret against information acquisition \citep{russoVanRoy2018}, while exploration by optimization jointly designs exploration distributions and loss estimators through a regret-bound objective \citep{lattimoreSzepesvari2020}.
Our distribution-design problem follows this principle, with the cost of information determined by the physical inventory decision.

\paragraph{Organization.}
\Cref{rn:sec:model} presents the inventory model and randomized learning method, and \cref{rn:sec:design} develops the distribution-design results.
\Cref{rn:sec:regret} establishes the regret bounds and lower bounds and treats {nonperishable} inventory.
\Cref{sec:numerical-studies} evaluates the methods using simulated demand and recorded retail sales, and \cref{sec:conclusion} concludes.

\paragraph{Notation.}
We use tildes for scalar quantities that may be random, such as demand $\tilde d_t$ and orders $\tilde q_t$.
The corresponding untilded symbols denote realizations or generic values, and the subscript $t$ indexes periods.
We write $(r)_+\coloneqq\max\{r,0\}$ and use $\mathbf 1\{\cdot\}$ for the indicator function.

%% file: _vcurrent/polished_section_1.tex
\section{Model and learning via randomization}
\label{rn:sec:model}

\subsection{Inventory model and regret}
\label{rn:sec:inventory}

We consider a perishable-inventory problem over a fixed horizon of $T$ periods, with unsold inventory discarded at the end of each period. (The extension to nonperishable inventory will be discussed in \cref{rn:sec:non-perishable}.)
Let $0<M<\infty$ be a known upper bound on both demand and order quantities.
In each period $t\in[T]\coloneqq\{1,\ldots,T\}$, the decision maker chooses an order quantity $\tilde q_t\in[0,M]$ before demand $\tilde d_t\in[0,M]$ is realized.
The decision maker observes only sales, $\tilde s_t\coloneqq\min\{\tilde d_t,\tilde q_t\}$.
If $\tilde s_t<\tilde q_t$, demand is revealed exactly; if $\tilde s_t=\tilde q_t<M$, it reveals only that demand is at least the order quantity. For the rest of this subsection we may consider a generic stochastic demand sequence $\tilde{d}_1,\ldots,\tilde{d}_T$ that may or may not be adaptive to the ordering policy; we will clarify its assumptions in \cref{rn:sec:dependence}.

Let $c\ge0$ denote the per-unit purchase cost, $p\ge c$ the selling price, and $\kappa\ge0$ the per-unit disposal cost, with $c+\kappa>0$.
A shortage of $r$ units incurs a penalty $\varphi(r)$.
This penalty need not represent a direct monetary payment: it may capture emergency procurement, service-level violations, or damage to customer relationships.
The ``profit'' from an order $q$ at demand $d$, net of the shortage penalty, is $\Pi(q,d)\coloneqq p\min\{q,d\}-cq-\kappa(q-d)_+-\varphi((d-q)_+)$.
Define $h\coloneqq c+\kappa$ and $\psi(r)\coloneqq\varphi(r)+(p-c)r$.
Rearranging the profit expression gives $\Pi(q,d)=(p-c)d-\ell(q,d)$, where
\begin{equation}
\label{rn:eq:loss}
\ell(q,d)\coloneqq h(q-d)_++\psi((d-q)_+),\qquad q,d\in[0,M].
\end{equation}
We interpret $\ell(q,d)$ as the loss from ordering $q$ when demand is $d$: it measures the profit shortfall relative to the maximum possible profit $(p-c)d$.
The excess-inventory cost $h > 0$ includes purchase and disposal costs, while the shortage cost $\psi$ combines the forgone sales profit with the additional shortage penalty. We formally describe assumptions on $\psi$ in \cref{rn:sec:dependence}. Because the benchmark $(p-c)d$ is independent of the order quantity, maximizing cumulative profit is equivalent to minimizing cumulative loss on the same demand sequence.

We evaluate a policy by its expected regret $\mathbb E[\tilde r_T]$, where
\begin{equation}
\label{rn:eq:regret}
\tilde r_T\coloneqq\sum_{t=1}^T\ell(\tilde q_t,\tilde d_t)-\min_{q\in[0,M]}\sum_{t=1}^T\ell(q,\tilde d_t).
\end{equation}
The comparator is the best fixed order in hindsight, selected after observing the realized demand sequence.
Both the policy and the comparator are evaluated on this same sequence.
The expectation is over the policy's randomization and, when demand is random, the demand sequence.
This criterion is at least as stringent as pseudo-regret, $\mathbb{E}[\sum_{t=1}^T\ell(\tilde q_t,\tilde d_t)] -
\min_{q\in[0,M]}\mathbb{E}[\sum_{t=1}^T\ell(q,\tilde d_t)]$, which compares the policy with a fixed order minimizing expected cumulative loss.

\subsection{Virtual orders, randomization and dependence}
\label{rn:sec:first-order}
\label{rn:sec:dependence}

We seek to learn order quantities directly from sales to minimize the expected regret $\mathbb E[\tilde r_T]$, without first estimating a demand distribution.
Our starting point is projected online gradient descent~{\citep{zinkevich2003}}, which adjusts orders using {subgradients of the inventory loss}.
The algorithm maintains an order target $\tilde x_t\in[0,M]$.
If demand were fully observed, we would start from a specified $\tilde x_1\in[0,M]$, implement this target directly as $\tilde q_t=\tilde x_t$, and use the observed demand $\tilde d_t$ to update $\tilde x_{t+1}\coloneqq\operatorname{Proj}_{[0,M]}\bigl(\tilde x_t-\eta_t g(\tilde x_t;\tilde d_t)\bigr)$.
Here $g(x;d)\in\partial_x\ell(x,d)$ is a selected subgradient of the loss at inventory level $x$ with demand $d$, $\eta_t>0$ is the step size, and $\operatorname{Proj}_{[0,M]}$ projects the update onto the feasible interval.

We now introduce the following convex structure on our losses $\ell$ via the shortage cost $\psi$; see \cref{rn:eq:loss}.
\begin{assumption}
\label{rn:ass:penalty}
The function $\psi:[0,M]\to\mathbb R_+$ is continuous, convex, and nondecreasing, with $\psi(0)=0$.
Its right derivative $\psi'_+(r)$ is finite for every $r\in[0,M)$, and the endpoint limit $\lim_{r\uparrow M}\psi'_+(r)$ is finite. Write $m(r)\coloneqq\psi'_+(r)$ for $r\in[0,M)$ and $m(M)\coloneqq\lim_{r\uparrow M}\psi'_+(r)$ for the marginal shortage cost.
\end{assumption}
Allowing $\psi$ to be nonlinear and convex captures increasing marginal consequences of shortages, representing aversion to large shortages. With censored sales, however, the subgradient needed for this update may be unavailable.
To see what is missing, fix an order target $x$ and demand $d$, and suppose that we order $q=x$ and observe sales $s=\min\{d,x\}$.
Consider the following choice of subgradient:
\begin{equation}\label{rn:eq:gradient-decomposition}
\begin{aligned}
    g(x;d)
    &\coloneqq h\mathbf 1\{d<x\}
    -m((d-x)_+)\mathbf 1\{d\ge x\} 
    = h\mathbf 1\{d<x\}
    -m(0)\mathbf 1\{d\ge x\}
    -\Delta((d-x)_+)\\
    &= h\mathbf 1\{s<x\}
    -m(0)\mathbf 1\{s=x\}
    -\Delta((d-x)_+),
\end{aligned}
\end{equation}
where $m(r)$ is the marginal shortage cost and $\Delta(r)\coloneqq m(r)-m(0)$ is its increase relative to $m(0)$.
Sales determine the first two terms in the final expression.
When $s<x$, demand is observed as $d=s$, and the subgradient equals $h$.
When $s=x<M$, we know only that $d\ge x$, so the shortage amount $d-x$ and the corresponding correction $\Delta(d-x)$ may remain unknown.

One possible method is to use $-m(0)$ whenever $s = q=x$, effectively setting the missing correction $\Delta(d-x)$ to zero.
For the convex shortage costs considered here, this value need not be a valid subgradient of $\ell(\cdot,d)$ at $x$, and using it can lead to linear regret.
For example, let $M=2$, $h=1$, and $\psi(r)=r^2/2$, with $\tilde d_t=2$ in every period.
Start the projected update at $\tilde x_1=1$, place $\tilde q_t=\tilde x_t$, and use the subgradient estimate $\tilde{g}_t = -m(0)$ whenever $\tilde s_t=\tilde q_t$.
Because $m(0)=0$, every update leaves the order unchanged, so $\tilde x_t=\tilde q_t=\tilde s_t=1$ for all $t$, regardless of the positive step sizes.
The true derivative at $(\tilde x,\tilde d)=(1,2)$ is $-m(1)=-1$, so zero is not a valid subgradient.
The policy incurs $\ell(1,2)=1/2$ per period, whereas the fixed order $q=2$ incurs zero loss, giving regret $\tilde r_T=T/2$.

Our aim is therefore to construct unbiased subgradient estimates while controlling the additional inventory used to obtain information.
We propose to use random additional inventory to estimate the loss subgradient at the current order target.
Let $\tilde x_t$ denote the target updated by the learning algorithm, which we call the \emph{virtual order}.
Using the observed history, choose a distribution $Q_t$ supported on $[0,M-\tilde x_t]$, draw additional inventory $\tilde u_t\sim Q_t$, and place the \emph{physical order} $\tilde q_t\coloneqq\tilde x_t+\tilde u_t$.
The physical order determines sales and operating loss, while the resulting sales are used to estimate the loss subgradient at the virtual order.

Before introducing our methods to choose $Q_t$, we clarify the dependence structure between our variables and the demand sequence. To do so, we introduce the following notation.
\begin{definition}\label{def:sigma-algebra}
Fix a time period $t \geq 1$. Let $\mathcal H_{t-1} := \sigma\left( \{ \tilde{x}_\tau, \tilde{u}_\tau, \tilde{d}_\tau \}_{\tau=1}^{t-1} \right)$ be the sigma-algebra generated by the full history prior to $t$, and let $\mathcal F_{t-1} := \sigma\left( \{ \tilde{x}_\tau,\tilde{u}_\tau, \tilde{s}_\tau \}_{\tau=1}^{t-1} \right) \subseteq\mathcal H_{t-1}$ be the sigma-algebra generated by the policy's randomization and observed sales.
Let $v_1^d,v_1^u,\ldots,v_T^d,v_T^u$ be a collection of $2T$ standard uniform random variables. Define $\widetilde{\mathcal{H}}_t := \sigma\left(\mathcal{H}_{t-1} \cup \sigma(v_t^d) \right)$ and $\widetilde{\mathcal{F}}_t := \sigma\left(\mathcal{F}_{t-1} \cup \sigma(v_t^u) \right)$.
\end{definition}
The $v_t^d,v_t^u$ are seed random variables introduced to allow for potential randomness in $\tilde{d}_t$ and $\tilde{u}_t$ from choosing a distribution based on the relevant histories, and then simulating the randomness with $v_t^d$ and $v_t^u$ respectively. 
For our dependence structure, we will assume the following.
\begin{assumption}\label{ass:dependence}
For each $t=1,\ldots,T$, the sigma-algebras $\sigma(v_t^d)$, $\sigma(v_t^u)$, and $\mathcal H_{t-1}$ are mutually independent. Furthermore, for each $t$, $\tilde d_t$ is $\widetilde{\mathcal H}_t$-measurable, $\tilde x_t$ is $\mathcal F_{t-1}$-measurable, and $\tilde u_t$ is $\widetilde{\mathcal F}_t$-measurable. Consequently, the variables $\tilde d_t$, $\tilde x_t$, and $\tilde u_t$ are mutually independent conditional on $\mathcal H_{t-1}$. Indeed, $\tilde x_t$ is determined by this history, while $\tilde d_t$ and $\tilde u_t$ depend on independent fresh seeds.
\end{assumption}

As a refinement of \cref{ass:dependence}, we also consider \emph{independent demand}: for each $t$, $\tilde d_t$ is independent of the full history $\mathcal H_{t-1}$ and has distribution $P_t$, fixed before the first period. These distributions may vary with $t$; independent and identically distributed demand is the special case $P_t=P$ for all $t$.


\subsection{Admissible randomization}
\label{rn:sec:admissible}

Having established the appropriate dependence structure, we now describe how to construct a useful subgradient estimate by specifying $Q_t$. Fix a period $t$; for convenience we will suppress the time index. Then $\tilde{x}$ is the virtual order, $\tilde{u} \sim Q$ is the randomized additional inventory, $\tilde{q} = \tilde{x} + \tilde{u}$ is the physical order, $\tilde{d}$ is demand, and $\tilde s\coloneqq\min\{\tilde{d},\tilde q\}$ is the observed sales quantity.
When $\tilde s<\tilde q$, demand is revealed since $\tilde{d} = \tilde{s}$, allowing us to evaluate the exact subgradient $g(\tilde x;\tilde d)$.
When $\tilde s=\tilde q$, we have $\tilde d\ge\tilde q\ge \tilde x$, so, according to Eq.~\eqref{rn:eq:gradient-decomposition}, the required subgradient is $-m(0)-\Delta(\tilde d- \tilde x)$.
We retain the known term $-m(0)$ and replace the potentially unobserved correction $\Delta(\tilde d- \tilde x)$ by $W(\tilde u)$.
Define
\begin{equation}
\label{rn:eq:gradient-estimator}
\tilde g_W(x,\tilde u,\tilde s)\coloneqq
\begin{cases}
g(x;\tilde s),&\tilde s<x+\tilde u,\\
-m(0)-W(\tilde u),&\tilde s=x+\tilde u.
\end{cases}
\end{equation}

We seek a correction $W$ such that
$\mathbb E[\tilde g_W(x,\tilde u,\tilde s)\mid x,d]=g(x;d)$
for every fixed demand $d$, where the expectation is over $\tilde u$ with the history, $x$, and $Q$ held fixed.
If $d<x$, this requirement holds automatically because demand is always revealed and the estimator equals $g(x;d)=h$.
For $d\ge x$, let $r=d-x$ and let $F(r)\coloneqq Q([0,r])$ be the distribution function.
Demand is revealed when $\tilde u>r$, whereas inventory is exhausted when $\tilde u\le r$.
Since the estimator in Eq.~\eqref{rn:eq:gradient-estimator} contains the same baseline $-m(0)$ in either case, unbiasedness of $\tilde g_W(x,\tilde u,\tilde s)$ requires recovering $\Delta(r)$ in expectation.
The event on which demand is revealed contributes $\{1-F(r)\}\Delta(r)$ toward this correction, so the stockout event must supply the remaining $F(r)\Delta(r)$.
Thus, unbiasedness for every demand is equivalent to
\begin{equation}
\label{rn:eq:unbiased-identity}
\int_{[0,r]}W(u)Q(\mathrm{d}u)=F(r)\Delta(r),
\qquad \forall 0\le r\le M-x.
\end{equation}
The same correction $W$, chosen without knowing $d$, must satisfy this identity for every possible shortage $r$.
To ensure \cref{ass:dependence} holds, we choose $Q$ and $W$ using only the observed history $\mathcal F_{t-1}$, then sample $\tilde u=F^{-1}(v_t^u)$ using a fresh independent uniform seed $v_t^u$.

However, not every distribution $Q$ admits such a correction. 
The next theorem shows that an unbiased correction exists exactly when a curvature measure defined below is absolutely continuous with respect to $Q$. The correction $W$ equals $\Delta$ plus the corresponding Radon--Nikodym derivative.
Since $\psi$ is convex, the marginal shortage cost $m$ is nondecreasing.
We describe its increases using the finite nonnegative Lebesgue--Stieltjes \emph{curvature measure} $\nu$ induced by $m$, which is specified by $\nu(\{0\})\coloneqq0$ and $\nu((a,b])\coloneqq m(b)-m(a)$ for $0\le a<b\le M$, so $\Delta(r)=\nu((0,r])$.
The curvature measure $\nu$  records both jumps and continuous increases in marginal shortage cost.
Write $F(u^-)\coloneqq Q([0,u))$ for $u>0$ and set $F(0^-)\coloneqq0$.
In the theorem, $\ll$ denotes absolute continuity of measures, and the derivative in \cref{rn:eq:rn-correction} is a Radon--Nikodym derivative.

\begin{theorem}
\label{rn:thm:estimator}
Fix $x\in[0,M]$ and a probability distribution $Q$ supported on $[0,M-x]$.
There exists a correction $W\in L^1(Q)$ for which the estimator in Eq.~\eqref{rn:eq:gradient-estimator} is unbiased for $g(x;d)$ for every fixed demand $d\in[0,M]$ if and only if $F(\cdot^-)\nu\ll Q$ on $(0,M-x]$.
When this condition holds, the correction is unique $Q$-almost surely and is given on $(0,M-x]$ by
\begin{equation}
\label{rn:eq:rn-correction}
W_Q\coloneqq\Delta+\frac{\mathrm{d}\{F(\cdot^-)\nu\}}{\mathrm{d}Q},
\end{equation}
with $W_Q(0)\coloneqq0$.
Write $\tilde g_Q\coloneqq\tilde g_{W_Q}$.
Then $\mathbb E[\tilde g_Q(x,\tilde u,\tilde s)\mid x,d]=g(x;d)$ for every fixed demand.
If, in addition, $W_Q\in L^2(Q)$, the conditional variance of $\tilde g_Q$ is nondecreasing in $(d-x)_+$, and its worst-case value is
\begin{equation}
\label{rn:eq:variance}
\sup_{d\in[0,M]}\operatorname{Var}(\tilde g_Q(x,\tilde u,\tilde s)\mid x,d)=V_x(Q)\coloneqq\int_{(0,M-x]}W_Q(u)^2Q(\mathrm{d}u)-\Delta(M-x)^2.
\end{equation}
\end{theorem}

\smallskip

The proof is given in Appendix~\ref{rn:app:estimator}.
The absolute-continuity condition requires $\int_BF(u^-)\nu(\mathrm{d}u)=0$ whenever $Q(B)=0$, for every Borel set $B\subseteq(0,M-x]$.
In particular, if marginal shortage cost jumps at $u$ and $Q$ assigns positive probability below $u$, then $Q$ must also assign positive probability to $u$.

We call a randomization distribution $Q$ \emph{admissible} if it satisfies the condition in Theorem~\ref{rn:thm:estimator} and $W_Q\in L^2(Q)$, and denote the class of such distributions on $[0,M-x]$ by $\mathcal Q_x$.
This class is nonempty: the distribution \(Q=\delta_{M-x}\), where \(\delta_a\) denotes the point mass at \(a\), is admissible and implements the physical order \(M\) with probability one.
For this choice, $W_Q(M-x)=\Delta(M-x)$ and the conditional variance is zero.

\begin{algorithm}[htb]
\caption{Online stochastic gradient descent with randomized ordering}
\label{rn:alg:randomized-osgd}
\begin{algorithmic}[1]
\Require A deterministic initial virtual order $\tilde x_1\in[0,M]$ and a deterministic, positive, nonincreasing step-size sequence $(\eta_t)_{t\ge1}$.
\For{$t=1,2,\ldots$}
\State Choose an $\mathcal F_{t-1}$-measurable randomization distribution $Q_t\in\mathcal Q_{\tilde x_t}$.
\State Sample additional inventory $\tilde u_t \sim Q_t$ using a fresh independent random seed $v_t^u$.
\State Place the physical order $\tilde q_t\coloneqq\tilde x_t+\tilde u_t$ and observe sales $\tilde s_t\coloneqq\min\{\tilde d_t,\tilde q_t\}$.
\State Compute $\tilde g_t\coloneqq\tilde g_{Q_t}(\tilde x_t,\tilde u_t,\tilde s_t)$ using Eq.~\eqref{rn:eq:gradient-estimator} with $W=W_{Q_t}$ from Eq.~\eqref{rn:eq:rn-correction}.
\State Update $\tilde x_{t+1}\coloneqq\operatorname{Proj}_{[0,M]}\bigl(\tilde x_t-\eta_t\tilde g_t\bigr)$.
\EndFor
\end{algorithmic}
\end{algorithm}

We now use these estimators in the sequential ordering policy specified in \cref{rn:alg:randomized-osgd},
which ensures that \cref{ass:dependence} holds. In particular, $\tilde u_t$ is independent of $\tilde d_t$ conditional on the history $\mathcal{H}_{t-1}$.
\cref{rn:thm:estimator} therefore gives $\mathbb E[\tilde g_t\mid\mathcal H_{t-1},\tilde d_t]=g(\tilde x_t;\tilde d_t)$ and $\operatorname{Var}(\tilde g_t\mid\mathcal H_{t-1},\tilde d_t)\le V_{\tilde x_t}(Q_t)$ almost surely.
The distribution $Q_t$ determines both the estimator's worst-case conditional variance and expected additional inventory for period $t$.
In \cref{rn:sec:design} we describe how to select $Q_t$ to minimize the weighted sum of these quantities that appears in the regret bound.

%% file: _vcurrent/polished_section_2.tex
\section{{Optimizing} randomization distributions}
\label{rn:sec:design}

\subsection{Regret bound and distribution selection}
\label{rn:sec:local-objective}

\cref{rn:alg:randomized-osgd} permits any $\cF_{t-1}$-measurable admissible distribution $Q_t \in \mathcal{Q}_{\tilde{x}_t}$ of additional inventory in each period.
The additional inventory $\tilde u_t$ is added to the virtual order $\tilde x_t$, giving the physical order $\tilde q_t=\tilde x_t+\tilde u_t$.
Choosing $Q_t$ therefore affects both the variance of the subgradient estimator and the inventory actually ordered.
We hence choose $Q_t$ to minimize their combined contribution to the regret bound.

To account for the physical order, we compare its loss with the loss at the virtual order.
Adding $u$ units gives $\ell(x+u,d)-\ell(x,d)\le hu$ for $x,d\in[0,M]$ and $0\le u\le M-x$, because the excess-inventory cost increases by at most $hu$ while the shortage {cost} cannot increase.
{Let $G:=\max\{h,m(M)\}$, so the exact subgradient satisfies $|g(x;d)|\le G$ for every $x,d\in[0,M]$.}
{This bound applies to the exact subgradient; the subgradient estimates need not be bounded by $G$.}
Combining {the loss-increase} bound with the projected-gradient analysis at the virtual orders gives the following result.

\begin{proposition}
\label{rn:prop:physical-regret}
Let the step sizes be deterministic, positive, and nonincreasing.
{Under the demand conditions in Assumption~\ref{ass:dependence} and the sampling rule in Algorithm~\ref{rn:alg:randomized-osgd}, the policy's expected regret satisfies the following bound for every admissible sequence of randomization distributions:}
\begin{equation}
\label{rn:eq:physical-regret}
\mathbb E[\tilde r_T]\le{\frac{3M^2}{4\eta_T}}
+\frac{G^2}{2}\sum_{t=1}^T\eta_t
+{\frac32}\mathbb E\!\Big[\sum_{t=1}^T\Big\{\frac{\eta_t}{2}V_{\tilde x_t}(Q_t)+h\int u\,Q_t(\mathrm{d}u)\Big\}\Big].
\end{equation}
{Both expectations are taken over the policy's randomization and any randomness in demand.}
\end{proposition}

The proof is given in Appendix~\ref{rn:app:physical-regret}.
It combines a standard regret bound for the virtual orders $\tilde{x}_t$ with a martingale difference argument to account for stochastic gradients. These contribute the first two terms as well as the first term of the final sum. The second term in the final sum is the contribution from using the physical order $\tilde{q}_t$ instead of $\tilde{x}_t$.

To choose the distribution $Q_t$ for the current period, condition on the observed history $\cF_{t-1}$ and denote the current virtual order and step size by $x\in[0,M]$ and $\eta>0$.
Although $Q_t$ affects subsequent virtual orders, we select it by minimizing {only} its current-period contribution to the regret bound.
This results in the following optimization problem: for an admissible distribution $Q\in\mathcal Q_x$, define the one-period objective and its infimum by
\begin{equation}
\label{rn:eq:local-objective}
J_x(Q;\eta):=\frac\eta2V_x(Q)+h\int u\,Q(\mathrm{d}u),
\qquad J_x^\star(\eta):=\inf_{Q\in\mathcal Q_x}J_x(Q;\eta).
\end{equation}

To solve this problem, we first identify where $Q$ needs to place its support to minimize $J_x$.
Recall that the curvature measure $\nu$ records increases in marginal shortage cost, and its support identifies points around which these increases occur.
The next lemma shows that it is sufficient to consider \(Q\) supported on \(\{0\}\cup\operatorname{supp}(\nu|_{(0,M-x]})\).
Here $\nu|_{(0,M-x]}$ denotes the restriction of $\nu$ to $(0,M-x]$, and $\operatorname{supp}(\cdot)$ denotes the closed support of a measure.

\begin{lemma}\label{rn:lem:support}
Fix $x\in[0,M]$, and set $K_x:=\{0\}\cup\operatorname{supp}(\nu|_{(0,M-x]})$.
For every $Q\in\mathcal Q_x$, there is a distribution $\bar Q\in\mathcal Q_x$ supported on $K_x$ such that $V_x(\bar Q)\le V_x(Q)$ and $\int u\,\bar Q(\mathrm{d}u)\le\int u\,Q(\mathrm{d}u)$.
The second inequality is strict whenever $Q([0,M-x]\setminus K_x)>0$.
\end{lemma}

The proof in Appendix~\ref{rn:app:support} moves the probability assigned to each additional inventory $u$ to the largest point of $K_x$ not exceeding $u$ and averages the corrections for quantities moved to the same point.
This reassignment cannot increase expected additional inventory, and conditional averaging cannot increase the correction's second moment.
Because $h>0$, Lemma~\ref{rn:lem:support} implies that every optimizer of the one-period objective in Eq.~\eqref{rn:eq:local-objective} is supported on $K_x$.

\subsection{Piecewise-linear shortage {costs}}
\label{rn:sec:pwl}

Fix a virtual order $x\in[0,M]$ and step size $\eta>0$.
For a convex piecewise-linear {shortage cost} $\psi$, let $0<u_1<\cdots<u_J\le M-x$ be its breakpoints in the feasible range of additional inventory, and let $a_j>0$ be the jump in marginal shortage cost at $u_j$.
The curvature measure on this range is $\nu|_{(0,M-x]}=\sum_{j=1}^Ja_j\delta_{u_j}$.
Write $u_0:=0$ and $\Delta_j:=\sum_{i=1}^ja_i=\Delta(u_j)$ for $j=1,\ldots,J$, with $\Delta_0:=0$.
If $J=0$, then $\delta_0$ is optimal with objective value zero, so no additional inventory is needed.
Hence, we assume below that $J\ge1$.

By Lemma~\ref{rn:lem:support}, it suffices to choose probabilities on $\{0,u_1,\ldots,u_J\}$.
Selecting the additional inventory $u_j$ means placing the physical order $x+u_j$.
For $j=0,\ldots,J$, let $F_j:=Q([0,u_j])$ be the probability of adding at most $u_j$ units.
The probability of adding exactly $u_j$ units is {$p_j:=Q(\{u_j\})$, with $p_0=F_0$ and $p_j=F_j-F_{j-1}$ for $j=1,\ldots,J$}.
Thus $F_0=Q(\{0\})$ is the probability of placing the virtual order without additional inventory, and $F_J=1$.

Admissibility restricts how these probabilities can be assigned.
At a breakpoint with $p_j>0$, Eq.~\eqref{rn:eq:rn-correction} gives $W_Q(u_j)=\Delta_j+a_jF_{j-1}/p_j$.
If $p_j=0$, the condition $F(\cdot^-)\nu\ll Q$ requires $a_jF_{j-1}=0$, or equivalently $F_{j-1}=0$.
Conversely, imposing this condition at every breakpoint with zero probability ensures admissibility on the finite support.
Thus an initial set of breakpoints may receive zero probability, but once positive probability is assigned below a breakpoint, that breakpoint must also receive positive probability.

In these variables, the variance is $V_x(Q)=\sum_{j=1}^Ja_j^2F_{j-1}F_j/p_j$, with each term satisfying $F_{j-1}=p_j=0$ interpreted as zero; the expansion is given in Appendix~\ref{rn:app:pwl}.
Using $F_{j-1}F_j/p_j=F_{j-1}+F_{j-1}^2/(F_j-F_{j-1})$ and $\mathbb E_Q[\tilde u]=\sum_{j=1}^Ju_jp_j$, the optimization problem in Eq.~\eqref{rn:eq:local-objective} becomes the finite convex program
\begin{equation}
\label{rn:eq:pwl-program}
\min_{0\le F_0\le\cdots\le F_J=1}
\left\{\frac\eta2\sum_{j=1}^Ja_j^2\left(F_{j-1}+\frac{F_{j-1}^2}{F_j-F_{j-1}}\right)
+h\sum_{j=1}^Ju_j(F_j-F_{j-1})\right\}.
\end{equation}
We interpret the quotient as zero when $F_{j-1}=F_j=0$ and as $+\infty$ when $F_{j-1}=F_j>0$.
This is the lower-semicontinuous extension of the quadratic-over-linear term; it preserves convexity and excludes inadmissible probability choices.
The next theorem solves the problem in Eq.~\eqref{rn:eq:pwl-program} in two passes through the ordered breakpoints: a forward pass computes scalar quantities, and a backward pass recovers the cumulative probabilities.
{In Sections~\ref{rn:sec:pwl} and~\ref{rn:sec:continuous}}, we suppress the dependence of optimal distributions and their cumulative distribution functions on $\eta$.

\begin{theorem}
\label{rn:thm:pwl}
For $j=1,\ldots,J$, define the forward recursion as follows.
\begin{equation}
\label{rn:eq:pwl-recursion}
\begin{aligned}
z_0:=0,\;\;
z_j^-:=z_{j-1}+h\cdot(u_j-u_{j-1}),\;\;
z_j:=\bigl(\sqrt{z_j^-}-\sqrt{\eta/2}\,a_j\bigr)_+^2,\;\;
\rho_j:=\bigl(1-\sqrt{\eta/(2z_j^-)}\,a_j\bigr)_+.
\end{aligned}
\end{equation}
Starting from $F_J^\star:=1$, define $F_{j-1}^\star:=\rho_jF_j^\star$ for $j=J,\ldots,1$, and let $Q_x^\star$ assign mass $F_0^\star$ to zero and mass $p_j^\star:=F_j^\star-F_{j-1}^\star$ to $u_j$.
Then $(F_0^\star,\ldots,F_J^\star)$ uniquely solves Eq.~\eqref{rn:eq:pwl-program}, and the associated distribution $Q_x^\star$ uniquely minimizes $J_x(Q;\eta)$ over $\mathcal Q_x$, with $J_x^\star(\eta)=hu_J-z_J$.
Whenever $p_j^\star>0$, its correction is $W_{Q_x^\star}(u_j)=\Delta_{j-1}+a_j/(1-\rho_j)=\Delta_{j-1}+\max\{a_j,\sqrt{2z_j^-/\eta}\}$.
The forward and backward passes require $O(J)$ arithmetic operations.
\end{theorem}

The proof is given in Appendix~\ref{rn:app:pwl}.
Since $z_j^->0$ and $0\le\rho_j<1$ for every $j$, the backward pass produces nondecreasing cumulative probabilities.
To implement the rule at a given virtual order and step size, retain the breakpoints of $\psi$ in $(0,M-x]$ and execute the two passes.
The $O(J)$ operation count applies once the breakpoints are ordered and their marginal-cost jumps are available.

For a single breakpoint $u\in(0,M-x]$ with marginal-cost jump $a>0$, let $p\in(0,1]$ be the probability of adding $u$ units to the virtual order, with the remaining probability assigned to no additional inventory.
When $d\ge x+u$, the subgradient estimator equals $-m(0)-a/p$ with probability $p$ and $-m(0)$ otherwise, giving the worst-case conditional variance $V_x(Q)=a^2(1-p)/p$.
The one-period objective is therefore $J_x(Q;\eta)=(\eta a^2/2)(1-p)/p+hup$, with minimizer $p^\star=\min\{1,a\sqrt{\eta/(2hu)}\}$.

Holding the other parameters fixed, the optimal probability is nondecreasing in the marginal-cost jump $a$.
For a fixed $p<1$, the estimator variance grows quadratically with $a$, so a larger jump increases the benefit of ordering $x+u$ more frequently to reduce this variance.
By contrast, a larger additional inventory $u$ increases the term $hup$, which bounds the expected loss increase from additional inventory, without changing the worst-case variance at a fixed $p$.
The optimal probability is therefore nonincreasing in $u$, balancing the benefit of more accurate subgradient estimates against the {bound on the expected loss increase from additional} inventory.

With $a$, $u$, and $h$ fixed, the optimal probability is also nonincreasing as the step size $\eta$ decreases, and decreases strictly once it is below one.
A smaller step size reduces the weight $\eta/2$ on estimator variance in the regret bound, while leaving the additional-inventory term unchanged for a fixed $p$.
Consequently, reducing variance becomes less valuable relative to reducing additional inventory, and the optimal rule orders the additional inventory less frequently, accepting higher estimator variance.

\subsection{Absolutely continuous marginal shortage costs}
\label{rn:sec:continuous}

We now solve the same distribution-selection problem when marginal shortage cost changes continuously rather than through discrete jumps.
Specifically, assume that $m$ is absolutely continuous with $m'\in L^\infty([0,M])$ nonnegative almost everywhere.
The optimization still ranges over all admissible distributions, including those that assign positive probability to zero additional inventory.
In this case, we construct the optimal distribution from an auxiliary function $z_\eta$ defined by a scalar ordinary differential equation (ODE).
Its argument $u$ denotes an additional-inventory level, not a time period.
For $\eta>0$, consider the ODE
\begin{equation}
\label{rn:eq:ode}
z_\eta'(u)=h-\sqrt{2\eta}\,m'(u)\sqrt{z_\eta(u)},\qquad z_\eta(0)=0.
\end{equation}
The differential equation is required to hold almost everywhere on $[0,M]$.

\begin{lemma}
\label{rn:lem:ode}
Suppose that $m$ is absolutely continuous with $m'\in L^\infty([0,M])$ nonnegative almost everywhere.
For every $\eta>0$, Eq.~\eqref{rn:eq:ode} has a unique nonnegative absolutely continuous solution on $[0,M]$.
It satisfies $0<z_\eta(u)\le hu$ for $u>0$, and $u\mapsto m'(u)/\sqrt{z_\eta(u)}$ is integrable on $[0,M]$.
\end{lemma}

The proof is given in Appendix~\ref{rn:app:continuous}.
These properties allow us to recover the optimal cumulative distribution function from $z_\eta$.

\begin{theorem}
\label{rn:thm:continuous}
Fix $x\in[0,M)$ and $\eta>0$.
Under the assumptions of this subsection, the unique minimizer $Q_x^\star$ of Eq.~\eqref{rn:eq:local-objective} has cumulative distribution function
\begin{equation}
\label{rn:eq:continuous-cdf}
F_x^\star(u):=\exp\left\{-\int_u^{M-x}\sqrt{\frac\eta2}\frac{m'(v)}{\sqrt{z_\eta(v)}}\,\mathrm{d}v\right\},\qquad 0\le u\le M-x.
\end{equation}
It has an atom of size $F_x^\star(0)$ at zero and an absolutely continuous component on $(0,M-x]$.
Choose $W_{Q_x^\star}(0):=0$; for $0<u\le M-x$, its correction is $W_{Q_x^\star}(u)=\Delta(u)+\sqrt{2z_\eta(u)/\eta}$.
The optimal value is $J_x^\star(\eta)=h\cdot(M-x)-z_\eta(M-x)$.
For $x=M$, the unique feasible distribution is $\delta_0$ and the same value formula holds.
\end{theorem}

The proof in Appendix~\ref{rn:app:continuous} obtains a lower bound on the one-period objective in Eq.~\eqref{rn:eq:local-objective} by completing the square.
An admissible distribution attains this bound precisely when $W_Q(u)-\Delta(u)=\sqrt{2z_\eta(u)/\eta}$ holds $Q$-almost surely.
Together with Eq.~\eqref{rn:eq:rn-correction} and the normalization $F(M-x)=1$, this condition determines the optimal cumulative distribution function in Eq.~\eqref{rn:eq:continuous-cdf}.
By Lemma~\ref{rn:lem:ode}, the integral defining $F_x^\star(0)$ is finite, so the optimal distribution assigns positive probability to zero additional inventory.
For a fixed step size $\eta$, the ODE needs to be solved only once on $[0,M]$, since changing the virtual order $x$ changes only the upper integration limit $M-x$ in the cumulative distribution function.

Equation~\eqref{rn:eq:ode} can be approximated numerically using a backward Euler scheme \citep{HairerWanner1996}; Appendix~\ref{rn:app:continuous} provides the construction and error analysis.
For fixed $\eta$ and problem parameters, this scheme approximates $z_\eta$ uniformly on $[0,M]$ to within $\epsilon$ using $O(\epsilon^{-1})$ grid intervals.
Substituting the resulting approximation into the distribution  and correction formulas in Theorem~\ref{rn:thm:continuous}, while retaining the original shortage cost, yields an admissible distribution whose one-period objective value is at most $\epsilon$ above the optimum.
This objective error can be easily included in the allowance $\epsilon_t$ in the subsequent regret analysis.

For some shortage costs, the ODE solution itself has a simpler representation, allowing the distribution to be evaluated without approximation.
For a quadratic shortage cost, the construction reduces to a single scalar equation, as the following example shows.

\begin{example}
\label{rn:ex:quadratic-design}
Suppose $\psi(r)=\beta r+ \gamma  r^2/2$, where $\beta\ge0$ and $ \gamma >0$.
Then $m(r)=\beta+ \gamma  r$, $\Delta(r)= \gamma  r$, and $m'(r)= \gamma $.
For $u\ge0$, let $w_\eta(u)\in[0,1)$ be the unique solution of $-w_\eta(u)-\log(1-w_\eta(u))=\eta \gamma ^2u/h$.
A direct differentiation shows that $z_\eta(u):=h^2w_\eta(u)^2/(2\eta \gamma ^2)$ solves Eq.~\eqref{rn:eq:ode}.
Therefore, Theorem~\ref{rn:thm:continuous} gives, for $0\le u\le M-x$,
\begingroup
\[
F_x^\star(u)=\frac{1-w_\eta(M-x)}{1-w_\eta(u)},\quad
W_{Q_x^\star}(u)= \gamma  u+\frac h{\eta \gamma }w_\eta(u),\quad
J_x^\star(\eta)=h(M-x)-\frac{h^2}{2\eta \gamma ^2}w_\eta(M-x)^2.
\]
\endgroup
Since $w_\eta(0)=0$, the probability of ordering without additional inventory is $F_x^\star(0)=1-w_\eta(M-x)$.
\end{example}

%% file: _vcurrent/polished_section_3.tex
\section{Regret guarantees and minimax optimality}
\label{rn:sec:regret}

Section~\ref{rn:sec:design} selects the distribution of additional inventory for a given virtual order and step size.
We now use the resulting one-period bounds to choose step sizes and control cumulative regret.
Under Assumption~\ref{rn:ass:penalty}, the policy achieves $O(T^{2/3})$ expected regret for adaptive demand and its special cases such as stationary or nonstationary independent demand.
Under independent demand, the rate improves to $O(\sqrt T)$ when the expected losses are uniformly strongly convex, or when they share a minimizer and satisfy a uniform restricted secant inequality around it.
Matching lower bounds establish that neither rate can be improved over its corresponding class with a fixed marginal-cost bound, even under iid demand.
These lower bounds apply to all nonanticipating policies that observe only orders and sales, not just Algorithm~\ref{rn:alg:randomized-osgd}.
We then give sufficient conditions for the faster rate and extend the analysis to nonperishable inventory.

\subsection{Distribution-free regret and the general minimax rate}
\label{rn:sec:general-regret}

The distributions in Section~\ref{rn:sec:design} depend on the shortage {cost}, the virtual order, and the step size, but their one-period objective values can be bounded in a common way.
To express this bound, define $B_{\nu,x}:=\int_{(0,M-x]}\sqrt u\,\nu(\mathrm{d}u)$ and $B_\nu:=\int_{(0,M]}\sqrt u\,\nu(\mathrm{d}u)$.
The square-root weight $\sqrt{u}$ arises from balancing estimator variance against the {bound on the} expected {loss increase from} additional inventory in Eq.~\eqref{rn:eq:local-objective}.
Both quantities are finite under Assumption~\ref{rn:ass:penalty}, and we call $B_\nu$ the curvature index.
They weight increases in marginal shortage cost by the square root of the additional inventory level needed to reach each shortage threshold.
The first quantity uses the feasible range of additional inventory at $x$, while the second gives a bound common to all virtual orders.
The next theorem converts this one-period bound into cumulative regret and allows {a nonnegative} additive {allowance above} the {analytic one-period bound} in each period.

\begin{theorem}
\label{rn:thm:general-regret}
Under Assumption~\ref{rn:ass:penalty}, for every $x\in[0,M]$ and $\eta>0$,
\begin{equation}
\label{rn:eq:local-bound}
J_x^\star(\eta)\le\sqrt{2h\eta}\,B_{\nu,x}\le\sqrt{2h\eta}\,B_\nu.
\end{equation}
Let $(\eta_t)_{t\ge1}$ be deterministic, positive, and nonincreasing, and let $(\epsilon_t)_{t\ge1}$ be deterministic and nonnegative.
In Algorithm~\ref{rn:alg:randomized-osgd}, suppose that the distributions $Q_t\in\mathcal Q_{\tilde x_t}$ are $\cF_{t-1}$-measurable and satisfy $J_{\tilde x_t}(Q_t;\eta_t)\le\sqrt{2h\eta_t}\,B_{\nu,\tilde x_t}+\epsilon_t$ almost surely in every period.
Let $(\tilde d_t)$ be any demand process supported on $[0,M]$, and suppose that $\tilde{u}_t \sim Q_t$ are sampled as described in \cref{rn:sec:admissible} so that \cref{ass:dependence} holds.
Then the expected regret of Algorithm~\ref{rn:alg:randomized-osgd}, relative to the best fixed order evaluated on the realized demand sequence, satisfies
\begin{equation}
\label{rn:eq:general-regret}
\mathbb E[\tilde r_T]\le\frac{3M^2}{4\eta_T}+\frac{G^2}{2}\sum_{t=1}^T\eta_t
+\frac32\sqrt{2h}\,B_\nu\sum_{t=1}^T\sqrt{\eta_t}+\frac32\sum_{t=1}^T\epsilon_t.
\end{equation}
The expectation is taken over the policy’s randomization and any randomness in demand.
In particular, $\eta_t=\eta_0t^{-2/3}$ with any $\eta_0>0$ gives expected regret $O(T^{2/3})$ whenever $\sum_{t=1}^T\epsilon_t=O(T^{2/3})$.
\end{theorem}

The proof is given in Appendix~\ref{rn:app:general-regret} and is divided into two steps.
Lemma~\ref{rn:lem:local-bound} first constructs an admissible distribution for every finite curvature measure, including measures with both atomic and nonatomic parts, and establishes Eq.~\eqref{rn:eq:local-bound}.
Substituting the one-period bound into Proposition~\ref{rn:prop:physical-regret} gives Eq.~\eqref{rn:eq:general-regret}.
The allowance $\epsilon_t$ bounds any excess of the one-period objective over the analytic bound in Eq.~\eqref{rn:eq:local-bound}.
An $\epsilon_t$-optimal solution of Eq.~\eqref{rn:eq:local-objective} satisfies this requirement, but $\epsilon_t$ need not equal the gap to the optimum.
It can also account for the objective error introduced by numerically approximating Eq.~\eqref{rn:eq:ode}, as discussed above.

The step-size schedule $\eta_t=\eta_0t^{-2/3}$ balances the terms $1/\eta_T$ and $\sum_{t=1}^T\sqrt{\eta_t}$, both of which have order $T^{2/3}$, while $\sum_{t=1}^T\eta_t$ has order $T^{1/3}$.
This schedule does not require advance knowledge of $T$.
The guarantee does not require demand to be independent or stationary across periods.
It applies to fixed demand sequences and adaptive demand whose conditional distribution may depend on the full past, provided current demand and the additional-inventory draw are conditionally independent given that past.

The same bound shows that the amount of additional inventory used for learning declines on average.
{The term $h\mathbb E[\tilde u_t\mid\mathcal F_{t-1}]$ bounds the expected loss increase from additional inventory}.
Since both terms in $J_x(Q;\eta)$ are nonnegative, $h\mathbb E[\tilde u_t\mid\mathcal F_{t-1}]\le J_{\tilde x_t}(Q_t;\eta_t)$, and hence
\[
\mathbb E\!\Big[\sum_{t=1}^T\tilde u_t\Big]
\le\sqrt{\frac2h}\,B_\nu\sum_{t=1}^T\sqrt{\eta_t}+\frac1h\sum_{t=1}^T\epsilon_t.
\]
Under the $t^{-2/3}$ schedule and $\sum_{t=1}^T\epsilon_t=O(T^{2/3})$, the expected average additional inventory is $O(T^{-1/3})$.
Since additional inventory is nonnegative, its time average therefore converges to zero in $L^1$.
If also $\epsilon_t\to0$, the one-period bound gives $\tilde u_t\to0$ in $L^1$.
These statements concern the additional inventory added to the virtual order, not the probability of adding any inventory.

When $\nu=0$, the shortage {cost} is linear and the selected subgradient is observable without additional inventory.
Taking $Q_t=\delta_0$ in every period and $\eta_t=\eta_0t^{-1/2}$ in Proposition~\ref{rn:prop:physical-regret} then gives an $O(\sqrt T)$ expected regret bound for fixed or adaptive demand under the same conditions.

We next consider whether the $T^{2/3}$ rate can also be improved uniformly over general convex shortage {costs}.
To resolve this, it suffices to establish a matching lower bound under iid demand, a special case of the demand environments in \cref{ass:dependence} covered by the upper bound.
Fix a marginal-cost bound $\bar m\ge4h/3$ and let $\mathcal C$ be the class of pairs $(\psi,P)$ such that $\psi$ satisfies Assumption~\ref{rn:ass:penalty}, $m(M)\le\bar m$, and $P$ is a demand distribution on $[0,M]$.
Within each instance, demands are drawn independently from $P$, and the policy knows $\psi$ but not $P$.
For fixed $M,h,\bar m$, Theorem~\ref{rn:thm:general-regret} gives a uniform $O(T^{2/3})$ upper bound over $\mathcal C$.

We prove the lower bound for pseudo-regret, which compares the policy with the fixed order minimizing expected loss under $P$, rather than the best fixed order chosen after observing the demand sequence.
As discussed in Section~\ref{rn:sec:inventory}, expected hindsight regret is at least as large: for every policy $\pi$ and instance $(\psi,P)\in\mathcal C$,
\[
\mathbb E_{\psi,P}^{\pi}[\tilde r_T]
\ge
\mathbb E_{\psi,P}^{\pi}\!\Big[\sum_{t=1}^T\ell(\tilde q_t,\tilde d_t)\Big]
-\min_{q\in[0,M]}\mathbb E_P\!\Big[\sum_{t=1}^T\ell(q,\tilde d_t)\Big].
\]
A matching lower bound on the right-hand side therefore establishes rate optimality for the original regret criterion, as stated in the next theorem.

\begin{theorem}
\label{rn:thm:general-lower}
Fix $M,h>0$.
Under iid demand, for every $T\ge512$,
\begin{equation}
\label{rn:eq:general-lower}
\inf_\pi\sup_{(\psi,P)\in\mathcal C}\mathbb E_{\psi,P}^{\pi}[\tilde r_T]
\ge\frac{hM}{3456}\,T^{2/3},
\end{equation}
where the infimum is over all nonanticipating policies that know the shortage {cost} and observe only orders and sales.
There is a fixed piecewise-linear {shortage cost}, independent of $T$ and satisfying $m(M)=4h/3$, for which the same lower bound holds when the supremum is taken only over $P$.
\end{theorem}

The statement {for a fixed shortage cost} highlights the difficulty of distinguishing demand distributions from censored sales, even when the shortage {cost} is fixed and known.
We prove this result using a construction that is closely related to that of \citet{yangShi2023}.
However, their lower-bound argument for integer orders does not apply directly to the continuous-order extension of their instance: continuous orders can reveal the same demand information as their computed larger integer order while incurring arbitrarily small additional expected loss under one of the demand distributions.
Our construction rules out such low-cost information acquisition and establishes the $T^{2/3}$ lower bound over the full continuous decision interval. Furthermore, our construction can be adapted to certify rate-optimality in more structured settings; see \cref{rn:sec:accelerated}.

To establish this property, we begin with the normalized instance $M=3$ and $h=3/2$.
Define the piecewise-linear shortage {cost} $\psi(z):=\int_0^z m(u)\,du$, where
\begin{equation}\label{rn:eq:mz}
    m(z):=
\begin{cases}
1/2, & 0\le z<1,\\
1,   & 1\le z<2,\\
2,   & 2\le z\le 3.
\end{cases}
\end{equation}
For $0<\varepsilon\le1/8$ and $\sigma\in\{+1,-1\}$, consider the demand distribution
\[
P_\sigma(\{0\})=\frac12,\qquad
P_\sigma(\{2\})=\frac14-\sigma\varepsilon,\qquad
P_\sigma(\{3\})=\frac14+\sigma\varepsilon.
\]
Relative to the reference distribution with masses $(1/2,1/4,1/4)$, $P_+$ transfers probability $\varepsilon$ from demand two to demand three, while $P_-$ transfers the same amount in the opposite direction.
Their expected losses are minimized at $q_+^\star=1$ and $q_-^\star=0$, respectively.
Yet every order $q\le 2$ produces the same sales distribution under both demand laws, because demands two and three both exhaust the available inventory.
Only orders above two distinguish the two instances, and each such order has expected loss bounded away from the minimum expected loss.

The policy therefore faces a choice between acquiring information at an operating cost and continuing to order without distinguishing the demand laws.
Obtaining sufficiently many informative observations incurs regret of order $\varepsilon^{-2}$; otherwise, choosing the wrong order region incurs regret of order $\varepsilon T$.
Balancing these terms with $\varepsilon=T^{-1/3}$ gives the $T^{2/3}$ lower bound.
Detailed arguments and the extension to arbitrary $M,h>0$ are provided in Appendix~\ref{rn:app:general-lower}.

\subsection{Faster regret under expected-loss curvature}
\label{rn:sec:accelerated}

The general lower bound does not rule out faster learning when the expected losses have additional structure.
In this subsection, suppose demands are independent, with $\tilde d_t\sim P_t$, where the distributions are fixed before the first period but need not be identical.
Define the expected loss $L_t(q):=\mathbb E_{P_t}[\ell(q,\tilde d_t)]$ and its selected subgradient $\bar g_t(q):=\mathbb E_{P_t}[g(q;\tilde d_t)]\in\partial L_t(q)$.
Let $q_T^\star\in\arg\min_{q\in[0,M]}\sum_{t=1}^TL_t(q)$ minimize cumulative expected loss.
This deterministic order is used only in the analysis; the regret benchmark remains the best fixed order on the realized demand sequence.
Under iid demand, we write $L_t=L$ and $\bar g_t=\bar g$.

{Under independent demand, the next theorem gives two alternative conditions on the expected losses under which Algorithm~\ref{rn:alg:randomized-osgd} achieves $O(\sqrt T)$ expected regret.}
The first requires each expected loss to be uniformly strongly convex.
The second requires the expected losses to share a minimizer and satisfy a restricted secant inequality around it.
This inequality ensures that, away from the common minimizer, the negative expected subgradient points toward it and has magnitude at least proportional to the distance.
Unlike strong convexity, this condition compares each order only with the common minimizer, rather than imposing a curvature bound between every pair of orders.
Under iid demand, the common-minimizer requirement is automatic, but the restricted secant inequality must still hold.

\begin{theorem}
\label{rn:thm:accelerated}
Under Assumption~\ref{rn:ass:penalty}, suppose demands are independent and, for a constant $\mu>0$, one of the following conditions holds.
\begin{enumerate}
\item[(i)] Each expected loss $L_t$ is $\mu$-strongly convex on $[0,M]$, so that $L_t(y)\ge L_t(x)+\bar g_t(x)(y-x)+(\mu/2)(y-x)^2$ for $x,y\in[0,M]$.
\item[(ii)] The expected losses $L_t,\, t=1, 2, \cdots, T$, have a common minimizer $q^\star\in[0,M]$, and their selected subgradients satisfy the restricted secant inequality
\begin{equation}
\label{rn:eq:rsi}
\bar g_t(q)(q-q^\star)\ge\mu(q-q^\star)^2,\qquad q\in[0,M],\quad t\in[T].
\end{equation}
\end{enumerate}
Set $a=1$ in case (i) and $a=1/2$ in case (ii).
Fix $t_0\ge1$, choose $\eta_t:=1/\{\mu(t+t_0)\}$, and in each period of Algorithm~\ref{rn:alg:randomized-osgd} choose $Q_t\in\mathcal Q_{\tilde x_t}$ satisfying $J_{\tilde x_t}(Q_t;\eta_t)\le\sqrt{2h\eta_t}\,B_{\nu,\tilde x_t}+\epsilon_t$ almost surely, where $\epsilon_t\ge0$ is deterministic.
Then
\begin{equation}
\label{rn:eq:accelerated-regret}
\begin{aligned}
\mathbb E[\tilde r_T]\le{}&\frac{\mu t_0M^2}{2a}+\frac{G^2}{2a\mu}\log\left(1+\frac T{t_0}\right) +\frac{2\sqrt{2h}\,B_\nu}{a\sqrt\mu}\bigl(\sqrt{T+t_0}-\sqrt{t_0}\bigr)
+\frac{\pi G^2}{\mu}+\frac1a\sum_{t=1}^T\epsilon_t.
\end{aligned}
\end{equation}
The expectation is over demand and the policy's randomization.
In particular, the expected regret is $O(\sqrt T)$ whenever $\sum_{t=1}^T\epsilon_t=O(\sqrt T)$.
\end{theorem}

To see where the improvement comes from, first compare the policy with the order $q_T^\star$ minimizing cumulative expected loss.
Both conditions strengthen the usual subgradient inequality to
\[
a\{L_t(x)-L_t(q_T^\star)\} \le
\bar g_t(x)(x-q_T^\star) - \frac{\mu}{2}(x-q_T^\star)^2.
\]
In case~(i), this follows directly from strong convexity.
In case~(ii), it follows from taking $q_T^\star=q^\star$ and averaging the convexity bound with $0\le\bar g_t(x)(x-q^\star)-\mu(x-q^\star)^2$, which follows from Eq.~\eqref{rn:eq:rsi}.
The negative squared-distance term makes the comparison tighter whenever the virtual order differs from $q_T^\star$.

With this comparison established, we follow the standard projected stochastic subgradient argument, using the squared-distance analysis for strongly convex online losses \citep{HazanAgarwalKale2007}.
Writing {$D_t:=\mathbb E[(\tilde x_t-q_T^\star)^2]$}, we apply the additional-inventory bound $L_t(\tilde q_t)-L_t(\tilde x_t)\le h\tilde u_t$ to obtain
\[
a\mathbb E[L_t(\tilde q_t)-L_t(q_T^\star)]
\le\frac{D_t-D_{t+1}}{2\eta_t}-\frac\mu2D_t
+\frac{\eta_tG^2}{2}+\mathbb E[J_{\tilde x_t}(Q_t;\eta_t)].
\]
With $\eta_t=1/\{\mu(t+t_0)\}$, we have $\eta_t^{-1}-\eta_{t-1}^{-1}=\mu$ for $t\ge2$, so the coefficients of {$D_2,\ldots,D_T$} cancel when these terms are summed.
Their combined sum is {$(\mu t_0/2)D_1-D_{T+1}/(2\eta_T)\le\mu t_0M^2/2$}, leaving an upper bound that does not grow with $T$.
The cumulative contribution $\eta_tG^2/2$ is $(G^2/2)\sum_{t=1}^T\eta_t\le\{G^2/(2\mu)\}\log(1+T/t_0)=O(\log T)$.
Since it is assumed that $\mathbb E[J_{\tilde x_t}(Q_t;\eta_t)]\le\sqrt{2h}\,B_\nu\sqrt{\eta_t}+\epsilon_t$ and $\sum_{t=1}^T\sqrt{\eta_t}=O(\sqrt T)$, this term contributes at most $O(\sqrt T)+\sum_{t=1}^T\epsilon_t$ over the horizon.
Combining the three contributions and dividing by $a$ bounds cumulative expected loss relative to $q_T^\star$ by $O(\sqrt T)+(1/a)\sum_{t=1}^T\epsilon_t$.
The analysis also shows that estimator variance and additional inventory account for the $O(\sqrt T)$ term, while the remaining terms are bounded or grow logarithmically in $T$.

This argument bounds expected loss relative to $q_T^\star$, but the best fixed order selected after observing demand can incur a smaller cumulative loss.
Expected hindsight regret therefore also includes the advantage of selecting the comparator after demand is observed: \(\min_q\sum_{t=1}^T L_t(q) - \mathbb E [\min_q\sum_{t=1}^T \ell(q,\tilde d_t)]\).
Curvature makes departures from $q_T^\star$ costly in expected loss, while independence and bounded marginal losses limit the random fluctuations that can make another order preferable on a realized demand sequence.
Lemma~\ref{rn:lem:comparator} uses these properties to show that $0 \leq \min_q\sum_{t=1}^T L_t(q) - \mathbb E [\min_q\sum_{t=1}^T \ell(q,\tilde d_t)] \leq \pi G^2/\mu$, independently of $T$.
Adding this constant gives Eq.~\eqref{rn:eq:accelerated-regret}, so the faster rate holds for the original hindsight regret, not only for pseudo-regret.
The full argument is given in Appendix~\ref{rn:app:acceleration}.

We next show that the $\sqrt T$ dependence cannot be improved uniformly over either class in Theorem~\ref{rn:thm:accelerated}.
Fix $\bar m\ge\max\{2h,8\mu M\}$ and let $\mathcal C_\mu$ denote either the class of independent-demand instances $(\psi,P_1,\ldots,P_T)$ satisfying condition (i) or the class satisfying condition (ii), in each case under Assumption~\ref{rn:ass:penalty} and $m(M)\le\bar m$.
Policies know $\psi$ and the supplied lower bound $\mu$, but not the demand distributions.

\begin{theorem}
\label{rn:thm:accelerated-lower}
Fix $M,h,\mu>0$.
For either structural class $\mathcal C_\mu$ and every $T\ge4096$,
\begin{equation}
\label{rn:eq:quadratic-lower}
\inf_\pi\sup_{(\psi,P_1,\ldots,P_T)\in\mathcal C_\mu}\mathbb E^\pi[\tilde r_T]
\ge\frac h{6144}\min\left\{M,\frac h{4\mu}\right\}\sqrt T.
\end{equation}
The infimum is over all nonanticipating policies that observe only orders and sales.
In particular, the lower bound holds under iid demand for a fixed quadratic {shortage cost} independent of $T$, whose expected losses satisfy both conditions in Theorem~\ref{rn:thm:accelerated}.
\end{theorem}

The proof uses the same pair of demand distributions and statistical-testing argument as Theorem~\ref{rn:thm:general-lower}, with the shortage {cost} replaced by $\psi(z)=z^2/2$ in the normalized instance.
Obtaining enough informative observations still incurs regret of order $\varepsilon^{-2}$, but the loss from choosing the wrong order region is now of order $\varepsilon^2 T$ rather than $\varepsilon T$.
Balancing these terms gives the $\sqrt{T}$ lower bound.

For $0<\mu\le h/(4M)$, rescaling the construction and setting $\psi(z)=hz^2/M$ yields the lower bound $hM\sqrt{T}/6144$.
For larger $\mu$, the construction is translated to $[M-b,M]$, where $b=h/(4\mu)$, and uses $\psi(z)=hz^2/b$.
Orders below $M-b$ can be replaced by $M-b$ without increasing loss or removing information.
This reduction preserves the lower bound while ensuring the required strong-convexity constant over the full decision interval.
The details are given in Appendix~\ref{rn:app:quadratic-lower}.

\subsection{Sufficient conditions from shortage costs and demand}
\label{rn:sec:sufficient}

To apply the faster regret guarantee, we need conditions on the shortage {cost} and demand that ensure the required structure of the expected losses.
Strong convexity of $\psi$ alone is insufficient, because the realized loss is affine in the order quantity whenever $q>d$.
For example, if $P_t([0,b])=1$ for some $b<M$, then $L_t(q)=h\cdot\{q-\mathbb E[\tilde d_t]\}$ on $[b,M]$, so $L_t$ cannot be strongly convex on the entire interval $[0,M]$.
The next proposition gives sufficient conditions that are easy to verify.

\begin{proposition}
\label{rn:prop:sufficient}
Suppose demands are independent and $\psi$ satisfies Assumption~\ref{rn:ass:penalty}.
The following conditions imply the corresponding structure in Theorem~\ref{rn:thm:accelerated}.
\begin{enumerate}
\item[(i)] Suppose there is a constant $ \gamma >0$ such that $m(v)-m(u)\ge \gamma (v-u)$ for $0\le u\le v\le M$, and $\inf_tP_t(\{M\})\ge p_0>0$.
Then each $L_t$ is $ \gamma  p_0$-strongly convex on $[0,M]$.
\item[(ii)] Suppose the absolutely continuous component of each $P_t$ has a density $f_t$ satisfying $f_t(q)\ge f_0>0$ for almost every $q\in(0,M)$, uniformly in $t$.
Then each $L_t$ is $(h+m(0))f_0$-strongly convex on $[0,M]$.
The distributions may also have atoms or singular continuous components.
\item[(iii)] Suppose the slope condition in part (i) holds, and the expected losses share a minimizer $q^\star\in[0,M)$.
If there is a $b\in(q^\star,M]$ with {$\inf_tP_t([b,M])\ge p_0>0$}, then Eq.~\eqref{rn:eq:rsi} holds with {${\mu=\gamma p_0(b-q^\star)/(M-q^\star)}$}.
\end{enumerate}
\end{proposition}

Parts~(i) and~(ii) obtain strong convexity in different ways.
In part~(i), demand equals $M$ with probability at least $p_0$, contributing the strongly convex loss $\psi(M-q)$ over the full decision interval.
All remaining demand realizations contribute convex losses, so taking expectations retains a uniform strong-convexity bound.
Part~(ii) instead uses demand mass between order levels: for $x<y$, we have \({\bar g_t(y)-\bar g_t(x)\ge (h+m(0))P_t([x,y))}\).
The density lower bound makes the expected subgradient increase at least linearly with the order quantity, even when neither $\psi$ nor $f_t$ is smooth.

Part~(iii) does not require global strong convexity.
Demand above $b$ provides sufficient curvature on $[0,b]$, which contains the common minimizer, and monotonicity of the expected subgradient extends the restricted secant inequality beyond $b$.
The argument also permits the boundary minimizer $q^\star=0$ and does not require $\bar g_t(q^\star)=0$.
The proofs of all three parts are given in Appendix~\ref{rn:app:sufficient}.

The constants in the proposition also provide the lower bound on $\mu$ needed for the $1/t$ step-size schedule in Theorem~\ref{rn:thm:accelerated}.
Exact knowledge of the demand distributions or the common minimizer is unnecessary when suitable uniform bounds are available.
For example, a tail bound $\inf_tP_t([b,M])\ge p_0$ and a positive separation {$q^\star\le b-\zeta$ with $\zeta>0$} allow the choice {\(\mu=\frac{ \gamma  p_0\zeta}{M}\)}.
These bounds are supplied as properties of the assumed demand family, rather than estimated by the algorithm from censored sales.
The distinction is that the shortage {cost} determines which marginal-cost information censoring may hide, while the {shortage cost} and demand together determine whether the expected losses support the faster rate.

\subsection{Extension to nonperishable inventory}
\label{rn:sec:non-perishable}

We now allow unsold inventory to carry forward rather than being discarded at the end of each period.
Let $h_{\mathrm H}>0$ denote the per-unit holding cost, while the purchase cost $c$, selling price $p$, and shortage penalty $\varphi$ remain as defined in Section~2.1.
Let $\tilde y_t$ denote inventory carried into period $t$ and $\tilde q_t^{\mathrm{ord}}\ge0$ the quantity purchased before demand, so that $\tilde q_t=\tilde y_t+\tilde q_t^{\mathrm{ord}}$ is the total available inventory.
{Thus $\tilde q_t$ now denotes available inventory, while $\tilde q_t^{\mathrm{ord}}$ denotes purchases; the two coincide in the perishable setting}.
Sales remain $\tilde s_t=\min\{\tilde d_t,\tilde q_t\}$, with $\tilde d_t\in[0,M]$.
Starting from $\tilde y_1=0$, inventory evolves as $\tilde y_{t+1}=\tilde q_t-\tilde s_t=(\tilde q_t-\tilde d_t)_+$ for $t\in[T]$.
Since inventory cannot be discarded or returned, the feasible stock levels satisfy $\tilde y_t\le\tilde q_t\le M$.
Thus, previous decisions and demand now restrict the stock levels available in subsequent periods.

The period profit, net of the shortage penalty, is $\tilde\Pi_t^{\mathrm{NP}}\coloneqq p\tilde s_t-c\tilde q_t^{\mathrm{ord}}-h_{\mathrm H}\tilde y_{t+1}-\varphi((\tilde d_t-\tilde q_t)_+)$.
Define $\ell^{\mathrm{NP}}(q,d)\coloneqq h_{\mathrm H}(q-d)_++\psi((d-q)_+)$, where $\psi(r)=\varphi(r)+(p-c)r$ is unchanged.
Using the inventory-balance identity $\tilde q_t^{\mathrm{ord}}=\tilde s_t+\tilde y_{t+1}-\tilde y_t$ gives
\[
\sum_{t=1}^T\tilde\Pi_t^{\mathrm{NP}}
=
(p-c)\sum_{t=1}^T\tilde d_t
-
\Big[
\sum_{t=1}^T\ell^{\mathrm{NP}}(\tilde q_t,\tilde d_t)
+
c\tilde y_{T+1}
\Big].
\]
For two policies evaluated on the same realized demand sequence, the demand-only term cancels, so their cumulative profit difference is the negative of the difference between their bracketed losses.
The period loss retains the form of Eq.~\eqref{rn:eq:loss}, with $h_{\mathrm H}$ replacing $h=c+\kappa$.
With no terminal salvage credit, the additional term $c\tilde y_{T+1}$ accounts for the purchase cost of inventory remaining at the end of the horizon.

For comparison, we use the fixed base-stock policy class considered in \citet{huhRusmevichientong2009,hihatEtAl2023}.
Each comparator is evaluated on the same realized demand sequence as the learning policy.
This policy starts from empty inventory and replenishes its own inventory to a fixed level $q\in[0,M]$ before demand realization in every period.
Its cumulative {period} loss is $\sum_{t=1}^T\ell^{\mathrm{NP}}(q,\tilde d_t)$, and its terminal inventory is $(q-\tilde d_T)_+$.
Regret against the best fixed base-stock level in hindsight is therefore
\[
\tilde r_T^{\mathrm{NP}}
\coloneqq
\sum_{t=1}^T\ell^{\mathrm{NP}}(\tilde q_t,\tilde d_t)
+c\tilde y_{T+1}
-
\min_{q\in[0,M]}
\Big\{
\sum_{t=1}^T\ell^{\mathrm{NP}}(q,\tilde d_t)
+c(q-\tilde d_T)_+
\Big\}.
\]
Both policies' terminal inventories lie in $[0,M]$, so including the terminal terms changes regret by at most $cM$ and does not affect its rate.

Since the period loss {retains} the same form as {in} the {perishable} case, the main difficulty in applying Algorithm~\ref{rn:alg:randomized-osgd} comes from the inventory constraint.
In the perishable setting, the randomized target $\tilde x_t+\tilde u_t$ is implemented by purchasing that quantity at the start of the period.
With carryover $\tilde y_t$, reaching the same target $\tilde x_t+\tilde u_t$ is infeasible when $\tilde y_t>\tilde x_t+\tilde u_t$.
We therefore retain the randomized target but purchase only the amount needed to reach it, placing no order when carryover already exceeds it.
Specifically, choose $Q_t\in\mathcal Q_{\tilde x_t}$ using the observed history, draw $\tilde u_t$ according to the sampling rule in Algorithm~\ref{rn:alg:randomized-osgd}, and purchase $\tilde q_t^{\mathrm{ord}}=(\tilde x_t+\tilde u_t-\tilde y_t)_+$.
The resulting stock is $\tilde q_t=\max\{\tilde x_t+\tilde u_t,\tilde y_t\}$, which satisfies $\tilde y_t\le\tilde q_t\le M$.
When this stock exceeds the randomized target {$\tilde x_t+\tilde u_t$}, we must account for \emph{the difference both in the observations supplied to the subgradient estimator} and \emph{in the loss entering regret}.

First, the estimator requires sales censored at the randomized target, whereas actual sales are censored at the available stock.
Since $\tilde q_t\ge\tilde x_t+\tilde u_t$, we can recover the required observation by defining the synthetic sales
\(\tilde s_t^{\mathrm{syn}}
\coloneqq
\min\{\tilde x_t+\tilde u_t,\tilde s_t\}
=
\min\{\tilde x_t+\tilde u_t,\tilde d_t\}.\)
The first expression uses only observed quantities, and the equality shows that these are exactly the sales that would have been observed at the randomized target.
Let $\tilde g_Q^{\mathrm{NP}}$ denote the estimator in Theorem~\ref{rn:thm:estimator} with $h_{\mathrm H}$ in place of $h$.
Because $\psi$ is unchanged, the admissible distributions $\mathcal Q_x$, corrections $W_Q$, and variance bounds $V_x(Q)$ are unchanged.
We compute $\tilde g_t=\tilde g_{Q_t}^{\mathrm{NP}}(\tilde x_t,\tilde u_t,\tilde s_t^{\mathrm{syn}})$ and update $\tilde x_{t+1}=\operatorname{Proj}_{[0,M]}(\tilde x_t-\eta_t\tilde g_t)$.
By Theorem~\ref{rn:thm:estimator}, conditional on $\mathcal H_{t-1}$ and current demand $\tilde d_t$, this estimate is unbiased for the selected subgradient of $\ell^{\mathrm{NP}}(\tilde x_t,\tilde d_t)$ and has variance at most $V_{\tilde x_t}(Q_t)$.
The virtual order can therefore be updated as before, while the replenishment rule separately enforces physical feasibility.

Second, recovering the correct feedback does not remove the loss incurred by stock held above the target.
The preceding regret analysis controls losses at the randomized targets, giving bounds on {the corresponding perishable comparison}
\[
\tilde r_T^{\mathrm P}
\coloneqq
\sum_{t=1}^T
\ell^{\mathrm{NP}}(\tilde x_t+\tilde u_t,\tilde d_t)
-
\min_{q\in[0,M]}
\sum_{t=1}^T\ell^{\mathrm{NP}}(q,\tilde d_t).
\]
{Thus $\tilde r_T^{\mathrm P}$ is regret evaluated at the randomized targets, with no carryover or terminal term.}
These bounds apply under their respective assumptions, with $h_{\mathrm H}$ in place of $h$ and {$\max\{h_{\mathrm H},m(M)\}$} in place of $G$.
Actual losses, however, are evaluated at $\tilde q_t$, which may exceed $\tilde x_t+\tilde u_t$.
Write $\tilde e_t\coloneqq\tilde q_t-(\tilde x_t+\tilde u_t)=(\tilde y_t-\tilde x_t-\tilde u_t)_+$ for the inventory held above the target.
This additional stock cannot increase shortage loss, and its additional holding loss is at most $h_{\mathrm H}\tilde e_t$.
Hence $\ell^{\mathrm{NP}}(\tilde q_t,\tilde d_t)-\ell^{\mathrm{NP}}(\tilde x_t+\tilde u_t,\tilde d_t)\le h_{\mathrm H}\tilde e_t$, and the regret comparison gives
\[
\tilde r_T^{\mathrm{NP}}
\le
\tilde r_T^{\mathrm P}
+
h_{\mathrm H}\sum_{t=1}^T\tilde e_t
+
cM.
\]
It remains to bound the cumulative inventory held above the randomized targets.

Recall that {$\tilde e_t\coloneqq\tilde q_t-(\tilde x_t+\tilde u_t)=(\tilde y_t-\tilde x_t-\tilde u_t)_+$}.
The inventory dynamics give
\begingroup
\[
\begin{aligned}
\tilde e_{t+1}
&=\bigl(\tilde e_t+\tilde x_t+\tilde u_t - \tilde d_t -(\tilde x_{t+1}+\tilde u_{t+1})\bigr)_+
\\ & 
=
\bigl(\tilde e_t-\tilde d_t
       +(\tilde x_t-\tilde x_{t+1})
       +(\tilde u_t-\tilde u_{t+1})\bigr)_+
\le {(\tilde e_t+\tilde u_t-\tilde d_t)_+ + h_{\mathrm H}\eta_t.}
\end{aligned}
\]
\endgroup
The first equality follows from {$\tilde y_{t+1}=(\tilde e_t+\tilde x_t+\tilde u_t-\tilde d_t)_+$} and $\tilde x_{t+1}+\tilde u_{t+1}\ge0$.
The inequality uses the nonnegativity of $\tilde u_{t+1}$ and $\tilde x_t-\tilde x_{t+1}\le h_{\mathrm H}\eta_t$, which follows from $\tilde g_t\le h_{\mathrm H}$.

The term $h_{\mathrm H}\eta_t$ bounds the excess inventory created when the virtual order decreases, while $(\tilde e_t+\tilde u_t-\tilde d_t)_+$ is the excess remaining after demand.
The following proposition uses a uniform positive lower bound on the conditional mean of demand to ensure that demand reduces excess inventory in expectation, allowing demand to depend on the full past.

\begin{proposition}
\label{prop:non-perishable-bound}
Consider the nonperishable case starting from $\tilde y_1=0$ with deterministic step sizes $\eta_t>0$.
Suppose that $\tilde u_t$ and $\tilde d_t$ are conditionally independent given $\mathcal H_{t-1}$ and that, for some constant $\underline d>0$, demand satisfies $\tilde d_t\in[0,M]$ and $\E[\tilde d_t\mid\mathcal H_{t-1}]\ge\underline d$ almost surely in every period.
Then, for every $T\ge1$,
\[
\E\!\Big[\sum_{t=1}^T \tilde e_t\Big]
\le
\Big(\frac{M}{\underline d}-1\Big)
\sum_{t=1}^{T-1}\E[\tilde u_t]
+
\frac{h_{\mathrm H}M}{\underline d}
\sum_{t=1}^{T-1}\eta_t.
\]
\end{proposition}

The proof is given in Appendix~\ref{app:non-perishable}.
Conditional independence of $\tilde d_t$ and $\tilde u_t$ given $\mathcal H_{t-1}$ implies
$\E[\tilde d_t\mid\mathcal H_{t-1},\tilde u_t]
=\E[\tilde d_t\mid\mathcal H_{t-1}]
\ge\underline d$.
Thus, the lower bound on conditional expected demand remains valid after conditioning on the current inventory increment.
This allows us to bound the expected cumulative inventory held above the randomized targets $\tilde x_t+\tilde u_t$ in terms of the expected total randomization increments and the cumulative step sizes.
Both quantities are controlled by the preceding regret analysis.
Combining these bounds yields regret guarantees for the nonperishable implementation of Algorithm~\ref{rn:alg:randomized-osgd}.
Specifically, under the assumptions of Proposition~\ref{prop:non-perishable-bound}, choosing $\eta_t=\eta_0t^{-2/3}$ with $\eta_0>0$ and selecting $Q_t$ as in Theorem~\ref{rn:thm:general-regret} gives $O(T^{2/3})$ expected regret whenever $\sum_{t=1}^T\epsilon_t=O(T^{2/3})$.
If demands are also independent and the expected losses associated with $\ell^{\mathrm{NP}}$ satisfy either curvature condition in Theorem~\ref{rn:thm:accelerated}, the step sizes and distribution choices in that theorem give $O(\sqrt T)$ expected regret whenever $\sum_{t=1}^T\epsilon_t=O(\sqrt T)$.
Thus, under the stated assumptions, inventory carryover changes the regret bounds but not their rates.

%% file: _vcurrent/numerical_section.tex

\section{Numerical studies}
\label{sec:numerical-studies}

We evaluate the proposed policy against alternative policies using synthetic demand and real data.
The synthetic experiments provide controlled comparisons under stationary demand and gradual distributional drift.
In the real-data experiments, recorded sales {serve} as demand {inputs, and each policy's censored observations are generated from those} inputs without fitting a demand model.

\subsection{Common experimental settings}

All experiments use the perishable inventory model in Section~2, with demand and physical orders in $[0,M]$ and the excess-inventory cost coefficient normalized to $h=1$.
We set $M=10$ in the synthetic experiments and specify the real-data demand bounds in the corresponding section.
We use piecewise-linear and quadratic shortage costs, given respectively by
\[
\psi_{\mathrm{PWL}}(r)=3r+(r-1)_++2(r-3)_++3(r-5.5)_+
\quad\text{and}\quad
\psi_{\mathrm{quad}}(r)=3r+1.5r^2.
\]

Our proposed policy, OPT, implements Algorithm~\ref{rn:alg:randomized-osgd} with the optimized randomization distribution $Q_t=Q^\star_{\tilde x_t}$.
We compute $Q_t$ at step size $\eta_t$, using the recursion in Theorem~\ref{rn:thm:pwl} for the piecewise-linear cost and the construction in Example~\ref{rn:ex:quadratic-design} for the quadratic cost.
The step size is $\eta_t=c_\eta(M/G)t^{-2/3}$, where $G=\max\{h,m(M)\}$ bounds the magnitude of an exact loss subgradient.

As a full-information reference, we consider FI-SGD, which uses the same projected subgradient update as OPT but computes the subgradient from exact demand.
It therefore does not require randomized ordering to reveal demand.
We set its step size to $\eta_t=c'_\eta(M/G)t^{-1/2}$ following the standard online subgradient descent algorithm.

We also consider two sales-only {comparison policies}: a learning--doing policy (LD), adapted from~\citet{yangShi2023} to our setting with a known demand bound and a continuous action space, and a sales-based sample average approximation policy (SAA).
Both use empirical loss minimization but differ in how they collect and use observations.
For a nonempty multiset $\mathcal D$ of observations treated as demand inputs, define
\[
\widehat q(\mathcal D)\in\arg\min_{q\in[0,M]}
\frac{1}{|\mathcal D|}\sum_{d\in\mathcal D}
\ell(q,d).
\]
When multiple minimizers exist, we select the smallest one.

LD divides the horizon into epochs, each beginning with one learning period followed by doing periods.
The epoch start times are $\tau_1=1$ and $\tau_j=\max\{\tau_{j-1}+1,\lceil a j^{\gamma_{\mathrm{LD}}}\rceil\}$ for $j\ge2$.
In each learning period, the policy orders $M$ units so that sales reveal demand exactly.
In each remaining period $t$ of the epoch, it orders $\widehat q(\mathcal D_t)$, where $\mathcal D_t$ contains all demand observations collected in earlier learning periods.

SAA uses a rolling window of sales observations and orders $M$ units with a fixed probability.
Its parameters are the window length $w$ and the randomization probability $p$.
For $t\ge2$, let $\mathcal S_t(w)$ contain the sales observations from the preceding $\min\{w,t-1\}$ periods, and set the operating target to $\widehat q_t=\widehat q(\mathcal S_t(w))$.
In every period $t\ge1$, the policy draws $\tilde b_t\sim\mathrm{Bernoulli}(p)$ independently of all past information.
It orders $M$ units if $\tilde b_t=1$ and $\widehat q_t$ units otherwise.
Sales from every period enter the rolling window and are treated as demand inputs, even when they may be censored.

We report mean loss and regret per period, evaluated at physical orders and including all inventory used for learning.
{We decompose total} loss into excess-inventory {loss} and shortage {loss}.
Regret is computed relative to the best fixed order in hindsight, chosen separately for each demand sequence as in Eq.~\eqref{rn:eq:regret}.

For each policy, we select hyperparameters on calibration data separate from the evaluation data, using mean loss per period as the selection criterion.
For OPT and FI-SGD, we search over $c_\eta, c'_\eta\in\{1/32,1/16,1/8,1/4,1/2,1,2,4\}$.
For LD, we search over $(a,\gamma_{\mathrm{LD}})\in\{0.25,0.5,1,2\}\times\{1.5,2\}$.
For SAA, we use $(p,w)\in\{0.05,0.1,0.15,0.2\}\times\{16,64\}$ for synthetic demand and $(p,w)\in\{0.05,0.2\}\times\{7,14,28,56\}$ for real data.
We use shorter rolling windows for the real-data experiments because the available demand sequences are short.
Ties are resolved in the listed candidate order, with Cartesian grids ordered lexicographically.
Hyperparameters are calibrated separately in the synthetic and real-data settings.
All policies begin each demand sequence without any initial demand observations.
We initialize OPT and FI-SGD at {$\tilde x_1=M/2$} and SAA at $\widehat q_1=M/2$.
Each policy follows its ordering rule from period~1; in particular, LD first orders $M$ units.

\subsection{Synthetic demand experiments}

We consider one stationary demand profile and one profile with gradual distributional drift.
Let $\mathrm{TN}(\mu,v)$ denote a normal distribution with parent mean $\mu$ and variance $v$, truncated to $[0,10]$.
The moments after truncation need not equal $(\mu,v)$.
The stationary bimodal surges profile is the mixture $0.95\,\mathrm{TN}(4.5,0.16)+0.05\,\mathrm{TN}(8.5,0.64)$.
Its first component represents regular demand concentrated near $4.5$, whereas its second component represents occasional demand surges near $8.5$.
Each period uses an independent draw from this mixture.
This profile examines the cost of acquiring information about infrequent high demand when ordering at the maximum demand bound can substantially exceed typical demand.

The gradual drift profile has independent demands with time-varying distributions \(P_t=\mathrm{TN}(\mu_t,0.64),
\;
\mu_t=4+3\frac{t-1}{1999},
\; t=1,\ldots,T.\) The denominator $1999$ is fixed across horizons, so shorter experiments use shorter prefixes of the same drift profile.
Over $T=2000$ periods, $\mu_t$ increases from $4$ to $7$.
This profile examines how policies adjust to increasing demand, particularly whether reliance on observations from earlier, lower-demand periods leads to insufficient orders and higher shortage costs.

For both profiles, we consider $T\in\{100,200,500,1000,2000\}$.
For each combination of shortage cost, demand profile, and horizon, we use $100$ independently generated demand sequences for calibration and $1000$ fresh sequences for evaluation.
All policies face the same demands within each sequence.
Each horizon is calibrated separately.

\begin{figure}[tbp]
\centering
\includegraphics[width=\linewidth]{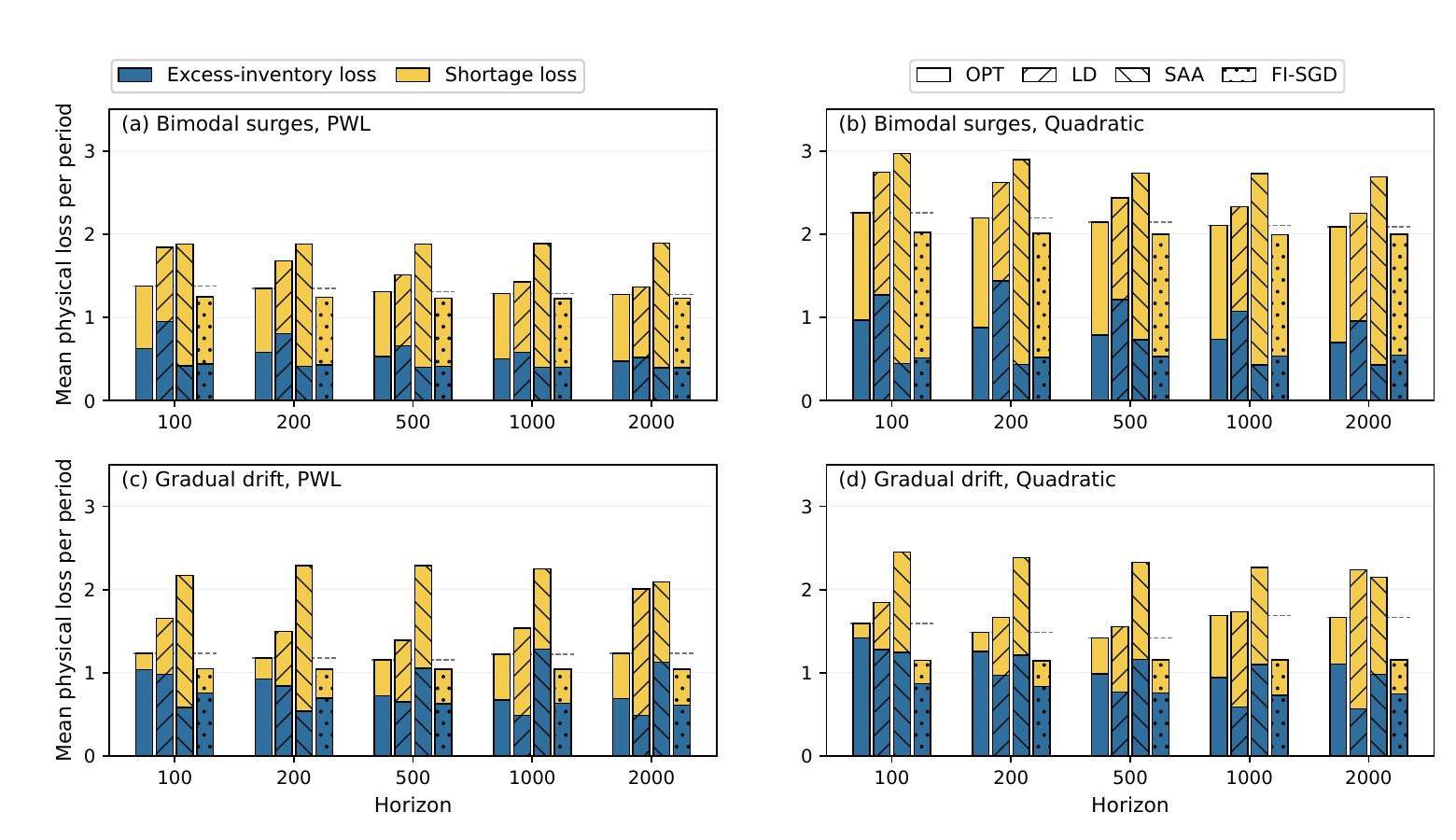}
\caption{Mean loss per period under synthetic demand.
Panels (a)--(d) show bimodal surges with the piecewise-linear and quadratic {costs}, followed by gradual drift with the piecewise-linear and quadratic {costs}.
Each horizon group contains OPT, LD, SAA, and FI-SGD, from left to right.
Blue and yellow segments are excess-inventory and shortage loss; patterns identify policies.
Dashed lines mark OPT's mean total loss per period.}
\label{fig:numerical-synthetic-loss}
\end{figure}

Figure~\ref{fig:numerical-synthetic-loss} reports mean loss per period for each policy and setting.
OPT has lower mean loss than both sales-only {comparison policies} at every reported horizon under both demand profiles and {costs}.
FI-SGD, which receives exact demand rather than censored sales, has lower loss than OPT in all displayed settings.

Under stationary bimodal surges demand, LD narrows the loss gap relative to OPT as the horizon increases.
As $T$ grows from $100$ to $2000$, its percentage gap in mean loss relative to OPT falls from $33.9\%$ to $6.9\%$ under the piecewise-linear {cost} and from $21.7\%$ to $7.7\%$ under the quadratic {cost}.
This narrowing gap is consistent with the accumulation of fully revealed demands from an unchanged distribution.
Nevertheless, ordering $M$ to obtain these observations can be expensive when typical demand is concentrated near $4.5$.
Here, our proposed randomization obtains useful first-order information at a lower operating cost than the LD strategy.

SAA improves less as the horizon increases.
At $T=2000$, its mean loss remains $48.4\%$ above OPT's under the piecewise-linear {cost} and $28.6\%$ above OPT's under the quadratic {cost}.
Its fixed-length window discards older observations even though the demand distribution is unchanged.
It also treats sales as demand: when demand exceeds the order quantity, the unmet demand is missing from its sample, which can lead to orders that are too low. 
This is also consistent with SAA's small excess-inventory loss.
Occasional orders of $M$ reveal demand but can leave substantial excess inventory.

Under gradual drift, the value of accumulating more observations must be balanced against their declining relevance to current demand.
At $T=2000$, LD incurs $62.9\%$ more loss than OPT under the piecewise-linear {cost} and $34.3\%$ more under the quadratic {cost}.
Because LD retains all learning observations, its empirical objective continues to reflect earlier, lower-demand periods as demand increases.
This is consistent with the larger shortage component in Figure~\ref{fig:numerical-synthetic-loss}(c)--(d).
SAA's rolling window limits the retention of outdated observations but does not eliminate the effects of censoring or the cost of randomized full revelation.
At $T=2000$, SAA incurs $69.8\%$ more loss than OPT under the piecewise-linear {cost} and $28.8\%$ more under the quadratic {cost}.

\begin{figure}[tbp]
\centering
\includegraphics[width=\linewidth]{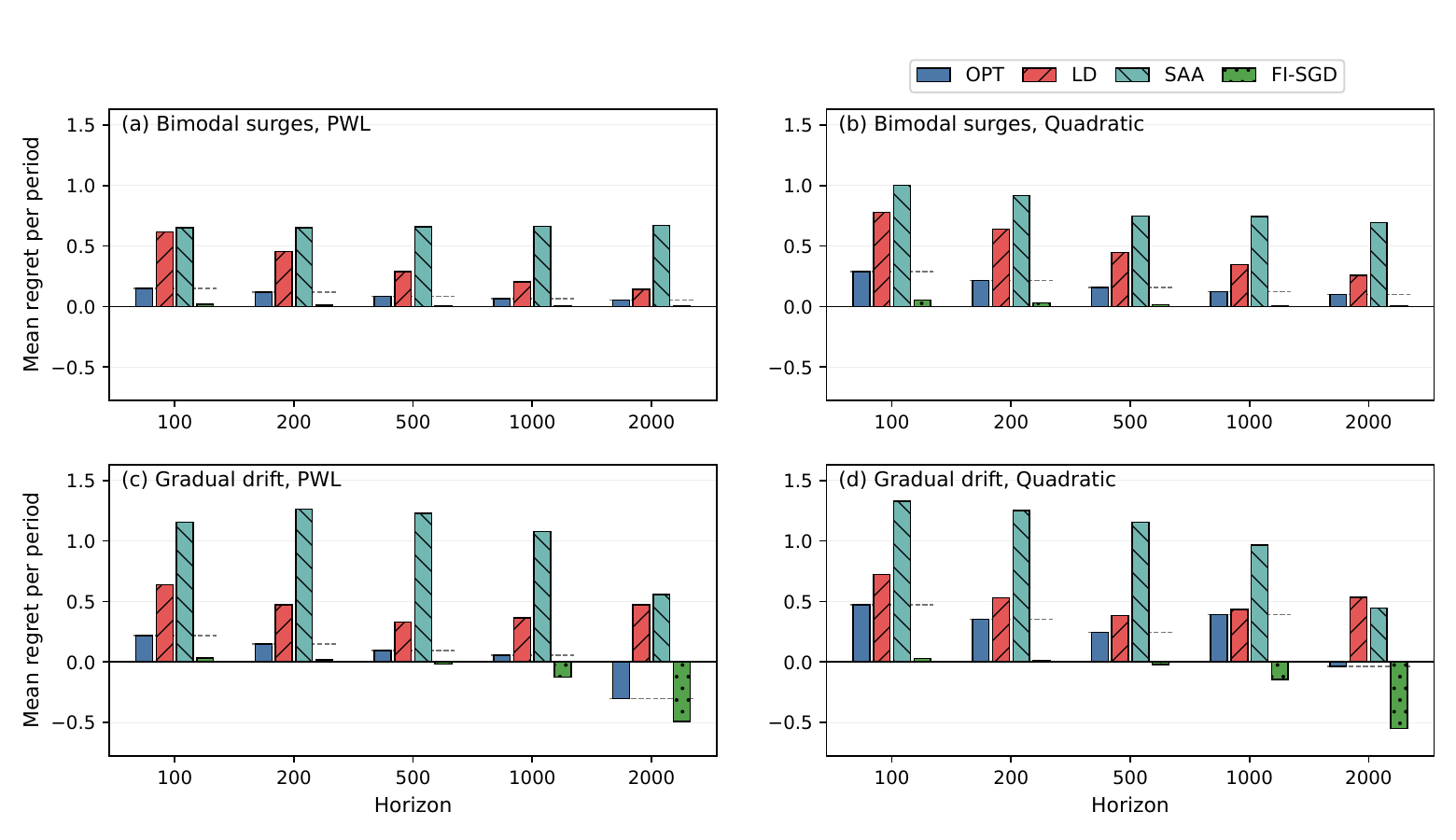}
\caption{Mean regret per period under synthetic demand.
Panel assignments and policy order are the same as in Figure~\ref{fig:numerical-synthetic-loss}.
The comparator is the best fixed order in hindsight for each demand sequence.
Dashed segments mark OPT's mean.
Negative regret indicates lower cumulative loss than the fixed-order comparator, not negative loss.}
\label{fig:numerical-synthetic-regret}
\end{figure}

Figure~\ref{fig:numerical-synthetic-regret} compares the policies with the best fixed order in hindsight.
Under stationary demand, mean regret per period declines for {OPT} and {LD} as the horizon increases, consistent with sublinear regret.
Under gradual drift, {OPT}'s mean regret per period becomes negative at $T=2000$: $-0.3024$ under the piecewise-linear {cost} and $-0.0373$ under the quadratic {cost}.
Its adaptive orders therefore incur lower cumulative loss on average than the comparator, which must use a single order quantity throughout each demand sequence.
This illustrates the benefit of adjusting orders as demand changes, even when the fixed-order comparator is chosen with full knowledge of the demand sequence.
For {LD}, mean regret per period first decreases and then increases with the horizon.
This pattern suggests that, under nonstationary demand, \textsc{LD}'s regret is driven primarily by limited data at short horizons and by reliance on outdated observations at longer horizons.
This comparison highlights the adaptability of our proposed policy, \textsc{OPT}, across different demand environments.

\subsection{Real data experiments}

The real-data experiments use the public Dingdong-Inc/FreshRetailNet-50K dataset \citep{dingdong2025freshretailnet,wang2025freshretailnet}.
We combine the released training and evaluation data and aggregate hourly sales over $[06{:}00,22{:}00)$ for each store--product pair and recorded date.
The resulting daily quantities are multiplied by $10$ and ordered chronologically to form the experimental demand sequences.
The {cost} parameters apply to these scaled quantities.
Zeros and observations with historical stockouts are retained.
The experiments thus evaluate performance on recorded sales inputs, which may already be censored by historical stockouts.
Each retained observation constitutes one period; missing dates neither create additional periods nor restart the policies.

We choose eligible pairs to have at least $42$ retained daily observations and no more than seven zeros among their first $14$ observations, yielding $49{,}108$ pairs from $897$ stores.
For each selected pair $i$, we process the complete sequence, including the first $14$ observations, chronologically under the common initialization rule.
Its horizon $T_i$ is the number of retained observations.
Before period~1, all policies receive the same ex post bound $M_i$, defined as the maximum quantity in that pair's complete scaled sequence.

We repeat the calibration and evaluation procedure independently $100$ times, referring to each repetition as a trial.
In each trial, we first sample $50$ evaluation pairs uniformly without replacement from the eligible pool.
We then sample $20$ calibration pairs uniformly without replacement from pairs whose stores do not appear in the evaluation set.
Thus, calibration and evaluation use different stores within each trial.
Pairs may appear again in later trials.
All policies use the same calibration and evaluation pairs within a trial.

For calibration, we evaluate every candidate parameter configuration on the $20$ calibration pairs.
Each randomized policy is run $30$ times on each pair's complete demand sequence; deterministic {policies} are run once.
These policy runs use the same recorded demand sequence and differ only in the policy's random choices.
For each configuration, we first average loss per period over its runs on each pair and then average equally across the $20$ pairs.
We select the configuration with the lowest resulting mean loss.
This parameter selection is repeated from scratch in every trial.

We then evaluate each selected policy under the calibrated parameter configuration on the $50$ evaluation pairs, using $200$ policy runs per pair for randomized policies and one run for deterministic policies.
For each performance metric, we first average its per-period values over runs on each pair and then average equally across the $50$ pairs.
Each pair receives equal weight, regardless of the length of its demand sequence.
This produces one evaluation average per policy and metric for that trial.
The reported performance is the average of these trial-level results over all $100$ trials.

The reported error bars are approximate $95\%$ Monte Carlo confidence intervals.
For a given policy and metric, let $z_b$ denote the evaluation average from trial $b$, for $b=1,\ldots,100$.
Writing $\bar z$ and $s$ for the sample mean and sample standard deviation of these $100$ values, we compute the interval as \(\bar z \pm t_{99,0.975}\frac{s}{\sqrt{100}}.\) Here, $t_{99,0.975}$ is the $97.5$th percentile of the Student-$t$ distribution with $99$ degrees of freedom.
The degrees of freedom are $100-1=99$ because the variance is estimated from the $100$ trial averages.
The policy runs within each trial have already been averaged before this calculation.
For regret comparisons between policies, we first compute the difference between their evaluation averages within each trial.
We then apply the same interval calculation to the resulting $100$ paired differences.
These intervals measure uncertainty from sampling calibration and evaluation pairs, selecting parameters, and randomizing policy decisions.

\begin{figure}[tbp]
\centering
\begin{minipage}[t]{0.485\linewidth}
\centering
\includegraphics[width=\linewidth]{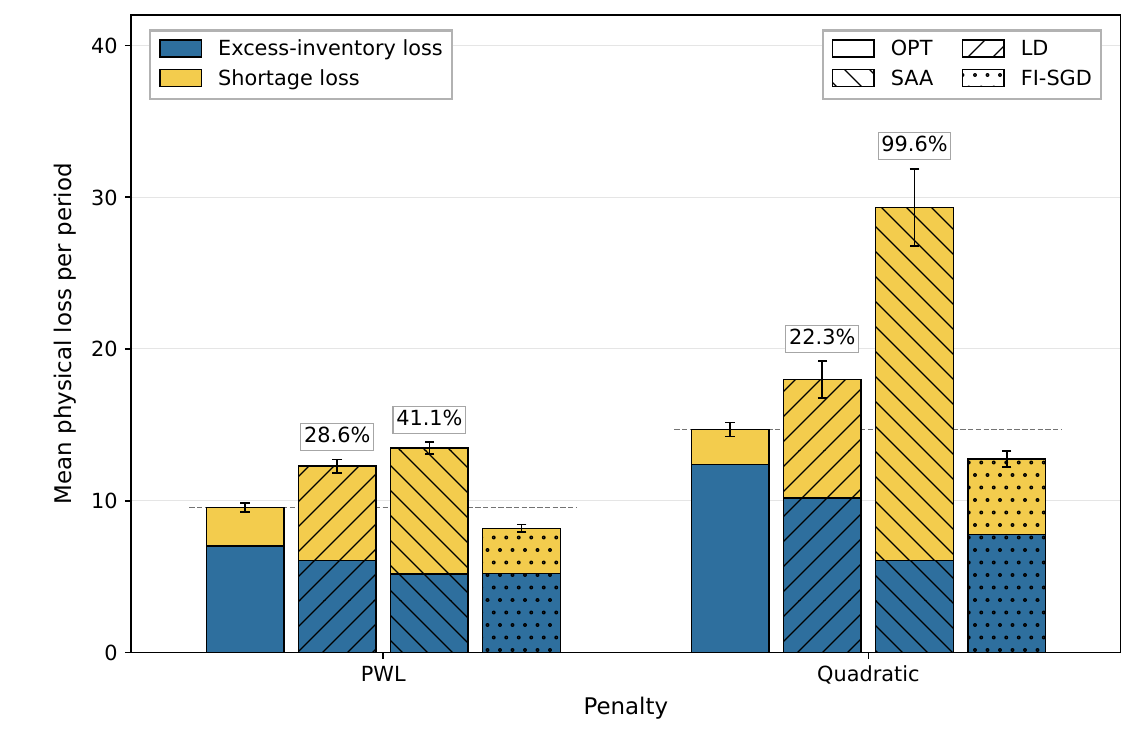}
\par\smallskip
(a) Mean loss per period.
\end{minipage}\hfill
\begin{minipage}[t]{0.485\linewidth}
\centering
\includegraphics[width=\linewidth]{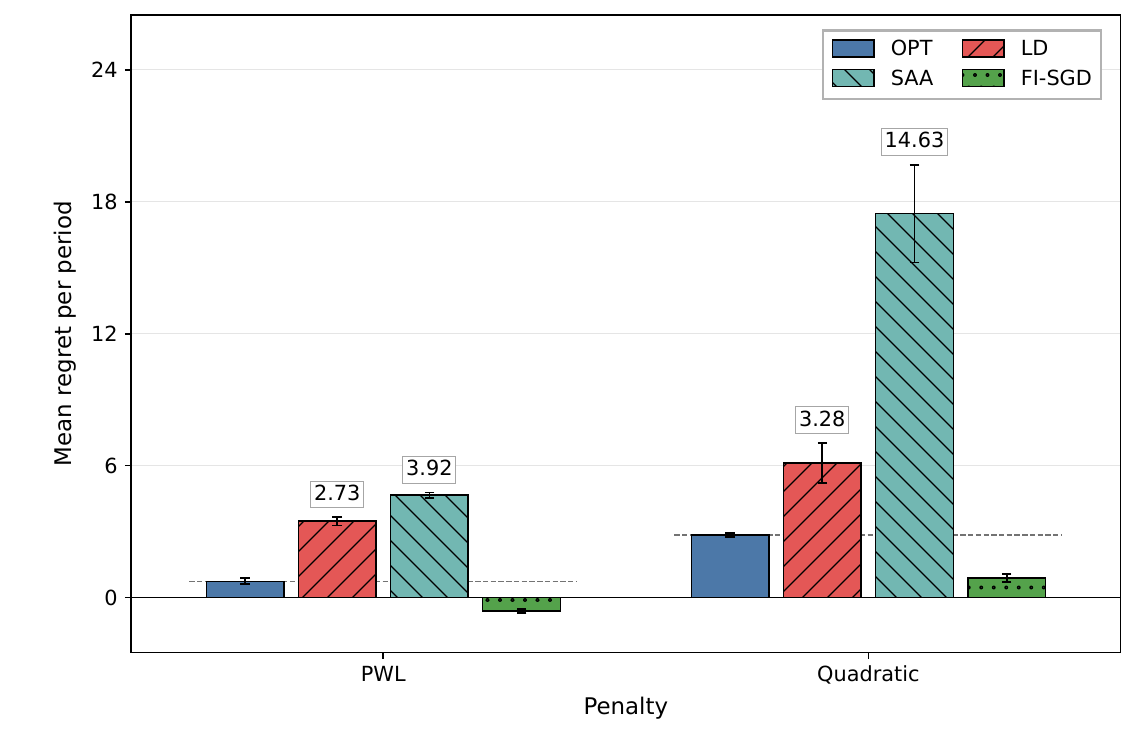}
\par\smallskip
(b) Mean regret per period.
\end{minipage}
\caption{Performance on real data, averaged over $100$ {trials}.
Within each {cost} group, policies appear as OPT, LD, SAA, and FI-SGD, from left to right.
In panel (a), blue and yellow segments represent excess-inventory and shortage loss; annotations show the percentage {increase in total} loss of LD and SAA relative to OPT.
In panel (b), annotations show their absolute regret differences from OPT.
Dashed lines mark OPT's mean.
Error bars are approximate $95\%$ Monte Carlo intervals conditional on the observed dataset.}
\label{fig:numerical-real-data}
\end{figure}

Figure~\ref{fig:numerical-real-data}(a) shows that OPT achieves the lowest mean loss among the sales-only policies under both shortage {costs}.
Under the piecewise-linear {cost}, LD and SAA incur $28.6\%$ and $41.1\%$ more loss than OPT, respectively.
Under the quadratic {cost}, the corresponding gaps are $22.3\%$ and $99.6\%$.
The advantage over SAA is therefore particularly pronounced {under the quadratic shortage cost}.

The loss decomposition explains this advantage: OPT incurs higher excess-inventory loss than either sales-only {comparison policy}, but its lower shortage loss more than offsets this increase.
For example, under the quadratic {cost}, OPT's excess-inventory loss is $12.3988$ per period, compared with $6.0709$ for SAA, whereas its shortage loss is only $2.2946$, compared with $23.2518$.
The same trade-off holds under the piecewise-linear {cost}.
These results suggest that businesses facing high shortage costs may benefit from ordering more to reduce shortages, even if this increases excess-inventory costs.

Figure~\ref{fig:numerical-real-data}(b) presents these losses relative to each sequence's best fixed order in hindsight.
Because the policies share the same comparators, the panel shows their remaining gaps to those comparators without changing their ranking.
FI-SGD has lower mean loss than the sales-only policies and achieves negative mean regret under the piecewise-linear {cost}.
Its better performance illustrates the potential benefit of {access to the uncensored experimental} demand {inputs}.
In these experiments, the small gap between our proposed policy {and} FI-SGD shows that our policy can approach full-information performance using only censored sales at a modest additional operating cost.

%% file: _vcurrent/conclusion.tex
\section{Conclusion}
\label{sec:conclusion}

We study inventory learning under censored demand and nonlinear convex shortage costs.
{A stockout can conceal the shortage amount and hence the marginal loss needed to update the order.}
Our policy overcomes this difficulty by adding {random additional inventory} to a virtual order and using the resulting sales to construct an unbiased subgradient estimate.
The randomization is chosen to balance subgradient estimation accuracy against {a bound on} the {loss increase from} additional inventory.
The unbiased subgradient {estimate} allows {the} policy to learn directly from sales without first estimating the demand distribution.

Our proposed approach does not require {stationarity} or {independence across periods in} the perishable inventory {model}.
Demand may respond to earlier orders and random draws, provided it is conditionally independent of the current period's randomization given the full past.
It achieves $O(T^{2/3})$ expected regret against the best fixed order in hindsight on the same realized demand sequence, including the cost of the additional randomized inventory used to acquire information.
Under independent demand and the stated conditions on expected losses, the rate improves to $O(\sqrt{T})$.
Matching lower bounds establish that both rates are optimal for the corresponding classes with a fixed marginal-cost bound.
For nonperishable inventory, a uniform positive lower bound on expected demand conditional on the full past preserves the $O(T^{2/3})$ rate against the best fixed base-stock policy in hindsight.
The $O(\sqrt{T})$ rate additionally requires independent demand and the stated expected-loss curvature conditions.

The numerical experiments {illustrate the value} of this approach.
The proposed policy achieves lower mean loss and {mean} regret {per period} than \textsc{LD} and \textsc{SAA} in the reported synthetic and real-data experiments.
{In the real-data experiments}, its lower shortage loss more than offsets its higher excess-inventory loss.
Under gradual drift, adjusting orders also allows the policy to outperform the best fixed order in hindsight on average at the longest horizon.
These findings illustrate how learning from censored sales can improve ordering decisions, {even though demand may remain censored.}

Several questions remain open.
First, incorporating a variation budget could yield guarantees that adapt to the magnitude of distributional changes.
An appropriate time-varying comparator would further distinguish the cost of tracking changing demand from the benefit of adaptation relative to a fixed order.
Second, the nonperishable extension relies on demand conditions that control the persistence of excess inventory.
Understanding how to design randomization when these conditions are unavailable would broaden the treatment of intertemporal inventory dynamics.

Finally, historical censored observations could be used to refine the randomization design as information about demand accumulates.
The current one-period objective protects against the worst-case demand, which need not be representative of a particular demand environment.
{A promising direction is therefore to combine distribution-free protection with instance-dependent improvements in information acquisition and operating performance.}

%% file: _vcurrent/appendix.tex

\section{Proofs of results in Section~\ref{rn:sec:model}}
\label{rn:app:estimator}

We first prove Theorem~\ref{rn:thm:estimator}, then justify the auxiliary-randomization claim in Section~\ref{rn:sec:admissible}.

\noindent
\textbf{Proof of \thmref{rn:thm:estimator}{Theorem}:} Set $L \coloneqq M-x$.
The argument is pointwise in the realized demand $d$, so the only randomness in the conditional expectation comes from $\tilde u$.
{For an integrable correction $W$, define the auxiliary estimator of the increase in marginal shortage cost by}

\begin{equation}
\label{rn:eq:curvature-estimator}
\tilde z_W(x,\tilde u,\tilde s) \coloneqq 
\begin{cases}
\Delta((\tilde s-x)_+),&\tilde s<x+\tilde u,\\
W(\tilde u),&\tilde s=x+\tilde u.
\end{cases}
\end{equation}
{The subgradient estimate in Eq.~\eqref{rn:eq:gradient-estimator} can then be written as}
\[
\tilde g_W(x,\tilde u,\tilde s)
=h\mathbf 1\{\tilde s<x\}-m(0)\mathbf 1\{\tilde s\ge x\}-\tilde z_W(x,\tilde u,\tilde s).
\]
{Write $\tilde z_Q \coloneqq \tilde z_{W_Q}$ when the unbiased correction $W_Q$ exists.}
{We also use the auxiliary measure $\mu_Q$ on $[0,L]$ defined by $\mu_Q([0,r]) \coloneqq F(r)\Delta(r)$ for $0\le r\le L$.}
{This is a finite nonnegative Lebesgue--Stieltjes measure because $F\Delta$ is nondecreasing and right-continuous, $F(0)\Delta(0)=0$, and $F(L)\Delta(L)=\Delta(L)<\infty$}.

Suppose first that $d<x$.
Since $\tilde u\ge0$, we have $\tilde q=x+\tilde u\ge x>d$ and hence $\tilde s=d<\tilde q$.
The first branch in the definition of $\tilde z_W$ therefore applies and gives $\tilde z_W=\Delta((d-x)_+)=\Delta(0)=0$.
Thus the curvature estimator is unbiased automatically when $d<x$.

Suppose now that $d\ge x$ and set $r \coloneqq d-x\in[0,L]$.
If $\tilde u\le r$, then $\tilde q=x+\tilde u\le d$, so $\tilde s=\tilde q=x+\tilde u$ and the estimator equals $W(\tilde u)$.
If $\tilde u>r$, then $\tilde q>d$, so $\tilde s=d<\tilde q$ and the estimator equals the observed value $\Delta(d-x)=\Delta(r)$.
Splitting the conditional expectation over these two events gives
\[
\begin{aligned}
\mathbb E[\tilde z_W(x,\tilde u,\tilde s)\mid x,d]
&=\int_{[0,r]}W(u)Q(\mathrm{d}u)+\Delta(r)\mathbb P(\tilde u>r)=
\int_{[0,r]}W(u)Q(\mathrm{d}u)+\Delta(r)\{1-F(r)\}.
\end{aligned}
\]
This expression equals the target $\Delta(r)$ if and only if $\int_{[0,r]}W\,\mathrm{d}Q=F(r)\Delta(r)$.
Since every $r$ corresponds to a unique $d\in[x,M]$, unbiasedness for every $d\in[0,M]$ is equivalent to Eq.~\eqref{rn:eq:unbiased-identity}.

It remains to characterize when a function satisfying Eq.~\eqref{rn:eq:unbiased-identity} exists.
Suppose first that $W\in L^1(Q)$ satisfies Eq.~\eqref{rn:eq:unbiased-identity}, and define the finite signed measure $\lambda_W$ by $\lambda_W(B) \coloneqq \int_BW(u)Q(\mathrm{d}u)$ for every Borel set $B\subseteq[0,L]$.
For every $r\in[0,L]$, Eq.~\eqref{rn:eq:unbiased-identity} gives $\lambda_W([0,r])=F(r)\Delta(r)=\mu_Q([0,r])$.
It follows by subtraction that $\lambda_W$ and $\mu_Q$ agree on every interval $(a,b]$ and on the singleton $\{0\}$.
These sets generate the Borel sigma-algebra on $[0,L]$, so the two finite signed measures agree on every Borel set.
Consequently, $\mu_Q=\lambda_W=WQ$, which implies $\mu_Q\ll Q$.
The same measure identity shows that $W$ is a version of the Radon--Nikodym derivative $\mathrm{d}\mu_Q/\mathrm{d}Q$.

Conversely, suppose that $\mu_Q\ll Q$.
The Radon--Nikodym theorem provides a measurable function $W_Q \coloneqq \mathrm{d}\mu_Q/\mathrm{d}Q$ satisfying $\mu_Q(B)=\int_BW_Q\,\mathrm{d}Q$ for every Borel set $B\subseteq[0,L]$.
The measure $\mu_Q$ is finite because $F(L)=1$ and $\mu_Q([0,L])=F(L)\Delta(L)=\Delta(L)<\infty$.
Therefore $W_Q\in L^1(Q)$.
Applying the Radon--Nikodym identity to $B=[0,r]$ gives $\int_{[0,r]}W_Q\,\mathrm{d}Q=\mu_Q([0,r])=F(r)\Delta(r)$.
Thus $W_Q$ satisfies Eq.~\eqref{rn:eq:unbiased-identity} and hence yields an unbiased curvature estimator.

Finally, if $W_1$ and $W_2$ are two integrable unbiased corrections, then both are Radon--Nikodym derivatives of $\mu_Q$ with respect to $Q$.
The uniqueness of the Radon--Nikodym derivative therefore gives $W_1=W_2$ $Q$-almost surely.
The Stieltjes product rule gives $\mathrm{d}\mu_Q=\Delta\,\mathrm{d}Q+F(\cdot^-)\nu(\mathrm{d}u)$, which proves the {absolute-continuity} characterization and the {correction formula} in Eq.~\eqref{rn:eq:rn-correction}.

Because $\tilde s=\min\{d,\tilde q\}$ and $\tilde q=x+\tilde u\ge x$, we have $\mathbf 1\{\tilde s<x\}=\mathbf 1\{d<x\}$ and $\mathbf 1\{\tilde s\ge x\}=\mathbf 1\{d\ge x\}$.
The first two terms of $\tilde g_Q$ therefore coincide with the corresponding terms of $g(x;d)$ in Eq.~\eqref{rn:eq:gradient-decomposition}.
The preceding argument establishes $\mathbb E[\tilde z_Q\mid x,d]=\Delta((d-x)_+)$, so the remaining term also agrees in expectation, giving $\mathbb E[\tilde g_Q\mid x,d]=g(x;d)$.

For the variance claims, suppose $W_Q\in L^2(Q)$.
Since the first two terms of $\tilde g_Q$ are fixed conditional on $x$ and $d$, $\tilde g_Q$ and $\tilde z_Q$ have the same conditional variance.
We will show that this variance equals $V_Q((d-x)_+)$, where $V_Q(r) \coloneqq \int_{(0,r]}W_Q(u)^2Q(\mathrm{d}u)-F(r)\Delta(r)^2$ for $0\le r\le L$.
Taking $r=0$ in Eq.~\eqref{rn:eq:unbiased-identity} gives $W_Q(0)Q(\{0\})=F(0)\Delta(0)=0$.
Hence $W_Q(0)^2Q(\{0\})=0$, so a possible atom at zero contributes nothing to the second moment and can be omitted from the integral.

If $d<x$, then $\tilde z_Q$ is identically zero and $(d-x)_+=0$.
Its conditional variance is therefore $0=V_Q(0)$, as required.
Now suppose that $d\ge x$ and set $r \coloneqq d-x$.
As in the unbiasedness calculation, the estimator equals $W_Q(\tilde u)$ on $\{\tilde u\le r\}$ and equals $\Delta(r)$ on $\{\tilde u>r\}$.
Consequently, the conditional second moment has the following form.
\[
\mathbb E[\tilde z_Q(x,\tilde u,\tilde s)^2\mid x,d]
=\int_{(0,r]}W_Q(u)^2Q(\mathrm{d}u)+\Delta(r)^2\{1-F(r)\}
=\Delta(r)^2+V_Q(r).
\]
The conditional mean is $\Delta(r)$ by the first part of the theorem, so subtracting its square gives $\operatorname{Var}(\tilde z_Q\mid x,d)=V_Q(r)$.
This proves the variance identity.

It remains to show that $V_Q$ is nondecreasing.
Fix $0\le r<s\le L$.
Subtracting Eq.~\eqref{rn:eq:unbiased-identity} at $r$ from the same identity at $s$ yields $\int_{(r,s]}W_Q\,\mathrm{d}Q=F(s)\Delta(s)-F(r)\Delta(r)=Q((r,s])\Delta(s)+F(r)\{\Delta(s)-\Delta(r)\}$.
If $Q((r,s])=0$, then the integral on the left is zero because $(r,s]$ is $Q$-null.
The preceding identity therefore gives $F(r)\{\Delta(s)-\Delta(r)\}=0$.
Moreover, $F(s)=F(r)$, and hence $V_Q(s)-V_Q(r)=-F(r)\{\Delta(s)^2-\Delta(r)^2\}=-F(r)\{\Delta(s)-\Delta(r)\}\{\Delta(s)+\Delta(r)\}=0$.

Suppose next that $Q((r,s])>0$.
We have the following bound on the variance difference.
\[
\begin{aligned}
V_Q(s)-V_Q(r)
&=\int_{(r,s]}W_Q(u)^2Q(\mathrm{d}u)-F(s)\Delta(s)^2+F(r)\Delta(r)^2\\
&\ge\frac{\bigl(Q((r,s])\Delta(s)+F(r)\{\Delta(s)-\Delta(r)\}\bigr)^2}{Q((r,s])}
-F(s)\Delta(s)^2+F(r)\Delta(r)^2\\
&=\frac{F(r)F(s)}{Q((r,s])}\{\Delta(s)-\Delta(r)\}^2\ge0,
\end{aligned}
\]
where the inequality follows from Cauchy--Schwarz, $(\int_{(r,s]}W_Q\,\mathrm{d}Q)^2\le(\int_{(r,s]}W_Q^2\,\mathrm{d}Q)Q((r,s])$.
Thus $V_Q$ is nondecreasing on $[0,L]$.
Since $(d-x)_+$ ranges over this entire interval, the worst case is attained at $r=L$.
Finally, $F(L)=1$, so $V_Q(L)=\int_{(0,L]}W_Q^2\,\mathrm{d}Q-\Delta(L)^2=V_x(Q)$.
\hfill
\Halmos

\section{Proofs of results in Section~\ref{rn:sec:design}}
\label{rn:app:design}

\subsection{Regret decomposition and the {one-period} objective}
\label{rn:app:physical-regret}
\label{rn:app:support}

\noindent\textbf{Proof of \thmref{rn:prop:physical-regret}{Proposition}:}
Write $\tilde\xi_t:=\tilde g_t-g(\tilde x_t;\tilde d_t)$, $\tilde V_t:=V_{\tilde x_t}(Q_t)$, and $\tilde S_T:=\sum_{t=1}^T\tilde\xi_t$, and let $\tilde q_T^{\mathrm{hin}}$ be the smallest minimizer of $\sum_{t=1}^T\ell(q,\tilde d_t)$ over $[0,M]$.

The conditional independence in Assumption~\ref{ass:dependence} and Theorem~\ref{rn:thm:estimator} give $\mathbb E[\tilde\xi_t\mid\mathcal H_{t-1},\tilde d_t]=0$ and $\mathbb E[\tilde\xi_t^2\mid\mathcal H_{t-1},\tilde d_t]\le\tilde V_t$. Thus, $(\tilde\xi_t)$ is a square-integrable martingale difference sequence with respect to $(\mathcal H_t)$, so $\mathbb E[\tilde S_T]=0$ and $\mathbb E[\tilde S_T^2]=\sum_{t=1}^T\mathbb E[\tilde\xi_t^2]\le\sum_{t=1}^T\mathbb E[\tilde V_t]$. Moreover, predictability of $\tilde x_t$ gives $\mathbb E[\tilde\xi_t\tilde x_t]=0$, while conditional unbiasedness and $|g(\tilde x_t;\tilde d_t)|\le G$ imply $\mathbb E[\tilde g_t^2]\le G^2+\mathbb E[\tilde V_t]$.

\citet[Theorem~2.13]{orabona2026online} provides a projected-gradient regret bound for nonincreasing step sizes. Fixing an arbitrary sample path and applying this bound to the linear losses $q\mapsto\tilde g_tq$ yields $\sum_{t=1}^T\tilde g_t(\tilde x_t-q)\le M^2/(2\eta_T)+(1/2)\sum_{t=1}^T\eta_t\tilde g_t^2$ simultaneously for all $q\in[0,M]$. In particular, the inequality holds for the hindsight comparator $q=\tilde q_T^{\mathrm{hin}}$. Combining it with convexity and $\ell(\tilde x_t+\tilde u_t,\tilde d_t)-\ell(\tilde x_t,\tilde d_t)\le h\tilde u_t$, and taking expectations, yields $\mathbb E[\tilde r_T]\le M^2/(2\eta_T)+(G^2/2)\sum_{t=1}^T\eta_t+\sum_{t=1}^T\mathbb E[J_{\tilde x_t}(Q_t;\eta_t)]+\mathbb E[(\tilde q_T^{\mathrm{hin}}-M/2)\tilde S_T]$, where we used $\mathbb E[\tilde u_t]=\mathbb E[\int u\,Q_t(\mathrm du)]$ and $\mathbb E[\tilde S_T]=0$.


Since $|\tilde q_T^{\mathrm{hin}}-M/2|\le M/2$, Cauchy--Schwarz inequality gives $\mathbb E[(\tilde q_T^{\mathrm{hin}}-M/2)\tilde S_T]\le(M/2)(\sum_{t=1}^T\mathbb E[\tilde V_t])^{1/2}$. By Young's inequality, this is at most $M^2/(4\eta_T)+(\eta_T/4)\sum_{t=1}^T\mathbb E[\tilde V_t]\le M^2/(4\eta_T)+(1/2)\sum_{t=1}^T\mathbb E[J_{\tilde x_t}(Q_t;\eta_t)]$, where the last inequality uses $\eta_T\le\eta_t$ and $J_{\tilde x_t}(Q_t;\eta_t)\ge(\eta_t/2)\tilde V_t$. Substituting this estimate proves the claimed bound.
\hfill\Halmos

\medskip

\noindent
\textbf{Proof of \thmref{rn:lem:support}{Lemma}:} Set $L \coloneqq M-x$.
The proof moves probability mass inside each curvature-free interval to the largest point of $K_x$ not exceeding the corresponding {additional inventory}.
Equivalently, increases of the cdf inside such an interval are reassigned to that earlier point.
This can only reduce the {physical} order, and averaging the old correction over points that are merged by this map can only reduce its second moment.
We now formalize this argument.

Define a map $\pi(u) \coloneqq \max(K_x\cap[0,u])$, which is nondecreasing and well defined for every $u\in[0,L]$ because the set $K_x$ is compact and contains zero.
Let $\bar Q \coloneqq \pi_\#Q$, the law of $\pi(\tilde u)$ when $\tilde u\sim Q$.
Then $\bar Q$ is supported on $K_x$.
Moreover, $\pi(u)\le u$ for every $u$, with strict inequality whenever $u\notin K_x$.
Therefore $\mathbb E_{\bar Q}[\tilde u]\le\mathbb E_Q[\tilde u]$, and the inequality is strict if $Q([0,L]\setminus K_x)>0$.

Let $\bar F$ be the cdf of $\bar Q$.
For $r\in[0,L]$, define $A_r \coloneqq \pi^{-1}([0,r])$.
Because $\pi$ is nondecreasing, $A_r$ is an initial interval, possibly open at its right endpoint, and $Q(A_r)=\bar F(r)$.
Any part of $A_r$ lying to the right of $r$ is contained in a connected component of $[0,L]\setminus K_x$.
The curvature measure vanishes on such a component, so $\Delta$ is constant there and equals $\Delta(r)$.
By the endpoint formula for the Lebesgue--Stieltjes measure induced by $F\Delta$, with a left limit when $A_r$ is open at its right endpoint, we obtain the following identity.
\[
(\pi_\#\mu_Q)([0,r])=\mu_Q(A_r)=Q(A_r)\Delta(r)=\bar F(r)\Delta(r)=\mu_{\bar Q}([0,r]).
\]
The two finite measures agree on every initial interval, and hence $\mu_{\bar Q}=\pi_\#\mu_Q$.

Let $\tilde u\sim Q$ and write $\tilde y \coloneqq \pi(\tilde u)$, so that $\tilde y\sim\bar Q$.
Since $\mu_Q=W_QQ$ and $\mu_{\bar Q}=\pi_\#\mu_Q$, the new Radon--Nikodym derivative may be chosen as $W_{\bar Q}(\tilde y)=\mathbb E[W_Q(\tilde u)\mid\tilde y]$ almost surely.
Indeed, integrating either side against any bounded Borel function of $\tilde y$ gives the same integral with respect to $\mu_{\bar Q}$.
Conditional Jensen's inequality now yields the following second-moment bound.
\[
\int W_{\bar Q}(v)^2\bar Q(\mathrm{d}v)\le\int W_Q(u)^2Q(\mathrm{d}u)<\infty.
\]
Also, $\mu_{\bar Q}(\{0\})=\bar F(0)\Delta(0)=0$, so we may choose $W_{\bar Q}(0)=0$.
Thus $\bar Q\in\mathcal Q_x$.
Finally, both variance terms are obtained from the corresponding second moments by subtracting the same terminal term $\Delta(L)^2$.
Hence the preceding inequality implies $V_x(\bar Q)\le V_x(Q)$.
\hfill
\Halmos

\subsection{Piecewise-linear shortage {costs}}
\label{rn:app:pwl}

We first verify the variance formula used in Section~\ref{rn:sec:pwl}, with the same notation and zero-mass convention.
Whenever $p_j>0$, direct expansion gives
\[
\begin{aligned}
&p_jW_Q(u_j)^2
-(F_j\Delta_j^2-F_{j-1}\Delta_{j-1}^2)
=p_j\Big(\Delta_j+\frac{ a_j F_{j-1}}{p_j}\Big)^2
  -(F_{j-1}+p_j)\Delta_j^2
  +F_{j-1}(\Delta_j- a_j )^2\\
&=p_j\Delta_j^2
  +2 a_j F_{j-1}\Delta_j
  +\frac{ a_j ^2F_{j-1}^2}{p_j} -(F_{j-1}+p_j)\Delta_j^2
  +F_{j-1}\Delta_j^2
  -2 a_j F_{j-1}\Delta_j
  + a_j ^2F_{j-1}\\
&=\frac{ a_j ^2F_{j-1}^2}{p_j}
  + a_j ^2F_{j-1} 
  =\frac{ a_j ^2F_{j-1}(F_{j-1}+p_j)}{p_j}
  =\frac{ a_j ^2F_{j-1}F_j}{p_j}.
\end{aligned}
\]
Under this convention, the same identity holds at zero-mass {breakpoints}, with both sides equal to zero.
Because $F_J=1$ and $\Delta_0=0$, telescoping gives $\sum_{j=1}^J(F_j\Delta_j^2-F_{j-1}\Delta_{j-1}^2)=\Delta_J^2=\Delta(M-x)^2$.
This is exactly the terminal term subtracted in Eq.~\eqref{rn:eq:variance}, so $V_x(Q)=\sum_{j=1}^Ja_j^2F_{j-1}F_j/p_j$.

\medskip

\noindent
\textbf{Proof of \thmref{rn:thm:pwl}{Theorem}:} Write $c \coloneqq \eta/2$ and $d_j \coloneqq u_j-u_{j-1}$.
By Lemma~\ref{rn:lem:support}, restricting to distributions supported on $\{0,u_1,\ldots,u_J\}$ does not change the infimum.
Moreover, because $h>0$, any optimizer over $\mathcal Q_x$ must be supported on this set: otherwise, support reduction strictly decreases the inventory term without increasing the variance.
Since $z_{j-1}\ge0$, $h>0$, and $d_j>0$, the forward recursion gives $z_j^->0$ and hence $0\le\rho_j<1$ for every $j$.

To prove the result, for each $j$, let us consider a scalar function $\phi_j:[0,1)\to\mathbb R$ defined by $\phi_j(r) \coloneqq ca_j^2r/(1-r)-z_j^-r$.
This function is strongly convex because $\phi_j''(r)=2ca_j^2/(1-r)^3\ge2ca_j^2>0$.
If $z_j^-\le ca_j^2$, then $\phi_j'(0)\ge0$, so the unique minimizer is $r=0=\rho_j$ and the minimum is $0=-z_j$.
If $z_j^->ca_j^2$, then the unique solution of $\phi_j'(r)=0$ is $r=1-a_j\sqrt{c/z_j^-}=\rho_j$, and substitution gives $\phi_j(\rho_j)=-(\sqrt{z_j^-}-a_j\sqrt c)^2=-z_j$.
Consequently, $\phi_j(r)\ge-z_j$ for every $r\in[0,1)$, with equality if and only if $r=\rho_j$.

We use this result to lower-bound the objective in~\eqref{rn:eq:pwl-program}.
Let $(F_0,\ldots,F_J)$ be a feasible cdf with finite objective.
For $F_j>0$, set $r_j \coloneqq F_{j-1}/F_j$, and set $r_j \coloneqq 0$ when $F_j=0$.
Then $r_j\in[0,1)$ and $F_{j-1}=F_jr_j$ for every $j$.
Summation by parts and the identity $F_{j-1}+F_{j-1}^2/(F_j-F_{j-1})=F_jr_j/(1-r_j)$ give
\begin{equation}
\label{rn:eq:pwl-telescope}
\begin{aligned}
J_x(Q;\eta)
&=hu_J+\sum_{j=1}^JF_j\left\{ca_j^2\frac{r_j}{1-r_j}-hd_jr_j\right\} 
=
hu_J+\sum_{j=1}^JF_j\{\phi_j(r_j)+z_{j-1}r_j\}
\\ &
\ge hu_J+\sum_{j=1}^J\{-z_jF_j+z_{j-1}F_{j-1}\}=hu_J-z_J.
\end{aligned}
\end{equation}
The inequality follows from $\phi_j(r_j) \geq -z_j$ for all $r_j \in [0, 1)$.
The final sum telescopes because $F_J=1$ and $z_0=0$.
Moreover, from the above analysis of $\phi_j$, equality holds if and only if $F_{j-1}/F_j=\rho_j$ at every index with $F_j>0$.

Now, we demonstrate optimality and uniqueness of the proposed randomization distribution.
Starting from $F_J^\star=1$, define $F_{j-1}^\star \coloneqq \rho_jF_j^\star$ recursively for $j=J,\ldots,1$.
Since $0\le\rho_j<1$, we have $0\le F_{j-1}^\star\le F_j^\star$, so $(F_0^\star,\ldots,F_J^\star)$ is a feasible cdf.
If $F_j^\star>0$, then $r_j^\star=F_{j-1}^\star/F_j^\star=\rho_j$, and hence equality holds in the scalar bound $\phi_j(r_j^\star)\ge-z_j$.
If $F_j^\star=0$, then the recursion also gives $F_{j-1}^\star=0$, so both sides of the corresponding term in Eq.~\eqref{rn:eq:pwl-telescope} are zero.
Thus equality holds at every index in Eq.~\eqref{rn:eq:pwl-telescope}, and consequently $J_x(Q_x^\star;\eta)=hu_J-z_J$.
Thus $Q_x^\star$ is optimal and the stated value formula follows.
If another feasible cdf is optimal, then equality in Eq.~\eqref{rn:eq:pwl-telescope} forces $F_{j-1}=\rho_jF_j$ whenever $F_j>0$.
When $F_j=0$, monotonicity gives $F_{j-1}=0=\rho_jF_j$, so the same recursion holds at every index.
Starting from $F_J=1$, this backward recursion determines the cdf uniquely, and hence $Q_x^\star$ is the unique optimizer.

It remains to verify admissibility and the correction formula.
The mass at $u_j$ is $p_j^\star=(1-\rho_j)F_j^\star$.
Since $1-\rho_j>0$, the equality $p_j^\star=0$ implies $F_j^\star=F_{j-1}^\star=0$.
The jump of the measure induced by $F^\star\Delta$ at $u_j$ is $p_j^\star\Delta_j+a_jF_{j-1}^\star$, so it vanishes whenever $p_j^\star=0$.
The same measure has no mass at zero because $\Delta_0=0$, and therefore $\mu_{Q_x^\star}\ll Q_x^\star$.
Whenever $p_j^\star>0$, its Radon--Nikodym derivative is
\(W_{Q_x^\star}(u_j)=\Delta_j+a_j\frac{F_{j-1}^\star}{p_j^\star}
=\Delta_{j-1}+\frac{a_j}{1-\rho_j}.\)
This correction is finite on the finite support, so taking $W_{Q_x^\star}(0)=0$ gives $W_{Q_x^\star}\in L^2(Q_x^\star)$.
Finally, the definition of $\rho_j$ yields $a_j/(1-\rho_j)=\max\{a_j,\sqrt{2z_j^-/\eta}\}$.
Each forward or backward step uses a constant number of arithmetic operations, so the total complexity is $O(J)$.
\hfill
\Halmos

\subsection{Absolutely continuous {marginal shortage costs}}
\label{rn:app:continuous}
\label{rn:app:quadratic-design}

\noindent
\textbf{Proof of \thmref{rn:lem:ode}{Lemma}:} Write $\alpha \coloneqq \sqrt{2\eta}$ and $K \coloneqq \|m'\|_\infty$.
All differential identities and inequalities below hold almost everywhere.
The square root is not locally Lipschitz at zero, so existence and uniqueness are handled separately.

Extend the equation to all real values of $z$ by writing $z_+ \coloneqq \max\{z,0\}$ and $f(u,z) \coloneqq h-\alpha m'(u)\sqrt{z_+}$.
For each fixed $z$, the map $u\mapsto f(u,z)$ is measurable, and for almost every $u$, the map $z\mapsto f(u,z)$ is continuous.
Moreover, $|f(u,z)|\le h+\alpha K\sqrt{z_+}\le h+\alpha K(1+|z|)$.
The Carath\'eodory existence theorem therefore gives a local absolutely continuous solution to $z'=f(u,z)$ with $z(0)=0$ \citep[Chapter~3]{Walter1998ODE}.
The linear-growth bound and Gronwall's inequality rule out finite-time blow-up, so the solution extends to all of $[0,M]$.

Next, let us show that $z(u)$ is nonnegative.
On the set where $z<0$, the extension satisfies $z'=h$ because $z_+=0$.
For the negative part $z_- \coloneqq (-z)_+$, the chain rule for absolutely continuous functions gives $(z_-)'=-z'\mathbf 1\{z<0\}=-h\mathbf 1\{z<0\}\le0$.
Since $z_-(0)=0$ and $z_-\ge0$, it follows that $z_-$ is identically zero.
Thus the extended solution is nonnegative and hence solves the original ODE.
Because $m'\ge0$, the ODE gives $z_\eta'\le h$.
Integrating from zero yields $z_\eta(u)\le hu$ for every $u\in[0,M]$.

Now, we show strict positivity of $z(u)$.
If $K=0$, then the ODE reduces to $z_\eta'=h$, so $z_\eta(u)=hu$ and the claim is immediate.
Suppose now that $K>0$, and set $\theta \coloneqq h^2/(8\eta K^2)$.
Whenever $0\le z_\eta(u)\le\theta$, the derivative satisfies the following lower bound.
\[
z_\eta'(u)=h-\alpha m'(u)\sqrt{z_\eta(u)}\ge h-\alpha K\sqrt\theta=\frac h2.
\]
Assume for contradiction that $z_\eta(\tau)=0$ for some $\tau>0$.
By continuity, there is a $\delta\in(0,\tau)$ such that $z_\eta\le\theta$ on $[\tau-\delta,\tau]$.
Integrating the preceding derivative bound on this interval gives $0=z_\eta(\tau)\ge z_\eta(\tau-\delta)+h\delta/2>0$, a contradiction.
Hence $z_\eta(u)>0$ for every $u>0$.

As for the integrability of $m'/\sqrt{z_\eta}$, the only possible singularity is at $u=0$.
Let $\varepsilon \coloneqq \min\{M,\theta/h\}$.
If $0<u\le\varepsilon$ and $0\le s\le u$, then the upper bound gives $z_\eta(s)\le hs\le\theta$.
Thus $z_\eta'\ge h/2$ throughout $[0,u]$ in the almost-everywhere sense, and integration gives $z_\eta(u)\ge hu/2$.
Consequently, $m'(u)/\sqrt{z_\eta(u)}\le K\sqrt{2/(hu)}$ for almost every $0<u\le\varepsilon$.
The right-hand side is integrable at zero.
If $\varepsilon<M$, continuity and strict positivity give a positive minimum of $z_\eta$ on $[\varepsilon,M]$, so the ratio is bounded there almost everywhere.
This proves integrability on $[0,M]$; its value at the single point zero is immaterial.

For uniqueness, let $z_1$ and $z_2$ be two nonnegative solutions.
Subtracting the equations gives $z_1'-z_2'=-\alpha m'(u)\bigl(\sqrt{z_1(u)}-\sqrt{z_2(u)}\bigr)$.
Using $z_1-z_2=(\sqrt{z_1}-\sqrt{z_2})(\sqrt{z_1}+\sqrt{z_2})$, the chain rule yields the following derivative identity.
\[
\bigl((z_1-z_2)^2\bigr)'
=-2\alpha m'(u)\bigl(\sqrt{z_1(u)}-\sqrt{z_2(u)}\bigr)^2
\bigl(\sqrt{z_1(u)}+\sqrt{z_2(u)}\bigr)\le0.
\]
Thus $(z_1-z_2)^2$ is nonincreasing.
It is nonnegative and starts from zero, so it must be zero identically.
This proves uniqueness.
\hfill
\Halmos

\medskip

\noindent
\textbf{Proof of \thmref{rn:thm:continuous}{Theorem}:} The case $x=M$ is immediate, because the only distribution supported on $[0,M-x]=\{0\}$ is $\delta_0$ and both terms in the objective are zero.
Hence fix $x<M$ and write $L \coloneqq M-x>0$.
All integrals below are over $[0,L]$ unless their limits are shown explicitly.

We first derive a lower bound for $J_x(Q;\eta)$ by completing the square in its integral representation.
Let $Q\in\mathcal Q_x$ have cdf $F$, and define $C_Q \coloneqq W_Q-\Delta$.
Because $\Delta$ is absolutely continuous with derivative $m'$, the Stieltjes product rule gives $\mathrm{d}(F\Delta)=\Delta\,\mathrm{d}Q+Fm'\,\mathrm{d}u$.
On the other hand, the definition of $W_Q$ gives $\mathrm{d}(F\Delta)=W_Q\,\mathrm{d}Q$.
Subtracting the common term $\Delta\,\mathrm{d}Q$ yields the measure identity $C_Q\,\mathrm{d}Q=Fm'\,\mathrm{d}u$.
Applying the product rule to $F\Delta^2$ and using $F(L)=1$ and $\Delta(0)=0$ gives $\Delta(L)^2=\int\Delta^2\,\mathrm{d}Q+2\int\Delta Fm'\,\mathrm{d}u$.
With these equalities, it follows that
\[
\begin{aligned}
\int C_Q^2\,\mathrm{d}Q
&=\int W_Q^2\,\mathrm{d}Q-\int\Delta^2\,\mathrm{d}Q-2\int\Delta C_Q\,\mathrm{d}Q\\
&=\int W_Q^2\,\mathrm{d}Q-\int\Delta^2\,\mathrm{d}Q-2\int\Delta Fm'\,\mathrm{d}u=\int W_Q^2\,\mathrm{d}Q-\Delta(L)^2=V_x(Q).
\end{aligned}
\]
Thus the worst-case variance is exactly the squared $L^2(Q)$ norm of the excess correction $C_Q$.

Define $b_\eta(u) \coloneqq \sqrt{2z_\eta(u)/\eta}$, with $b_\eta(0) \coloneqq 0$.
The ODE in Eq.~\eqref{rn:eq:ode} is equivalent to $\eta b_\eta(u)m'(u)=h-z_\eta'(u)$.
Also, Stieltjes integration by parts, together with $z_\eta(0)=0$ and $F(L)=1$, gives $\int z_\eta\,\mathrm{d}Q=z_\eta(L)-\int F(u)z_\eta'(u)\,\mathrm{d}u$.
Using the preceding measure identity and ODE identity gives the following square expansion.
\[
\begin{aligned}
\frac\eta2\int(C_Q-b_\eta)^2\,\mathrm{d}Q
&=\frac\eta2\int C_Q^2\,\mathrm{d}Q-\eta\int b_\eta Fm'\,\mathrm{d}u+\int z_\eta\,\mathrm{d}Q\\
&=\frac\eta2V_x(Q)-\int F\cdot(h-z_\eta')\,\mathrm{d}u+z_\eta(L)-\int Fz_\eta'\,\mathrm{d}u\\
&=\frac\eta2V_x(Q)+z_\eta(L)-h\int F(u)\,\mathrm{d}u.
\end{aligned}
\]
For a random variable supported on $[0,L]$, the tail-integral identity gives $\mathbb E_Q[\tilde u]=\int\{1-F(u)\}\,\mathrm{d}u=L-\int F(u)\,\mathrm{d}u$.
Substitution produces the following exact square-completion identity.
\begin{equation}
\label{rn:eq:square-completion}
J_x(Q;\eta)=hL-z_\eta(L)+\frac\eta2\int\{C_Q(u)-b_\eta(u)\}^2Q(\mathrm{d}u).
\end{equation}
The square is nonnegative, so every admissible distribution has objective value at least $hL-z_\eta(L)$.

Now, we show that the lower bound is achieved by the distribution in the statement.
By Lemma~\ref{rn:lem:ode}, $b_\eta(u)>0$ for $u>0$ and $m'(u)/b_\eta(u)=\sqrt{\eta/2}\,m'(u)/\sqrt{z_\eta(u)}$ is integrable on $[0,L]$.
Hence Eq.~\eqref{rn:eq:continuous-cdf} is well defined.
Its exponent is nonpositive, so $0<F_x^\star(u)\le1$, it is nondecreasing, and $F_x^\star(L)=1$.
Moreover, it is absolutely continuous on $(0,L]$ and satisfies $(F_x^\star)'(u)=F_x^\star(u)m'(u)/b_\eta(u)$ for almost every $u>0$.
Extending this cdf by zero to the left of zero shows that it defines a probability measure $Q_x^\star$ with an atom of size $F_x^\star(0)$ at zero and the displayed density on $(0,L]$.
Because $b_\eta(0)=0$, the atom at zero does not affect the following measure identity on $(0,L]$.
\[
b_\eta(u)Q_x^\star(\mathrm{d}u)=F_x^\star(u)m'(u)\,\mathrm{d}u.
\]
The product rule now gives $\mathrm{d}(F_x^\star\Delta)=(\Delta+b_\eta)\,\mathrm{d}Q_x^\star$.
Thus the Radon--Nikodym correction is $W_{Q_x^\star}=\Delta+b_\eta$, with value zero chosen at the atom at zero.
The bounds $z_\eta(u)\le hL$ and $\Delta(u)\le\Delta(L)$ show that this correction is bounded, and hence belongs to $L^2(Q_x^\star)$.
Therefore $Q_x^\star$ is admissible.
For this distribution, the preceding measure identity says exactly that $C_{Q_x^\star}=b_\eta$ $Q_x^\star$-almost surely.
The square in Eq.~\eqref{rn:eq:square-completion} vanishes, so the lower bound is attained and the optimal value is $hL-z_\eta(L)$.

For uniqueness, let $Q$ be any optimizer.
Equality in Eq.~\eqref{rn:eq:square-completion} forces $C_Q=b_\eta$ $Q$-almost surely.
Combining this equality with the excess-correction measure identity gives $b_\eta\,\mathrm{d}Q=Fm'\,\mathrm{d}u$.
Since $b_\eta$ is strictly positive on $(0,L]$, the restriction of $Q$ to this interval is absolutely continuous, and its cdf satisfies $F'=Fm'/b_\eta$ almost everywhere.
For any $u>0$, solving this scalar linear equation backward from the terminal condition $F(L)=1$ gives $F(u)=\exp\{-\int_u^Lm'(v)/b_\eta(v)\,\mathrm{d}v\}=F_x^\star(u)$.
Right continuity then gives $F(0)=F_x^\star(0)$ and fixes the atom at zero.
Thus $Q=Q_x^\star$, proving uniqueness.
\hfill
\Halmos

\medskip

\noindent
\textbf{Numerical approximation and objective error.}
We construct an approximation of $z_\eta$ and show that the resulting distribution remains admissible, with the one-period objective error bounded by the uniform approximation error.
Fix $\eta>0$ and write $\alpha\coloneqq\sqrt{2\eta}$ and $K\coloneqq\|m'\|_\infty$.

We first control the variation of $\sqrt{z_\eta}$, which will be used to bound the discretization error.
Choose $c>0$ sufficiently small that $c+\alpha K\sqrt{cM}\le h$.
Wherever $z_\eta(u)<cu$, the ODE gives
$z_\eta'(u)-c\ge h-\alpha K\sqrt{cu}-c\ge0$ almost everywhere.
Thus, the negative part of $z_\eta(u)-cu$ cannot increase from its initial value zero, proving $z_\eta(u)\ge cu$.
Together with $z_\eta(u)\le hu$, the ODE also gives
$|z_\eta'(u)|\le B\coloneqq h+\alpha K\sqrt{hM}$ almost everywhere.
Away from zero, the chain rule therefore gives
$|(\sqrt{z_\eta(u)})'|\le B/(2\sqrt{cu})$.
This bound is integrable at zero, so the chain-rule identity extends to $[0,M]$.
Consequently, $\sqrt{z_\eta}$ is absolutely continuous and
\(\int_0^M
|\bigl(\sqrt{z_\eta(u)}\bigr)'|\,\mathrm{d}u
\le
\int_0^M\frac{B}{2\sqrt{cu}}\,\mathrm{d}u
=B\sqrt{M/c}.\)

Divide $[0,M]$ into $N$ equal intervals, with
$\delta\coloneqq M/N$ and $u_i\coloneqq i\delta$.
On each interval, use the average coefficient
$a_i\coloneqq\{m(u_i)-m(u_{i-1})\}/\delta$.
Starting from $Z_0=0$, compute
\[
Z_i+\alpha\delta a_i\sqrt{Z_i}
=Z_{i-1}+h\delta,
\qquad i=1,\ldots,N.
\]
Because $a_i\ge0$, the left-hand side is strictly increasing from zero to infinity.
Each step therefore has a unique positive solution, obtained by solving a quadratic equation for $\sqrt{Z_i}$.
Also, $Z_i\le Z_{i-1}+h\delta$, so induction gives $0<Z_i\le hu_i$.
Define the piecewise-constant approximation by
$\widehat z_\eta(0)=0$ and
$\widehat z_\eta(u)=Z_i$ for $u\in(u_{i-1},u_i]$.

To bound the error, substitute the exact values $z_\eta(u_{i-1})$ and $z_\eta(u_i)$ into the discrete update.
Integrating the ODE over the interval shows that the residual is
$\alpha\int_{u_{i-1}}^{u_i}
m'(v)\{\sqrt{z_\eta(u_i)}-\sqrt{z_\eta(v)}\}\,\mathrm{d}v$.
The function $s\mapsto s+\alpha\delta a_i\sqrt{s}$ increases by at least the change in $s$.
Hence, a perturbation of the right-hand side changes its solution by no more than the size of that perturbation.
Bounding the residual by the variation of $\sqrt{z_\eta}$ on the interval therefore gives
\(|Z_i-z_\eta(u_i)| \leq 
|Z_{i-1}-z_\eta(u_{i-1})| 
+\alpha K\delta
\int_{u_{i-1}}^{u_i}
|\bigl(\sqrt{z_\eta(v)}\bigr)'|\,\mathrm{d}v.\)
Summing this inequality bounds the error at the grid points.
Within each interval, the exact solution varies by at most $B\delta$.
Combining these two bounds yields
\[
\|\widehat z_\eta-z_\eta\|_\infty
\le
B\delta\{1+\alpha K\sqrt{M/c}\}
=\frac{C_\eta}{N},
\qquad
C_\eta\coloneqq MB\{1+\alpha K\sqrt{M/c}\}.
\]

We next use $\widehat z_\eta$ to construct a distribution, while retaining the original marginal shortage cost $m$.
For a virtual order $x<M$, set $L\coloneqq M-x$ and
$\widehat b_\eta\coloneqq\sqrt{2\widehat z_\eta/\eta}$.
Define $\widehat Q_x$ through its cdf
\[
\widehat F_x(u)
\coloneqq
\exp\Big\{
-\int_u^L\frac{m'(v)}{\widehat b_\eta(v)}\,\mathrm{d}v
\Big\},
\qquad 0\le u\le L.
\]
For each fixed grid, $\widehat b_\eta$ is bounded away from zero on $(0,M]$, so the integral is finite.
The distribution has mass $\widehat F_x(0)$ at zero and density
$\widehat F_xm'/\widehat b_\eta$ on $(0,L]$.
Substituting this density into Eq.~\eqref{rn:eq:rn-correction} gives
$W_{\widehat Q_x}(u)=\Delta(u)+\widehat b_\eta(u)$ for $u>0$, with
$W_{\widehat Q_x}(0)=0$.
This correction is bounded, so $\widehat Q_x$ is admissible for the original shortage cost.

Finally, Eq.~\eqref{rn:eq:square-completion} shows how the approximation error affects the objective.
Using $(\sqrt a-\sqrt b)^2\le|a-b|$ for $a,b\ge0$, we obtain
\[
\begin{aligned}
0\le J_x(\widehat Q_x;\eta)-J_x^\star(\eta)
&=
\int_{[0,L]}
\{\sqrt{\widehat z_\eta(u)}-\sqrt{z_\eta(u)}\}^2
\,\widehat Q_x(\mathrm{d}u) \le
\|\widehat z_\eta-z_\eta\|_\infty
\le\frac{C_\eta}{N}.
\end{aligned}
\]
For $x=M$, take $\widehat Q_M=\delta_0$.
Since $C_\eta$ does not depend on $x$, choosing $N\ge C_\eta/\epsilon$ gives objective error at most $\epsilon$ uniformly over all virtual orders.

Both the cell averages and the cdf integrals can be evaluated using increments of $m$, because $\widehat b_\eta$ is constant within each grid interval.
Thus, continuity of $m'$ is not required.
For fixed $\eta$ and problem parameters, $O(\epsilon^{-1})$ grid intervals suffice.
This count excludes the cost of evaluating $m$ and inverting the resulting cdf.
\hfill
\Halmos

\medskip

\noindent
\textbf{Verification of \thmref{rn:ex:quadratic-design}{Example}:} The function $w\mapsto-w-\log(1-w)$ is continuous and strictly increasing from zero to infinity on $[0,1)$, so the scalar equation has a unique solution for each $u\ge0$.
For $u>0$, implicit differentiation gives $w_\eta'(u)=(\eta\gamma^2/h)\{1-w_\eta(u)\}/w_\eta(u)$.
Hence $z_\eta(u) \coloneqq h^2w_\eta(u)^2/(2\eta\gamma^2)$ satisfies $z_\eta'(u)=h\cdot\{1-w_\eta(u)\}=h-\sqrt{2\eta}\,\gamma\sqrt{z_\eta(u)}$ for $u>0$.
Since $w_\eta(0)=0$ and $0\le z_\eta'(u)\le h$, this solution extends absolutely continuously to zero with $z_\eta(0)=0$.
Lemma~\ref{rn:lem:ode} therefore identifies it with the solution used in Theorem~\ref{rn:thm:continuous}.
Also, $\{\log(1-w_\eta(u))\}'=-\eta\gamma^2/\{h w_\eta(u)\}$ for $u>0$.
Substituting $m'=\gamma$ and the formula for $z_\eta$ into Eq.~\eqref{rn:eq:continuous-cdf} gives $F_x^\star(u)=\{1-w_\eta(M-x)\}/\{1-w_\eta(u)\}$, with the value at zero determined by continuity.
The correction, optimal value, and mass at zero then follow directly from Theorem~\ref{rn:thm:continuous}.
\hfill
\Halmos


\section{Proofs of results in Section~\ref{rn:sec:regret}}
\label{rn:app:regret}

\subsection{Distribution-free regret and the general minimax rate}
\label{rn:app:local-bound}
\label{rn:app:general-regret}
\label{rn:app:lower}
\label{rn:app:information}
\label{rn:app:general-lower}

We first prove an auxiliary {one-period objective}~\eqref{rn:eq:local-objective} bound and use it to establish Theorem~\ref{rn:thm:general-regret}.
We then establish the statistical-testing inequalities needed for the general minimax lower bound in Theorem~\ref{rn:thm:general-lower}.

\begin{lemma}
\label{rn:lem:local-bound}
For every $x\in[0,M]$ and $\eta>0$, the optimized {one-period} objective satisfies Eq.~\eqref{rn:eq:local-bound}.
There is an admissible distribution $Q$ with $J_x(Q;\eta)\le\sqrt{2h\eta}\,B_{\nu,x}$.
\end{lemma}

\noindent
\textbf{Proof of \thmref{rn:lem:local-bound}{Lemma}:}
Set $L \coloneqq M-x$.
If $L=0$, then $Q=\delta_0$ is the only feasible distribution, its objective is zero, and all claims follow immediately.
Assume henceforth that $L>0$, and recall that $\Delta(u)=\nu((0,u])$ and $B_{\nu,x}=\int_{(0,L]}\sqrt u\,\nu(\mathrm{d}u)$.
The objective in Eq.~\eqref{rn:eq:local-objective} is $J_x(Q;\eta)=\eta V_x(Q)/2+h\mathbb E_Q[\tilde u]$.
In this proof, for each $c>0$, we construct an admissible distribution $Q_c$ satisfying $\mathbb E_{Q_c}[\tilde u]\le B_{\nu,x}/c$ and $V_x(Q_c)\le cB_{\nu,x}$.
After verifying admissibility and these two bounds, we choose $c$ to balance the two terms in the objective and conclude the proof.

Let us start with the construction.
Fix $c>0$.
We choose the distribution so that the correction in Theorem~\ref{rn:thm:estimator} will take the form $W_c(u)=\Delta(u)+c\sqrt u$ for $u>0$.
We first define its cdf and check that it determines a probability distribution.
Write the restriction of $\nu$ to $(0,L]$ as $\nu^{\mathrm{na}}+\sum_{s\in A}a_s\delta_s$, where $\nu^{\mathrm{na}}$ has no atoms, $A$ is at most countable, and $a_s=\nu(\{s\})>0$.
Define, for $0<u\le L$,
\begin{equation}
\label{rn:eq:feasible-cdf}
\begin{aligned}
F_c(u) \coloneqq \exp\Big\{-\int_{(u,L]}
\frac{\nu^{\mathrm{na}}(\mathrm{d}v)}{c\sqrt v}\Big\}
\prod_{s\in A\cap(u,L]}
\Big(1+\frac{a_s}{c\sqrt s}\Big)^{-1},
\quad \text{and} \quad
F_c(0)  \coloneqq  \lim_{u\downarrow0}F_c(u).
\end{aligned}
\end{equation}
The exponential factor accounts for the nonatomic part of $\nu$, while each factor in the product accounts for one atom.
For fixed $u>0$, $\log(1+a_s/(c\sqrt s)) \leq a_s/(c\sqrt s)$ over atoms $s>u$ and $\int_{(u,L]}
\frac{\nu^{\mathrm{na}}(\mathrm{d}v)}{c\sqrt v} \leq \nu((u,L])/(c\sqrt u)<\infty$.
Thus the product converges and $0<F_c(u)\le1$ for $u>0$.
The integral and the sum of logarithms are right continuous and decrease as $u$ increases, so $F_c$ is nondecreasing and right continuous.
Together with $F_c(L)=1$ and the definition of $F_c(0)$, these properties show that $F_c$ is the cdf of a probability distribution $Q_c$ on $[0,L]$.

Now, we verify admissibility of $Q_c$ and identify the correction $W_c$. 
By Theorem~\ref{rn:thm:estimator} and the definition in Section~\ref{rn:sec:admissible}, admissibility requires $F_c(\cdot^-)\nu\ll Q_c$ and a correction in $L^2(Q_c)$.
By the correction formula in Eq.~\eqref{rn:eq:rn-correction}, an unbiased correction must satisfy $\int_B(W_c(u)-\Delta(u))\,Q_c(\mathrm{d}u)=\int_B F_c(u^-)\,\nu(\mathrm{d}u)$ for every Borel set $B\subseteq(0,L]$.
To obtain $W_c(u)-\Delta(u)=c\sqrt u$, we therefore establish the measure identity
\begin{equation}
\label{rn:eq:feasible-measure}
c\sqrt u\,Q_c(\mathrm{d}u)=F_c(u^-)\nu(\mathrm{d}u),
\qquad 0<u\le L.
\end{equation}
We then check that the resulting correction is square integrable.

Fix $\delta\in(0,L)$ and work first on $[\delta,L]$, where $1/\sqrt u$ is bounded.
Set $H_c(u) \coloneqq \log F_c(u)$ and $b_s \coloneqq \log(1+a_s/(c\sqrt s))$.
Since $F_c(\delta)>0$ and $F_c$ is nondecreasing, $H_c$ is finite and has bounded variation on this interval.
For $\delta\le a<b\le L$, Eq.~\eqref{rn:eq:feasible-cdf} gives $H_c(b)-H_c(a)=\int_{(a,b]}\nu^{\mathrm{na}}(\mathrm{d}u)/(c\sqrt u)+\sum_{s\in A\cap(a,b]}b_s$.
Thus the nonatomic part of the Stieltjes measure of $H_c$ is $\nu^{\mathrm{na}}(\mathrm{d}u)/(c\sqrt u)$.
Let $Q_c^{\mathrm{na}}$ denote the nonatomic part of $Q_c$.
Applying the chain rule for the nonatomic parts of functions of bounded variation to $F_c=\exp(H_c)$ gives $Q_c^{\mathrm{na}}(\mathrm{d}u)=F_c(u)\nu^{\mathrm{na}}(\mathrm{d}u)/(c\sqrt u)$ on $(\delta,L]$ \citep[Theorem~3.96]{ambrosio2000functions}.
This rule applies to Stieltjes measures, so $\nu^{\mathrm{na}}$ need not have a density with respect to Lebesgue measure.

At an atom $s\in A\cap(\delta,L]$, we have $H_c(s)-H_c(s^-)=b_s$.
Exponentiating gives $F_c(s)=F_c(s^-)(1+a_s/(c\sqrt s))$, and hence $Q_c(\{s\})=F_c(s^-)a_s/(c\sqrt s)$.
Since $A$ is countable and $\nu^{\mathrm{na}}$ has no atoms, $F_c(u)=F_c(u^-)$ for $\nu^{\mathrm{na}}$-almost every $u$ in this interval.
Combining the nonatomic and atomic parts proves Eq.~\eqref{rn:eq:feasible-measure} on $(\delta,L]$.
For any Borel set $B\subseteq(0,L]$, apply this identity to $B\cap(\delta,L]$ and let $\delta\downarrow0$.
Continuity from below proves Eq.~\eqref{rn:eq:feasible-measure} on all of $(0,L]$.

Equation~\eqref{rn:eq:feasible-measure} shows that $F_c(\cdot^-)\nu\ll Q_c$ on $(0,L]$, with Radon--Nikodym derivative $c\sqrt u$.
Theorem~\ref{rn:thm:estimator} therefore gives the unbiased correction $W_c(u)=\Delta(u)+c\sqrt u$ for $0<u\le L$, with $W_c(0)=0$.
Since $0\le W_c(u)\le\Delta(L)+c\sqrt L$ on $[0,L]$, we have $W_c\in L^2(Q_c)$.
Both requirements for admissibility are therefore satisfied, so $Q_c\in\mathcal Q_x$.

Next, we bound expected additional inventory and subgradient estimator variance under the constructed distribution $Q_c$.
Multiplying Eq.~\eqref{rn:eq:feasible-measure} by $\sqrt u/c$ and integrating gives $\mathbb E_{Q_c}[\tilde u]=c^{-1}\int_{(0,L]}\sqrt u\,F_c(u^-)\nu(\mathrm{d}u)$.
Since $F_c(u^-)\le1$, this proves $\mathbb E_{Q_c}[\tilde u]\le B_{\nu,x}/c$.

For the variance, Eq.~\eqref{rn:eq:variance} gives $V_x(Q_c)=\int_{(0,L]}W_c(u)^2Q_c(\mathrm{d}u)-\Delta(L)^2$.
We expand $W_c(u)^2=(\Delta(u)+c\sqrt u)^2$ and use the Stieltjes product rule to account for the terms involving $\Delta$.
The chain rule for functions of bounded variation \citep[Theorem~3.96]{ambrosio2000functions} gives $2\Delta(u)\nu^{\mathrm{na}}(\mathrm{d}u)$ as the nonatomic part of the Stieltjes measure of $\Delta^2$.
At an atom $s$, the jump is $\Delta(s)^2-\Delta(s^-)^2=2\Delta(s)a_s-a_s^2$, because $\Delta(s)-\Delta(s^-)=a_s$.
It follows that $\mathrm{d}(\Delta^2)(u)=2\Delta(u)\nu(\mathrm{d}u)-\sum_{s\in A}a_s^2\delta_s(\mathrm{d}u)$.

The Stieltjes product rule for $F_c\Delta^2$ remains valid when both factors jump \citep[Example~3.97]{ambrosio2000functions}.
Integrating it over $(0,L]$ gives $\Delta(L)^2=\int_{(0,L]}\Delta(u)^2Q_c(\mathrm{d}u)+\int_{(0,L]}F_c(u^-)\,\mathrm{d}(\Delta^2)(u)$, since $F_c(L)=1$ and $\Delta(0)=0$.
Substituting this expression for $\Delta(L)^2$ into the variance formula in Eq.~\eqref{rn:eq:variance} and expanding $W_c(u)^2=(\Delta(u)+c\sqrt u)^2$ cancels the common term $\int_{(0,L]}\Delta(u)^2Q_c(\mathrm{d}u)$.
This gives the first equality below.
The second equality follows from substituting $\mathrm{d}(\Delta^2)(u)=2\Delta(u)\nu(\mathrm{d}u)-\sum_{s\in A}a_s^2\delta_s(\mathrm{d}u)$ into the last integral and using Eq.~\eqref{rn:eq:feasible-measure} to replace $c\sqrt u\,Q_c(\mathrm{d}u)$ by $F_c(u^-)\nu(\mathrm{d}u)$ in the first integral, and rearranging terms.
\[
\begin{aligned}
V_x(Q_c)
&=2c\int_{(0,L]}\sqrt u\,\Delta(u)Q_c(\mathrm{d}u)
+c^2\int_{(0,L]}u\,Q_c(\mathrm{d}u) - \int_{(0,L]}F_c(u^-)\,\mathrm{d}(\Delta^2)(u)\\
&=c^2\int_{(0,L]}u\,Q_c(\mathrm{d}u)
+\sum_{s\in A}F_c(s^-)a_s^2.
\end{aligned}
\]
The sum is finite since $\sum_{s\in A}F_c(s^-)a_s^2\le\sum_{s\in A}a_s^2\le\nu((0,L])^2$.

We now express each of the two remaining terms as an integral against $\nu$.
For the first term, multiply Eq.~\eqref{rn:eq:feasible-measure} by $c\sqrt u$ and integrate to obtain $c^2\int_{(0,L]}u\,Q_c(\mathrm{d}u)=c\int_{(0,L]}\sqrt u\,F_c(u^-)\nu(\mathrm{d}u)$.
For the sum, apply the same measure identity to the set $\{s\}$, where $s\in A$, to get $c\sqrt s\,Q_c(\{s\})=F_c(s^-)a_s$.
Substituting $Q_c(\{s\})=F_c(s)-F_c(s^-)$ and multiplying by $a_s$ gives $F_c(s^-)a_s^2=c\sqrt s\,[F_c(s)-F_c(s^-)]a_s$.
Since $F_c(u)=F_c(u^-)$ for $\nu^{\mathrm{na}}$-almost every $u$, summing over atoms gives $\sum_{s\in A}F_c(s^-)a_s^2=c\int_{(0,L]}\sqrt u\,[F_c(u)-F_c(u^-)]\nu(\mathrm{d}u)$.
Adding the two integrals combines $F_c(u^-)$ and $F_c(u)-F_c(u^-)$ into $F_c(u)$.
Therefore $V_x(Q_c)=c\int_{(0,L]}\sqrt u\,F_c(u)\nu(\mathrm{d}u)\le cB_{\nu,x}$, where the inequality uses $F_c(u)\le1$.

Finally, let us choose $c$ to conclude the proof.
The bounds above and the objective definition in Eq.~\eqref{rn:eq:local-objective} give $J_x(Q_c;\eta)\le(\eta c/2+h/c)B_{\nu,x}$.
Choose $c=\sqrt{2h/\eta}$, so that $\eta c/2=h/c$.
The admissible distribution $Q_c$ then satisfies $J_x(Q_c;\eta)\le\sqrt{2h\eta}\,B_{\nu,x}$, proving the existence claim.
Since $J_x^\star(\eta)\le J_x(Q_c;\eta)$ and $B_{\nu,x}\le B_\nu$, Eq.~\eqref{rn:eq:local-bound} follows.
\hfill
\Halmos

\medskip

\noindent
\textbf{Proof of \thmref{rn:thm:general-regret}{Theorem}:} Lemma~\ref{rn:lem:local-bound} proves the first assertion.
Conditional on $\mathcal F_{t-1}$, the virtual order $\tilde x_t$ and the selected distribution $Q_t$ are fixed.
The allowance $\epsilon_t$ can include error from approximating the one-period objective.
The assumed one-period objective bound in the statement and $B_{\nu,\tilde x_t}\le B_\nu$ therefore give $J_{\tilde x_t}(Q_t;\eta_t)\le\sqrt{2h\eta_t}\,B_\nu+\epsilon_t$ almost surely.
Substituting this bound into Eq.~\eqref{rn:eq:physical-regret} proves Eq.~\eqref{rn:eq:general-regret} under the demand model in Assumption~\ref{ass:dependence}.
For $\eta_t=\eta_0t^{-2/3}$, we have $\eta_T^{-1}=\eta_0^{-1}T^{2/3}$, $\sum_{t=1}^T\eta_t\le3\eta_0T^{1/3}$, and $\sum_{t=1}^T\sqrt{\eta_t}\le(3/2)\sqrt{\eta_0}T^{2/3}$.
Substituting these bounds into Eq.~\eqref{rn:eq:general-regret} gives
\begin{equation}\label{rn:eq:anytime-regret}
    \mathbb E[\tilde r_T]
\le\Big\{\frac{3M^2}{4\eta_0}+\frac94\sqrt{2h\eta_0}\,B_\nu\Big\}T^{2/3}
+\frac32G^2\eta_0T^{1/3}+\frac32\sum_{t=1}^T\epsilon_t.
\end{equation}
The asserted $O(T^{2/3})$ conclusion follows immediately when $\sum_{t=1}^T\epsilon_t=O(T^{2/3})$.
\hfill
\Halmos

\medskip

We first establish two auxiliary inequalities for the proofs of Theorems~\ref{rn:thm:general-lower} and~\ref{rn:thm:accelerated-lower}: one bounds the information obtained from censored sales, and the other relates this information to regret.
Recall the normalized setting $M=3$, $h=3/2$, and the demand pair introduced in Section~\ref{rn:sec:general-regret}.
For $0<\varepsilon\le1/8$ and $\sigma\in\{+1,-1\}$, demands are iid with distribution
\begin{equation}
\label{rn:eq:hard-pair}
P_\sigma(\{0\})=\frac12,\qquad
P_\sigma(\{2\})=\frac14-\sigma\varepsilon,\qquad
P_\sigma(\{3\})=\frac14+\sigma\varepsilon.
\end{equation}

Fix a known shortage cost $\psi$, common to both demand laws.
Write $L_\sigma(q)\coloneqq\mathbb E_{P_\sigma}[\ell(q,\tilde d)]$ and choose $q_\sigma^\star\in\arg\min_{q\in[0,3]}L_\sigma(q)$.
Let $\pi$ be any nonanticipating policy with conditional order distribution $\pi_t(\mathrm{d}q\mid\tilde h_{t-1})$, where
$\tilde h_{t-1}\coloneqq(\tilde q_1,\tilde s_1,\ldots,\tilde q_{t-1},\tilde s_{t-1})$
is the observed history.
The policy may depend on $T$, $\psi$, and any supplied curvature bound, but its conditional order distribution at a fixed history must be the same under both demand laws.

Write $\mathbb E_\sigma^\pi$ and $\mathbb P_\sigma^\pi$ for expectation and probability under policy $\pi$ and demand law $P_\sigma$, and define the pseudo-regret by
$\mathcal E_\sigma^\pi(T)\coloneqq
\mathbb E_\sigma^\pi[\sum_{t=1}^T\{L_\sigma(\tilde q_t)-L_\sigma(q_\sigma^\star)\}]$.
As discussed in Section~\ref{rn:sec:general-regret}, expected regret relative to the best fixed order in hindsight is at least as large:
\begin{equation}
\label{rn:eq:population-lower}
\mathbb E_\sigma^\pi[\tilde r_T]\ge\mathcal E_\sigma^\pi(T).
\end{equation}
It therefore suffices to establish lower bounds for pseudo-regret on these iid instances.

For every fixed order $q\le2$, the sales kernels under $P_+$ and $P_-$ coincide: demand two and demand three are both censored at $q$.
Thus only orders above two provide information for distinguishing the demand laws.
Such an order reveals whether the demand was zero, two, or three, even when sales equal the order.
The next lemma bounds the resulting KL divergence by the expected number of such orders.

\begin{lemma}
\label{rn:lem:kl}
Let $\mathbb P_\sigma^\pi$ also denote the law of the history $(\tilde q_1,\tilde s_1,\ldots,\tilde q_T,\tilde s_T)$ generated by policy $\pi$ under $P_\sigma$.
Then
\begin{equation}
\label{rn:eq:kl-bound}
\operatorname{KL}(\mathbb P_+^\pi\Vert\mathbb P_-^\pi)
\le24\varepsilon^2\sum_{t=1}^T\mathbb P_+^\pi(\tilde q_t>2).
\end{equation}
\end{lemma}

\noindent
\textbf{Proof of \thmref{rn:lem:kl}{Lemma}:} We first compute the information from one order and its sales observation, and then sum the conditional contributions over time.
Let $\tilde h_{t-1} \coloneqq (\tilde q_1,\tilde s_1,\ldots,\tilde q_{t-1},\tilde s_{t-1})$ be the observable history before period $t$.
For a fixed physical order $q$, let $K_\sigma(\cdot\mid q)$ denote the law of the sales observation $\tilde s=\min\{\tilde d,q\}$ under $\tilde d\sim P_\sigma$.
Conditional on $\tilde h_{t-1}$, the joint law of the next order and sales observation factors as $\mathbb P_\sigma^\pi(\mathrm{d}q,\mathrm{d}s\mid\tilde h_{t-1})=\pi_t(\mathrm{d}q\mid\tilde h_{t-1})K_\sigma(\mathrm{d}s\mid q)$.
The policy kernel is the same under the two demand laws.
Therefore it contributes no likelihood ratio, and the conditional relative entropy has the following representation.
\[
\operatorname{KL}\bigl(\mathbb P_+^\pi(\cdot\mid\tilde h_{t-1})\Vert\mathbb P_-^\pi(\cdot\mid\tilde h_{t-1})\bigr)
=\int\operatorname{KL}\bigl(K_+(\cdot\mid q)\Vert K_-(\cdot\mid q)\bigr)\pi_t(\mathrm{d}q\mid\tilde h_{t-1}).
\]

We now compute the inner divergence.
If $q\le2$, demands two and three both produce the same censored sale, so $K_+(\cdot\mid q)=K_-(\cdot\mid q)$ and the divergence is zero.
If $q>2$, sales zero has probability $1/2$ under both laws, while the probabilities of sales two and sales $q$ are interchanged.
Thus $\operatorname{KL}(K_+(\cdot\mid q)\Vert K_-(\cdot\mid q))=2\varepsilon\log\{(1/4+\varepsilon)/(1/4-\varepsilon)\}$.
To bound this expression, write $y \coloneqq 4\varepsilon\le1/2$.
Then $\log\{(1+y)/(1-y)\}=\int_0^y2/(1-v^2)\,\mathrm{d}v\le(8/3)y\le3y$, giving the upper bound $24\varepsilon^2$.
Consequently, the conditional divergence is at most $24\varepsilon^2\pi_t((2,3]\mid\tilde h_{t-1})$.

For the second step, apply the chain rule for relative entropy to the sequence $(\tilde q_t,\tilde s_t)_{t=1}^T$.
Because the divergence is directed from $\mathbb P_+^\pi$ to $\mathbb P_-^\pi$, the conditional divergences are averaged under $\mathbb P_+^\pi$.
The chain rule and tower property therefore give the following bound.
\[
\begin{aligned}
\operatorname{KL}(\mathbb P_+^\pi\Vert\mathbb P_-^\pi)
&=\sum_{t=1}^T\mathbb E_+^\pi\!\Big[\operatorname{KL}\bigl(\mathbb P_+^\pi(\cdot\mid\tilde h_{t-1})\Vert\mathbb P_-^\pi(\cdot\mid\tilde h_{t-1})\bigr)\Big]\\
&\le24\varepsilon^2\sum_{t=1}^T\mathbb E_+^\pi[\pi_t((2,3]\mid\tilde h_{t-1})]
=24\varepsilon^2\sum_{t=1}^T\mathbb P_+^\pi(\tilde q_t>2).
\end{aligned}
\]
This proves Eq.~\eqref{rn:eq:kl-bound}.
\hfill
\Halmos

\medskip

The next lemma turns this information bound into a reusable regret inequality by relating the pseudo-regret to $\sum_{t=1}^T\mathbb P_+^\pi(\tilde q_t>2)$.

\medskip

\begin{lemma}
\label{rn:lem:testing}
Suppose that a Borel set $A\subseteq[0,3]$ and constants $c_H,\delta_\varepsilon>0$ satisfy $L_+(q)-L_+(q_+^\star)\ge c_H$ for $q>2$, $L_+(q)-L_+(q_+^\star)\ge\delta_\varepsilon$ for $q\notin A$, and $L_-(q)-L_-(q_-^\star)\ge\delta_\varepsilon$ for $q\in A$.
Then every nonanticipating policy satisfies
\begin{equation}
\label{rn:eq:testing-bound}
\max_{\sigma\in\{+,-\}}\mathcal E_\sigma^\pi(T)
\ge\min\left\{\frac{c_H}{192}\varepsilon^{-2},\frac38\delta_\varepsilon T\right\}.
\end{equation}
\end{lemma}

\noindent
\textbf{Proof of \thmref{rn:lem:testing}{Lemma}:} Let $I_T \coloneqq \sum_{t=1}^T\mathbb P_+^\pi(\tilde q_t>2)$, the expected number of informative orders under $P_+$.
There are two cases: sufficiently many informative orders directly incur regret, while too few leave the history laws close and force a substantial probability of choosing the wrong order region.
The threshold $1/(192\varepsilon^2)$ is chosen so that the second case gives a KL divergence below $1/8$.

First suppose that $I_T\ge(192\varepsilon^2)^{-1}$.
All population loss gaps are nonnegative, and the first assumption gives $\mathcal E_+^\pi(T)=\sum_{t=1}^T\mathbb E_+^\pi[L_+(\tilde q_t)-L_+(q_+^\star)]\ge c_HI_T$.
Therefore $\mathcal E_+^\pi(T)\ge c_H\varepsilon^{-2}/192$, which proves the first term in Eq.~\eqref{rn:eq:testing-bound}.

Now consider the second case, in which $I_T<(192\varepsilon^2)^{-1}$.
By Lemma~\ref{rn:lem:kl}, $\operatorname{KL}(\mathbb P_+^\pi\Vert\mathbb P_-^\pi)<1/8$.
Pinsker's inequality gives $\operatorname{TV}(\mathbb P_+^\pi,\mathbb P_-^\pi)\le\sqrt{\operatorname{KL}(\mathbb P_+^\pi\Vert\mathbb P_-^\pi)/2}<1/4$.
Here total variation is the supremum of the absolute probability difference over all events in the history space.

For each $t$, define the event $E_t \coloneqq \{\tilde q_t\in A\}$.
Because $\tilde q_t$ is part of the complete observable history, $E_t$ is measurable with respect to the history space on which the total-variation distance is evaluated.
Hence $|\mathbb P_+^\pi(E_t)-\mathbb P_-^\pi(E_t)|\le\operatorname{TV}(\mathbb P_+^\pi,\mathbb P_-^\pi)$.
It follows that the combined wrong-region probability satisfies the following bound.
\[
\begin{aligned}
\mathbb P_+^\pi(\tilde q_t\notin A)+\mathbb P_-^\pi(\tilde q_t\in A)
&=1-\mathbb P_+^\pi(E_t)+\mathbb P_-^\pi(E_t) 
\ge
1-|\mathbb P_+^\pi(E_t)-\mathbb P_-^\pi(E_t)|>\frac34.
\end{aligned}
\]
The event $\tilde q_t\notin A$ incurs the stated loss gap under $P_+$, whereas $\tilde q_t\in A$ incurs it under $P_-$.
Thus the preceding inequality prevents the policy from choosing a low-loss region with high probability under both demand laws.

The two wrong-region gap assumptions now imply the following combined regret bound.
\[
\mathcal E_+^\pi(T)+\mathcal E_-^\pi(T)
\ge\delta_\varepsilon\sum_{t=1}^T\{\mathbb P_+^\pi(\tilde q_t\notin A)+\mathbb P_-^\pi(\tilde q_t\in A)\}
>\frac34\delta_\varepsilon T.
\]
Thus, at least one of the two pseudo-regrets is greater than $3\delta_\varepsilon T/8$.
Combining the two cases proves Eq.~\eqref{rn:eq:testing-bound}.
\hfill
\Halmos

\medskip

We now apply these auxiliary results to the general lower bound in Theorem~\ref{rn:thm:general-lower}.
Let the piecewise-linear shortage {cost} be defined by $\psi(z) \coloneqq \int_0^z m(u)\,\mathrm{d}u$, where $m(z)$ is defined in Eq.~\eqref{rn:eq:mz}. 
For $T\ge512$, take $\varepsilon=T^{-1/3}$ in Eq.~\eqref{rn:eq:hard-pair}.
This specifies the normalized {shortage cost} and demand family used to prove Theorem~\ref{rn:thm:general-lower}.

\medskip

\noindent
\textbf{Proof of \thmref{rn:thm:general-lower}{Theorem}:} We proceed in three steps.
First, we determine the optimal orders in the normalized instance.
Second, we verify the three loss-gap conditions in Lemma~\ref{rn:lem:testing} and obtain the normalized regret bound.
Third, we scale demands, orders, and losses to arbitrary $M,h>0$.

For the first step, fix any nonanticipating policy and consider the expected losses under the demand pair in Eq.~\eqref{rn:eq:hard-pair}.
Away from demand points and breakpoints of $m$, differentiation under the finite expectation gives $L_\sigma'(q)=hP_\sigma([0,q))-\mathbb E_{P_\sigma}[m((\tilde d-q)_+)\mathbf 1\{\tilde d>q\}]$.

For $0<q<1$, only demand zero lies below the order, while shortages from demands two and three fall in the slope-one and slope-two regions of $\psi$.
For $1<q<2$, the corresponding shortage slopes are $1/2$ and $1$.
For $2<q<3$, demands zero and two lie below the order, and a demand of three has shortage slope $1/2$.
Substituting the masses from Eq.~\eqref{rn:eq:hard-pair} gives the following piecewise derivative.
\[
L_\sigma'(q)=\begin{cases}
-\sigma\varepsilon,&0<q<1,\\
3/8-\sigma\varepsilon/2,&1<q<2,\\
1-2\sigma\varepsilon,&2<q<3.
\end{cases}
\]
When $\varepsilon\le1/8$, the first derivative is negative for $\sigma=+$ and positive for $\sigma=-$, while all derivatives to the right of one are positive.
Convexity therefore gives $q_+^\star=1$ and $q_-^\star=0$.

For the second step, take $A \coloneqq [1/2,3]$.
The threshold $1/2$ lies between the two optimal orders, so choosing the opposite region leaves a loss gap under the corresponding demand law.
If $q\notin A$, then $q<1/2$ and the derivative of $L_+$ is $-\varepsilon$ on $(q,1)$.
Therefore $L_+(q)-L_+(1)=\varepsilon(1-q)\ge\varepsilon/2$.
Under $P_-$, all three displayed derivatives are positive, so $L_-$ is increasing.
Thus every $q\in A$ satisfies $L_-(q)-L_-(0)\ge L_-(1/2)-L_-(0)=\varepsilon/2$.
These are the two wrong-region inequalities with $\delta_\varepsilon=\varepsilon/2$.

To verify the informative-order gap, note that $L_+$ is increasing on $[1,3]$.
Every $q>2$ satisfies $L_+(q)-L_+(1)\ge L_+(2)-L_+(1)=3/8-\varepsilon/2\ge1/4$.
Hence the informative-order gap condition holds with $c_H=1/4$.
Applying Lemma~\ref{rn:lem:testing} gives the following lower bound.
\[
\max_{\sigma\in\{+,-\}}\mathcal E_\sigma^\pi(T)
\ge\min\left\{\frac1{768}\varepsilon^{-2},\frac3{16}\varepsilon T\right\}.
\]
The first term is the cost of obtaining enough informative observations, and the second is the loss from selecting the wrong region when the observations do not distinguish the two laws.
With $\varepsilon=T^{-1/3}$, both terms have order $T^{2/3}$ and the first is the smaller one.
The assumption $T\ge512$ ensures $\varepsilon\le1/8$.
Finally, Eq.~\eqref{rn:eq:population-lower} converts the pseudo-regret into the normalized {expected-regret} lower bound $T^{2/3}/768$.
The policy was arbitrary, so this is a lower bound for the infimum over all policies followed by the supremum over iid demand laws.
No restriction to Algorithm~\ref{rn:alg:randomized-osgd} has been imposed.

For the third step, let $M,h>0$ be arbitrary and denote the normalized shortage cost and loss by $\psi_0$ and $\ell_0$, respectively.
Define $\psi(z)\coloneqq(2hM/9)\psi_0(3z/M)$ and scale every normalized demand and order by $M/3$.
For every $q,d\in[0,3]$, the corresponding scaled pair satisfies
\[
\begin{aligned}
\ell\!\Big({Mq/3},{Md/3}\Big)
=\frac{hM}{3}(q-d)_+ + \frac{2hM}{9}\psi_0((d-q)_+) 
=\frac{2hM}{9}\ell_0(q,d).
\end{aligned}
\]
Thus, quantities scale by $M/3$ and losses by $2hM/9$.
Sales scale by the same quantity factor because
$\min\{Mq/3,Md/3\}=(M/3)\min\{q,d\}$.
Hence, any policy in the scaled problem can be implemented in the normalized problem by rescaling its orders and observed sales.
This preserves nonanticipation and allows arbitrary policy randomization.
Since both cumulative loss and the best-fixed-order benchmark scale by $2hM/9$, regret scales by this factor as well.
The normalized lower bound therefore becomes
$(2hM/9)T^{2/3}/768=hMT^{2/3}/3456$,
proving Eq.~\eqref{rn:eq:general-lower}.
The scaled shortage cost satisfies Assumption~\ref{rn:ass:penalty} and has marginal cost $m(M)=4h/3\le\bar m$, so the constructed instance belongs to $\mathcal C$.
The shortage cost is fixed independently of $T$, and the lower bound applies even when it is known to the policy.
\hfill
\Halmos

\subsection{Faster regret under expected-loss curvature}
\label{rn:app:acceleration}
\label{rn:app:comparator}
\label{rn:app:accelerated-proof}
\label{rn:app:quadratic-lower}

Before proving Theorem~\ref{rn:thm:accelerated}, we establish an auxiliary result in Lemma~\ref{rn:lem:comparator}.

\begin{lemma}
\label{rn:lem:comparator}
Suppose demands are independent and either condition (i) or condition (ii) of Theorem~\ref{rn:thm:accelerated} holds with constant $\mu>0$.
Then, for every $T\ge1$,
\begin{equation}
\label{rn:eq:comparator-gap}
0\le\min_{q\in[0,M]}\sum_{t=1}^TL_t(q)
-\mathbb E\!\Big[\min_{q\in[0,M]}\sum_{t=1}^T\ell(q,\tilde d_t)\Big]
\le\frac{\pi G^2}{\mu}.
\end{equation}
\end{lemma}

\noindent
\textbf{Proof of \thmref{rn:lem:comparator}{Lemma}:}
The gap in Eq.~\eqref{rn:eq:comparator-gap} is the expected reduction in cumulative loss obtained by choosing the fixed order after observing demand rather than minimizing expected loss.
Let $\tilde f_T(q)\coloneqq\sum_{t=1}^T\ell(q,\tilde d_t)$ and
$\bar L_T(q)\coloneqq T^{-1}\sum_{t=1}^T L_t(q)$.
We use subscripts $+$ and $-$ on derivatives to denote right and left derivatives, respectively; for example $\bar L'_{T,+}$ and $\bar L'_{T,-}$ are the right and left derivatives, respectively.
Choose a minimizer $q^\star$ of $\bar L_T$; in case (ii), take the common minimizer.
Then
\[
\min_q\sum_{t=1}^T L_t(q)
-\mathbb E[\min_q\tilde f_T(q)]
=
\mathbb E[
\tilde f_T(q^\star)-\min_q\tilde f_T(q)].
\]
The expression inside the expectation is nonnegative, proving the lower bound.
We bound its expectation from above by controlling the possible reduction in realized loss on either side of $q^\star$.

Both structural conditions ensure that expected loss increases as the order moves away from $q^\star$.
More precisely, for $u\in(0,M)$,
\[
\bar L'_{T,+}(u)\ge\mu(u-q^\star)
\quad\text{if }u>q^\star,
\qquad
\bar L'_{T,-}(u)\le-\mu(q^\star-u)
\quad\text{if }u<q^\star.
\]
In case (i), $\bar L_T$ is $\mu$-strongly convex.
At an interior minimizer, zero is a valid subgradient, so strong monotonicity gives these inequalities.
At a boundary minimizer, the same argument uses
$\bar L'_{T,+}(0)\ge0$ or $\bar L'_{T,-}(M)\le0$.
In case (ii), average Eq.~\eqref{rn:eq:rsi} over $t$ and divide by $u-q^\star$, reversing the inequality when $u<q^\star$.
The resulting bounds on the average selected subgradient imply the stated slope bounds because that subgradient lies between the left and right derivatives of $\bar L_T$.
Neither argument requires the selected subgradient at $q^\star$ to be zero.

Consider first orders to the right of $q^\star$.
Increasing the order can reduce realized loss only where the realized slope is negative.
Since $\tilde f_T$ is absolutely continuous, for every $q\in[q^\star,M]$,
\[
\tilde f_T(q^\star)-\tilde f_T(q)
=
-\int_{q^\star}^q\tilde f'_{T,+}(u)\,\mathrm{d}u
\le
\int_{q^\star}^M
\bigl(-\tilde f'_{T,+}(u)\bigr)_+\,\mathrm{d}u.
\]
We therefore bound the expected negative part of this slope.

For fixed $u>q^\star$, $\tilde f'_{T,+}(u)$ is a sum of independent random variables, each bounded in absolute value by $G$.
Bounded difference quotients and dominated convergence give
$\mathbb E[\tilde f'_{T,+}(u)]
=T\bar L'_{T,+}(u)\ge T\mu(u-q^\star)$.
Thus, the slope can become negative only through a downward deviation from its positive mean.
For $z\ge0$, Hoeffding's inequality gives
$\mathbb P(\tilde f'_{T,+}(u)\le-z)
\le\exp\{-(T\mu(u-q^\star)+z)^2/(2TG^2)\}$.
Integrating this tail bound yields
\[
\begin{aligned}
\mathbb E[
\bigl(-\tilde f'_{T,+}(u)\bigr)_+
]
&\le
\int_0^\infty
\exp\Big\{
-\frac{(T\mu(u-q^\star)+z)^2}{2TG^2}
\Big\}\,\mathrm{d}z
\le
G\sqrt{\frac{\pi T}{2}}\,
\exp\Big\{
-\frac{T\mu^2(u-q^\star)^2}{2G^2}
\Big\}.
\end{aligned}
\]
The last inequality follows by separating the two squared terms in the exponent and evaluating the Gaussian integral.
Applying Tonelli's theorem to the preceding pathwise bound and extending the integration interval gives
\[
\begin{aligned}
\mathbb E\Big[
\sup_{q\in[q^\star,M]}
\{\tilde f_T(q^\star)-\tilde f_T(q)\}
\Big]
&\le
G\sqrt{\frac{\pi T}{2}}
\int_{q^\star}^\infty
\exp\Big\{
-\frac{T\mu^2(u-q^\star)^2}{2G^2}
\Big\}\,\mathrm{d}u
=
\frac{\pi G^2}{2\mu}.
\end{aligned}
\]

To the left of $q^\star$, reducing the order can lower realized loss only where the left derivative is positive.
For $q\le q^\star$, absolute continuity gives
$\tilde f_T(q^\star)-\tilde f_T(q)
=\int_q^{q^\star}\tilde f'_{T,-}(u)\,\mathrm{d}u$.
The possible reduction is therefore bounded by
$\int_0^{q^\star}(\tilde f'_{T,-}(u))_+\,\mathrm{d}u$.
Since
$\mathbb E[\tilde f'_{T,-}(u)]
\le-T\mu(q^\star-u)$,
the same argument, now using Hoeffding's upper-tail bound, shows that the expected reduction over all orders in $[0,q^\star]$ is at most $\pi G^2/(2\mu)$.
If $q^\star$ is an endpoint, the corresponding side contains only $q^\star$ and contributes zero.

The best order in hindsight lies on one of these two sides.
Its reduction in realized loss is the larger of the two one-sided reductions and is therefore at most their sum.
Adding the two bounds gives
$\mathbb E[\tilde f_T(q^\star)-\min_q\tilde f_T(q)]
\le\pi G^2/\mu$,
which proves the upper bound in Eq.~\eqref{rn:eq:comparator-gap}.
\hfill
\Halmos

\medskip

With the comparator bound established, we can prove the faster regret bound.

\medskip

\noindent
\textbf{Proof of \thmref{rn:thm:accelerated}{Theorem}:}
Recall that $q_T^\star$ is a deterministic minimizer of
$\sum_{t=1}^T L_t(q)$ over $[0,M]$.
In case (ii), we take $q_T^\star$ to be the common minimizer of the expected losses.

We first separate the policy's expected loss relative to $q_T^\star$ from the advantage of the hindsight comparator.
Under the independent-demand assumption and the conditional sampling rule, the current order is independent of current demand, so
$\mathbb E[\ell(\tilde q_t,\tilde d_t)]
=\mathbb E[L_t(\tilde q_t)]$.
Consequently,
\[
\begin{aligned}
\mathbb E[\tilde r_T]
={}&
\underbrace{
\mathbb E\Big[
\sum_{t=1}^T\{L_t(\tilde q_t)-L_t(q_T^\star)\}
\Big]
}_{A}
+
\underbrace{
\sum_{t=1}^T L_t(q_T^\star)
-\mathbb E\Big[
\min_{q\in[0,M]}\sum_{t=1}^T\ell(q,\tilde d_t)
\Big]
}_{B}.
\end{aligned}
\]
Lemma~\ref{rn:lem:comparator} bounds term B by $\pi G^2/\mu$.
It therefore remains to bound term A.

Both structural conditions give the following comparison:
\begin{equation}
\label{rn:eq:shared-curvature}
a\{L_t(x)-L_t(q_T^\star)\}
\le
\bar g_t(x)(x-q_T^\star)
-\frac{\mu}{2}(x-q_T^\star)^2,
\qquad x\in[0,M],
\end{equation}
where $a=1$ in case (i) and $a=1/2$ in case (ii).
In case (i), this is the strong-convexity inequality.
It holds for any comparison order, so $q_T^\star$ need not minimize each $L_t$ separately.
In case (ii), convexity gives
$L_t(x)-L_t(q_T^\star)\le\bar g_t(x)(x-q_T^\star)$,
while the restricted secant inequality gives
$0\le\bar g_t(x)(x-q_T^\star)-\mu(x-q_T^\star)^2$.
Adding these inequalities and dividing by two proves
Eq.~\eqref{rn:eq:shared-curvature} with $a=1/2$.

We now use the projected update to bound the loss difference in
Eq.~\eqref{rn:eq:shared-curvature}.
Conditional on $\mathcal F_{t-1}$, the virtual order $\tilde x_t$ and distribution $Q_t$ are fixed, and demand has law $P_t$.
Pointwise unbiasedness therefore gives
$\mathbb E[\tilde g_t\mid\mathcal F_{t-1}]
=\bar g_t(\tilde x_t)$.
Theorem~\ref{rn:thm:estimator} also gives
$\mathbb E[\tilde g_t^2\mid\mathcal F_{t-1}]
\le G^2+V_{\tilde x_t}(Q_t)$.
Indeed, conditional on the full past and current demand $\tilde d_t$, the estimator has mean $g(\tilde x_t;\tilde d_t)$ and variance at most $V_{\tilde x_t}(Q_t)$; the bound follows by using $|g(\tilde x_t;\tilde d_t)|\le G$ and taking conditional expectations (on the demand $\tilde d_t$) again.
These second moments are integrable because
$V_{\tilde x_t}(Q_t)\le2J_{\tilde x_t}(Q_t;\eta_t)/\eta_t$
and the assumed objective bound is finite.

Write $D_t\coloneqq\mathbb E[(\tilde x_t-q_T^\star)^2]$.
Projection onto $[0,M]$ cannot increase the distance from $q_T^\star$.
Expanding the squared distance after the update and using the conditional moment bounds yields
$D_{t+1}\le D_t
-2\eta_t\mathbb E[\bar g_t(\tilde x_t)(\tilde x_t-q_T^\star)]
+\eta_t^2\{G^2+\mathbb E[V_{\tilde x_t}(Q_t)]\}$.
Rearranging and applying Eq.~\eqref{rn:eq:shared-curvature} bounds the expected loss at the virtual order.
To account for the physical order, use
$L_t(\tilde q_t)-L_t(\tilde x_t)\le h\tilde u_t$.
Since $a\le1$, the resulting variance and inventory terms satisfy
$(\eta_t/2)\mathbb E[V_{\tilde x_t}(Q_t)]
+ah\mathbb E[\tilde u_t]
\le\mathbb E[J_{\tilde x_t}(Q_t;\eta_t)]$.
Thus,
\begin{equation}
\label{rn:eq:accelerated-step}
a\,\mathbb E[L_t(\tilde q_t)-L_t(q_T^\star)]
\le
\frac{D_t-D_{t+1}}{2\eta_t}
-\frac{\mu}{2}D_t
+\frac{\eta_tG^2}{2}
+\mathbb E[J_{\tilde x_t}(Q_t;\eta_t)].
\end{equation}
The choice $\eta_t^{-1}=\mu(t+t_0)$ cancels the intermediate $D_t$ terms when we sum over periods:
\[
\begin{aligned}
\sum_{t=1}^T
\Big\{
\frac{D_t-D_{t+1}}{2\eta_t}-\frac{\mu}{2}D_t
\Big\}
={}&
\Big(\frac1{2\eta_1}-\frac{\mu}{2}\Big)D_1+\frac12\sum_{t=2}^T
(\eta_t^{-1}-\eta_{t-1}^{-1}-\mu)D_t
-\frac{D_{T+1}}{2\eta_T}\\
={}&
\frac{\mu t_0}{2}D_1-\frac{D_{T+1}}{2\eta_T}
\le\frac{\mu t_0M^2}{2}.
\end{aligned}
\]
Here the second equality holds because
$\eta_t^{-1}-\eta_{t-1}^{-1}=\mu$ and the final inequality uses $D_1\le M^2$ and $D_{T+1}\ge0$.

The assumed one-period objective bound and $B_{\nu,\tilde x_t}\le B_\nu$ give
$J_{\tilde x_t}(Q_t;\eta_t)
\le\sqrt{2h\eta_t}\,B_\nu+\epsilon_t$.
Summing Eq.~\eqref{rn:eq:accelerated-step} over $t$ yields
\[
\begin{aligned}
a \mathbb E\!\left[
\sum_{t=1}^T\{L_t(\tilde q_t)-L_t(q_T^\star)\}
\right]
\le{}&
\frac{\mu t_0M^2}{2}
+\frac{G^2}{2\mu}\sum_{t=1}^T\frac1{t+t_0}
+\frac{\sqrt{2h}\,B_\nu}{\sqrt\mu}
\sum_{t=1}^T\frac1{\sqrt{t+t_0}}
+\sum_{t=1}^T\epsilon_t.
\end{aligned}
\]
Since the summands are decreasing, comparison with their integrals gives
$\sum_{t=1}^T(t+t_0)^{-1}\le\log(1+T/t_0)$
and
$\sum_{t=1}^T(t+t_0)^{-1/2}
\le2(\sqrt{T+t_0}-\sqrt{t_0})$.
Applying these inequalities and adding the bound
$\pi G^2/\mu$ of term $B$ proves
Eq.~\eqref{rn:eq:accelerated-regret}.
For fixed problem parameters, the resulting bound is $O(\sqrt T)$ whenever
$\sum_{t=1}^T\epsilon_t=O(\sqrt T)$.
\hfill
\Halmos

\medskip

We next prove the corresponding minimax lower bound by applying the information and loss-gap inequalities to a quadratic shortage {cost}.

\medskip

\noindent
\textbf{Proof of \thmref{rn:thm:accelerated-lower}{Theorem}:} We prove the result in three steps.
First, we prove the $\sqrt T$ lower bound on a normalized instance using the demand pair and information inequalities from Appendix~\ref{rn:app:information}.
Second, we extend the construction to arbitrary $M,h,\mu>0$ by rescaling and shifting demands and orders, and show that the lower bound applies to all policies on $[0,M]$.
Third, we verify the constructed instance satisfies both structural conditions in the statement.

For the first step, take $M=3$, $h=3/2$, and $\psi(z)=z^2/2$.
Use the demand pair in Eq.~\eqref{rn:eq:hard-pair} with $\varepsilon=T^{-1/4}$.
The condition $T\ge4096$ ensures $0<\varepsilon\le1/8$.
For the quadratic {shortage cost}, $m(z)=z$.
When $q\in(0,2)$, only demand zero lies below the order, and direct substitution gives
\[
L_\sigma'(q)=\frac12(q-1)-\sigma\varepsilon=\frac12(q-q_\sigma^\star),
\qquad q_\sigma^\star \coloneqq 1+2\sigma\varepsilon.
\]
Both optimizers lie in $(0,2)$, and convexity makes them global minimizers.
Integrating the derivative from $q_\sigma^\star$ to $q$ yields $L_\sigma(q)-L_\sigma(q_\sigma^\star)=(q-q_\sigma^\star)^2/4$ for every $0\le q\le2$.
For $2<q<3$, demands zero and two lie below the order, and only demand three contributes a shortage term.
Thus $L_\sigma'(q)=(3/2)(3/4-\sigma\varepsilon)+(1/4+\sigma\varepsilon)(q-3) \geq 7/8-5\varepsilon/2\ge9/16$ on the interval.
Hence, the expected losses under both demand profiles are increasing beyond two.

Take $A \coloneqq [1,3]$.
If $q\notin A$, then $q<1$ and its distance from $q_+^\star=1+2\varepsilon$ is at least $2\varepsilon$.
The quadratic gap formula therefore gives $L_+(q)-L_+(q_+^\star)\ge\varepsilon^2$.
If $q\in[1,2]$, its distance from $q_-^\star=1-2\varepsilon$ is at least $2\varepsilon$, so the same formula gives $L_-(q)-L_-(q_-^\star)\ge\varepsilon^2$.
If $q>2$, monotonicity beyond two gives a gap of at least $L_-(2)-L_-(q_-^\star)=(1+2\varepsilon)^2/4\ge\varepsilon^2$.
Hence the two wrong-region conditions in Lemma~\ref{rn:lem:testing} hold with $\delta_\varepsilon=\varepsilon^2$.

Every informative order $q>2$ satisfies $L_+(q)-L_+(q_+^\star)\ge L_+(2)-L_+(q_+^\star)=(1-2\varepsilon)^2/4\ge9/64$.
Thus the informative-order condition holds with $c_H=9/64$.
Lemma~\ref{rn:lem:testing} gives
\[
\max_{\sigma\in\{+,-\}}\mathcal E_\sigma^\pi(T)
\ge\min\left\{\frac3{4096}\varepsilon^{-2},\frac38\varepsilon^2T\right\}.
\]
Unlike the piecewise-linear instance, the wrong-region loss is quadratic in the separation between the optimal orders.
Choosing $\varepsilon=T^{-1/4}$ makes both terms proportional to $\sqrt T$, and the first is the smaller one.
By Eq.~\eqref{rn:eq:population-lower}, every policy in the normalized problem therefore satisfies $\max_\sigma\mathbb E_\sigma^\pi[\tilde r_T]\ge(3/4096)\sqrt T$.

For the second step, let $M,h,\mu>0$ be arbitrary and set $b \coloneqq \min\{M,h/(4\mu)\}$.
Define the shortage {cost} by $\psi(z) \coloneqq hz^2/b$ for every $z\in[0,M]$.
Let $P_\sigma^b$ be the distribution obtained by transforming a demand $d$ from Eq.~\eqref{rn:eq:hard-pair} into $M-b+(b/3)d$.
Its support is $\{M-b,M-b+2b/3,M\}$, with the same three probabilities as in the normalized instance.
Write $\ell_0$ for the normalized loss with excess-inventory coefficient $3/2$ and shortage {cost} $z^2/2$.
For normalized orders and demands $q,d\in[0,3]$, direct substitution gives
\begin{equation}
\label{rn:eq:translated-loss}
\ell\!\Big(M-b+\frac b3q,M-b+\frac b3d\Big)
=h \frac{b}{3}(q-d)_+
+\frac hb\Big\{\frac b3(d-q)_+\Big\}^2
=\frac{2hb}{9}\ell_0(q,d).
\end{equation}
Sales transform by the same affine map, because the minimum of the transformed order and demand is $M-b+(b/3)\min\{q,d\}$.
Consequently, a policy restricted to $[M-b,M]$ corresponds to a policy on $[0,3]$ with exactly the same normalized observation history.

Take any policy $\pi$ with orders in $[0,M]$.
Construct $\hat\pi$ by raising every proposed order below $M-b$ to $M-b$ and implementing all other proposals unchanged.
For subsequent decisions, update $\pi$ using its proposed order and the sale that order would have produced.
Specifically, for a proposal $q<M-b$, record $(q,q)$, since every transformed demand is at least $M-b$; otherwise, record the actual order and sale.
Each replacement also strictly reduces loss: for $q<M-b$ and every transformed demand $d$,
$\ell(M-b,d)=\psi(d-(M-b)) < \psi(d-q)=\ell(q,d)$.
Thus, $\hat\pi$ incurs no greater cumulative loss than $\pi$.
The hindsight benchmark is also unchanged when orders are restricted to $[M-b,M]$, because every fixed order below $M-b$ is dominated by $M-b$.
Consequently, $\hat\pi$ incurs no greater regret than $\pi$.

By construction, $\hat\pi$ orders only in $[M-b,M]$, so the restricted-policy lower bound applies.
Under at least one transformed demand law, its expected regret is at least $(hb/6144)\sqrt T$.
The expected regret of $\pi$ is no smaller under that same law.
Since $\pi$ was arbitrary, the lower bound holds for all policies on $[0,M]$.

For the third step, we show that the constructed instances satisfy the conditions in Theorem~\ref{rn:thm:accelerated}. 
Write $L_\sigma^b(q) \coloneqq \mathbb E_{P_\sigma^b}[\ell(q,\tilde d)]$.
The {shortage cost} $\psi(z)=hz^2/b$ is $2h/b$-strongly convex on all of $[0,M]$, not only on $[0,b]$.
Also, $P_\sigma^b(\{M\})=1/4+\sigma\varepsilon\ge1/8$.
On this event, the realized loss $\psi(M-q)$ is $2h/b$-strongly convex throughout the full decision interval $[0,M]$; all other realized losses are convex.
Averaging therefore makes each $L_\sigma^b$ at least $h/(4b)$-strongly convex on $[0,M]$.
Since $b\le h/(4\mu)$, this modulus is at least $\mu$, proving condition (i).
Strong convexity also implies the restricted secant inequality around the minimizer on $[0,M]$, with the same constant $h/(4b)\ge\mu$.
Within each instance, demand is iid, so all periodwise expected losses share this minimizer.
Thus condition (ii) holds as well.

The {shortage cost} satisfies Assumption~\ref{rn:ass:penalty} and depends only on $M,h,\mu$, not on $T$ or the sign $\sigma$.
Its marginal cost satisfies $m(M)=2hM/b=\max\{2h,8\mu M\}\le\bar m$.
Thus the transformed pair belongs to both stated structural classes, with one fixed known quadratic {shortage cost}.
Taking the supremum over either class and then the infimum over policies gives Eq.~\eqref{rn:eq:quadratic-lower}, because $b=\min\{M,h/(4\mu)\}$.
When $\mu\le h/(4M)$, the construction has $b=M$ and reduces to ordinary scaling; larger $\mu$ is covered by the translation and domination argument above.
\hfill
\Halmos

\subsection{Sufficient conditions from shortage costs and demand}
\label{rn:app:sufficient}

\noindent
\textbf{Proof of \thmref{rn:prop:sufficient}{Proposition}:} We prove the three parts in order, using mass at the maximum demand, mass between two order levels, and tail mass above the common minimizer, respectively.
Throughout, fix a period $t$; all lower bounds are uniform in $t$.

For part (i), the slope condition makes $\psi$ $\gamma$-strongly convex on $[0,M]$.
Indeed, the right derivative of $z\mapsto\psi(z)-\gamma z^2/2$ is $m(z)-\gamma z$, which is nondecreasing by assumption, so this function is convex, which implies that $\psi$ is $\gamma$-strongly convex.
On the event $\tilde d_t=M$, the realized loss is $\ell(q,M)=\psi(M-q)$ throughout $[0,M]$, so it is $\gamma$-strongly convex in $q$.
For every other demand realization, the loss remains convex.
Averaging over $P_t$ therefore makes $L_t$ at least $\gamma P_t(\{M\})$-strongly convex, and the assumed mass bound gives the modulus $\gamma p_0$ uniformly in $t$.

For part (ii), fix $0\le x<y\le M$.
The selected subgradient $g(q;d)$ is nondecreasing in $q$ for every fixed $d$.
If $x\le d<y$, its increase is $g(y;d)-g(x;d)=h+m(d-x)\ge h+m(0)$; for all other demands its increase is nonnegative.
Taking expectations and using the density lower bound therefore gives
\[
\begin{aligned}
\bar g_t(y)-\bar g_t(x)
&\ge(h+m(0))P_t([x,y)) 
\ge
(h+m(0))\int_x^y f_t(v)\,\mathrm{d}v
\ge(h+m(0))(y-x)f_0.
\end{aligned}
\]
This inequality uses only the absolutely continuous part to obtain a lower bound; any atoms or singular continuous component make an additional nonnegative contribution.

Write $\mu \coloneqq (h+m(0))f_0$ for this part.
The function $L_t$ is $G$-Lipschitz and hence absolutely continuous, and its selected subgradient agrees with its derivative almost everywhere.
For $x<y$, integrating $\bar g_t(v)\ge\bar g_t(x)+\mu(v-x)$ over $v\in(x,y)$ gives
\[
L_t(y)-L_t(x)
=\int_x^y\bar g_t(v)\,\mathrm{d}v
\ge\bar g_t(x)(y-x)+\frac\mu2(y-x)^2.
\]
For $y<x$, integrate $\bar g_t(v)\le\bar g_t(x)-\mu(x-v)$ over $v\in(y,x)$ and reverse the sign of the integral to obtain the same inequality.
The case $x=y$ is immediate, so $L_t$ is $\mu$-strongly convex on $[0,M]$.

For part (iii), the slope condition again makes $\psi$ $\gamma$-strongly convex.
Fix a demand realization $d\ge b$.
For every $x\in[0,b]$, the realized loss is $\ell(x,d)=\psi(d-x)$, which is $\gamma$-strongly convex in $x$ on this interval.
For realizations with $d<b$, the loss remains convex there.
Averaging over $P_t$ therefore makes $L_t$ at least $\gamma P_t([b,M])$-strongly convex on $[0,b]$, and thus at least {$\gamma p_0$}-strongly convex.

Optimality of $q^\star$ and strong convexity give, for $x\in[0,b]$ and $0<\lambda<1$,
\(L_t(q^\star)
\le L_t((1-\lambda)q^\star+\lambda x)
\le
(1-\lambda)L_t(q^\star)+\lambda L_t(x)
-\frac{\gamma p_0}{2}\lambda(1-\lambda)(x-q^\star)^2.\)
Subtracting $L_t(q^\star)$, dividing by $\lambda$, and letting $\lambda\downarrow0$ yields
$L_t(x)-L_t(q^\star)\ge(\gamma p_0/2)(x-q^\star)^2$.
This argument also holds at $q^\star=0$.
Applying the strong-convexity inequality at $x$, with subgradient $\bar g_t(x)$, now gives
\[
\begin{aligned}
\bar g_t(x)(x-q^\star)
\ge L_t(x)-L_t(q^\star)
+\frac{\gamma p_0}{2}(x-q^\star)^2
\ge\gamma p_0(x-q^\star)^2,
\qquad x\in[0,b].
\end{aligned}
\]
Since the stated constant
$\mu=\gamma p_0(b-q^\star)/(M-q^\star)$
is no larger than $\gamma p_0$, Eq.~\eqref{rn:eq:rsi} holds on $[0,b]$.

It remains to consider $x>b$ when $b<M$.
Monotonicity gives $\bar g_t(x)\ge\bar g_t(b)$, and the preceding inequality at $b$ gives {$\bar g_t(b)\ge\gamma p_0(b-q^\star)$} because $b-q^\star>0$.
Consequently,
\begingroup\color{black}
\[
\begin{aligned}
\bar g_t(x)(x-q^\star)
&\ge\gamma p_0(b-q^\star)(x-q^\star)
\ge
\gamma p_0\frac{b-q^\star}{M-q^\star}(x-q^\star)^2.
\end{aligned}
\]
\endgroup
The second inequality uses $x-q^\star\le M-q^\star$.
This proves Eq.~\eqref{rn:eq:rsi} on the remaining interval, uniformly over periods.
\hfill
\Halmos

\subsection{Extension to nonperishable inventory}\label{app:non-perishable}


\noindent
\textbf{Proof of \cref{prop:non-perishable-bound}:} 
Recall that $\tilde e_t\coloneqq\tilde q_t-(\tilde x_t+\tilde u_t)$.
As shown in the main text,
\[
\tilde e_{t+1}
\le (\tilde e_t+\tilde u_t-\tilde d_t)_+
+h_{\mathrm H}\eta_t 
\le (\tilde e_t+\tilde u_t)
\Big(1-\frac{\tilde d_t}{M}\Big)
+h_{\mathrm H}\eta_t.
\]
The second inequality follows from $(z-d)_+\le z(1-d/M)$ for $z,d\in[0,M]$, since
$0\le\tilde e_t+\tilde u_t=\tilde q_t-\tilde x_t\le M$
and $\tilde d_t\in[0,M]$.

Set $\rho\coloneqq1-\underline d/M\in[0,1)$.
Conditional on $\mathcal H_{t-1}$ and $\tilde u_t$, $\tilde e_t+\tilde u_t$ is fixed and nonnegative.
Since $\tilde d_t$ and $\tilde u_t$ are conditionally independent given $\mathcal H_{t-1}$,
$\E[\tilde d_t\mid\mathcal H_{t-1},\tilde u_t]
=\E[\tilde d_t\mid\mathcal H_{t-1}]
\ge\underline d$.
Taking conditional expectations and then averaging gives
\[
\begin{aligned}
\E[\tilde e_{t+1}]
&\le
\E\!\Big[
(\tilde e_t+\tilde u_t)
\Big(1-\frac{\E[\tilde d_t\mid\mathcal H_{t-1},\tilde u_t]}{M}\Big)
\Big]
+h_{\mathrm H}\eta_t 
\le
\rho\E[\tilde e_t]
+\rho\E[\tilde u_t]
+h_{\mathrm H}\eta_t.
\end{aligned}
\]

Let $S_T\coloneqq\sum_{t=1}^T\E[\tilde e_t]$.
Summing over $t=1,\ldots,T-1$ gives
\[\sum_{t=1}^{T-1}\E[\tilde e_{t+1}]
\le
\rho\sum_{t=1}^{T-1}\E[\tilde e_t]
+\rho\sum_{t=1}^{T-1}\E[\tilde u_t]
+h_{\mathrm H}\sum_{t=1}^{T-1}\eta_t.\]
Since $\tilde e_1=0$, the sum on the left equals $S_T$ and $\sum_{t=1}^{T-1}\E[\tilde e_t] = S_T-\E[\tilde e_T]$.
Thus,
\[S_T
\le
\rho\bigl(S_T-\E[\tilde e_T]\bigr)
+\rho\sum_{t=1}^{T-1}\E[\tilde u_t]
+h_{\mathrm H}\sum_{t=1}^{T-1}\eta_t.\]
Rearranging, using $\tilde e_T\ge0$, and dividing by $1-\rho=\underline d/M$ establishes
\[
S_T
\le
\Big(\frac{M}{\underline d}-1\Big)
\sum_{t=1}^{T-1}\E[\tilde u_t]
+
\frac{h_{\mathrm H}M}{\underline d}
\sum_{t=1}^{T-1}\eta_t.
\]
which is the desired result.
\hfill\Halmos